\documentclass[twoside,11pt]{article}

\usepackage{blindtext}

\usepackage[preprint]{jmlr2e}

\usepackage{algorithm}
\usepackage{algpseudocode}
\usepackage{float}
\usepackage{tabularx}
\usepackage{multicol}

\usepackage{amsmath}

\newcommand{\KL}{\operatorname{KL}\!}

\usepackage{todonotes}

\usepackage{enumitem}
\usepackage{booktabs}

\usepackage{xcolor, mdframed}

\usepackage{lastpage}

\ShortHeadings{Transfer Learning in Bandits}{Prevost et al.}
\firstpageno{1}

\begin{document}

\title{Transfer Learning in Bandits from Unknown Adaptive Policies: Problem-Dependent Lower and Upper Bounds}

\author{\name Adrien PREVOST \email adrien.prevost@inria.fr \\
       \addr Equipe Scool\\
       Univ. Lille, Inria, CNRS, Centrale Lille\\
       UMR 9189-CRIStAL, F-59000 Lille, France
       \AND
       \name Timothee MATHIEU \email timothee.mathieu@inria.fr \\
        \addr Equipe Scool\\
       Univ. Lille, Inria, CNRS, Centrale Lille\\
       UMR 9189-CRIStAL, F-59000 Lille, France
       \AND
       \name Waris RADJI \email waris.radji@inria.fr \\
       \addr Equipe Scool\\
       Univ. Lille, Inria, CNRS, Centrale Lille\\
       UMR 9189-CRIStAL, F-59000 Lille, France
       \AND
       \name Odalric-Ambrym MAILLARD \email odalric.maillard@inria.fr \\
       \addr Equipe Scool\\
       Univ. Lille, Inria, CNRS, Centrale Lille\\
       UMR 9189-CRIStAL, F-59000 Lille, France}

\maketitle

\begin{abstract}
Transfer learning in stochastic multi-armed bandits is typically studied assuming that prior data forms a fixed offline dataset or is generated by a known source. In practice, historical observations come from a previous adaptive decision process whose policy is unknown. We study this setting for bounded rewards in $[0,1]$.

Before the online game starts, the learner receives, for each arm $k$, $N'_k$ observations drawn i.i.d.\ from an unknown distribution $\nu'_k$, collected by an arbitrary and possibly data-dependent strategy, so the sample sizes $N'_k$ may be random. Source and target means satisfy $|\mu_k-\mu'_k|<L_k$ for known constants $(L_k)_k$.

We derive an asymptotic problem-dependent lower bound and propose KL-UCB-Transfer, which combines online and historical data and achieves logarithmic regret with an improved leading constant whenever $\mu'_a<\mu_1-L_a$, and recovers the classical KL-UCB guarantee otherwise. Since bounding the source-target discrepancy is often unrealistic, we relax the assumption to the one-sided condition $\mu_k-\mu'_k<L_k$ and show that logarithmic regret can survive under misspecification, characterizing three regimes ranging from improvement to a degraded logarithmic bound to a loss of guarantee. Simulations illustrate the practical benefits of exploiting process-generated historical data.
\end{abstract}


\section{Introduction}

Multi-armed bandit (MAB) algorithms provide a standard framework for sequential decision making under uncertainty, with applications ranging from clinical trials \citep{Robertson2023} and agriculture \citep{GAUTRON2024109249} to recommendation systems, algorithm selection, and hyperparameter optimization \citep{9185782}. These methods are particularly well suited to settings where data are scarce, costly, or slow to collect, which makes efficient exploration--exploitation strategies essential.

In many practical situations, however, learning does not start from scratch. Historical observations are often available before the online interaction begins and may substantially accelerate learning if they can be exploited reliably. Most existing transfer-learning approaches assume that these observations form a fixed offline dataset, or that the process generating them is known. In practice, however, historical data is often produced by a previous adaptive decision process whose policy is unknown. A previous bandit algorithm, for instance, naturally allocates more observations to promising arms than to poor ones. This raises the question of whether such process-generated data can still be exploited while retaining the asymptotic guarantees of classical bandit algorithms.

Motivated by this observation, we consider a variant of the stochastic bandit setting in which, before the online interaction begins, the learner receives, for each arm, a batch of observations drawn from a source distribution. The learner has no access to the policy that generated these observations. Consequently, the number of available samples for each arm may be random and depend on the outcomes of the previous decision process. We further allow the source and target distributions to differ, assuming only that their means satisfy the discrepancy condition $|\mu_k-\mu'_k|<L_k$ for known constants $(L_k)_k$. We refer to this setting as \emph{transfer with process-generated samples}. A typical example is a farmer choosing among several fertilization strategies on a given plot of land using historical yield data collected on a different plot. Since the two plots are related but not identical, the historical observations only transfer approximately.

As we discuss in detail in \autoref{sec:related}, existing work on transfer learning in bandits typically relaxes only one or two of these difficulties at a time. Prior data is generally assumed to be an offline dataset of fixed size \citep{prevost2025asymptoticallyoptimalproblemdependentbandit}, or the relation between source and target is described through a contextual or causal structure \citep{10.1214/23-AOS2341, ijcai2017p186}, or the source policy is known and coordinated with the learner \citep{barnea2025individual}. Our setting relaxes all these assumptions simultaneously. The prior data may originate from an arbitrary, unknown, and possibly data-dependent policy, while the source and target environments are connected only through the discrepancy bounds $(L_k)_k$.
\paragraph{Our contributions} We summarize our main contributions below:
\begin{enumerate}
\item[(i)] \textbf{Lower bound.} We prove an asymptotic problem-dependent lower bound for transfer learning with an unknown, possibly adaptive source policy $\pi'$. Existing results require offline data of fixed size. We generalize them to distributions in $[0,1]$ (\autoref{sec:lowerbound}).
\item[(ii)] \textbf{Algorithm.} We adapt the KL-UCB-Transfer algorithm of \citet{prevost2025asymptoticallyoptimalproblemdependentbandit}, originally proposed for Gaussian rewards with known variance, offline prior observations, and deterministic sample sizes, to the case of bounded distributions with process-generated prior samples (\autoref{sec:algo}).
\item[(iii)] \textbf{Regret upper bound.} We provide an upper bound on the regret of KL-UCB-Transfer (Theorem \ref{thm:regret1}).
\item[(iv)] \textbf{Robustness to misspecification.} Motivated by the practical difficulty of accurately bounding discrepancy in real-world scenarios, we show that the symmetric condition $|\mu_k - \mu'_k| < L_k$ can be relaxed to the one-sided condition $\mu_k - \mu'_k < L_k$. Furthermore, through a unified analysis based on an effective transfer level $\mu^\dagger$ (Theorem~\ref{thm:general}), we characterize precisely how regret behaves across three regimes: well-specified (best constant), mildly misspecified (logarithmic regret with degraded constant), and severely misspecified (\autoref{sec:robust}).
\item[(v)]\textbf{Extension to multiple sources.} We show that the analysis of Theorem \ref{thm:general} extends to aggregating $M-1$ independent prior sources, each with its own discrepancy
bound, and recovers Theorem \ref{thm:general} as the special case $M=2$ (\autoref{sec:multisource}).
\item[(vi)] \textbf{Empirical validation.} Numerical experiments illustrate the practical benefits of exploiting process-generated historical data from past decision processes (\autoref{sec:simu}).
\end{enumerate}

\paragraph{Outline} \autoref{sec:related} discusses related work in more detail. \autoref{sec:formulation} formalizes the transfer setting and notation. \autoref{sec:lowerbound} states our lower bound. \autoref{sec:algo} recalls KL-UCB and introduces KL-UCB-Transfer. \autoref{sec:regret} states our regret upper bound, and \autoref{sec:robust} studies its robustness to misspecification of $L$. \autoref{sec:multisource}  extends the analysis to multiple independent prior sources. \autoref{sec:simu} reports numerical experiments, and \autoref{sec:conclusion} concludes.

\section{Related work and context}\label{sec:related}

Table~\ref{tab:related-work} summarizes how existing lines of work exploit side information in bandit problems, along the three dimensions that matter most for our setting: the nature of the source data, the mechanism linking source and target, and whether the policy generating the source data is known. We discuss each line of work in more detail below.

\begin{table}[h]
\centering
\small
\caption{Representative transfer learning settings in stochastic bandits.}
\label{tab:related-work}
\begin{tabularx}{\textwidth}{@{}l X X X@{}}
\toprule
Work & Source data & Link source-target & Source policy \\
\midrule
Bayesian & Prior distribution & Bayesian prior & --- \\
Contextual & Offline samples & Context & Fixed offline \\
Causal & Offline samples & DAG & Fixed offline \\
Cooperative & Online interaction & Same environment & Known \\
Fixed-size transfer & Offline samples & Mean discrepancy & Fixed offline \\
\textbf{Ours} & \textbf{Offline samples} & \textbf{Mean discrepancy} & \textbf{Unknown (adaptive)} \\
\bottomrule
\end{tabularx}
\end{table}

\paragraph{Classical multi-armed bandits}
The multi-armed bandit (MAB) problem has a long history, with early contributions in the statistics and clinical-trial literature \citep{robbins1952some,dc35850b-2ca1-314f-9e0d-470713436b17}. The problem-dependent asymptotic lower bound was established in \cite{LAI19854}, and a variety of algorithms attain this bound in different settings. For bounded rewards, KL-UCB achieves this optimality \citep{Cappe2013KL}. Nonparametric Thompson Sampling have also been shown to match the lower bound in bounded-reward settings \citep{riou2020bandit}. Alternative algorithms such as DMED and IMED also match this optimality for bounded rewards \citep{Honda2010AnAO,JMLR:v16:honda15a}.

\paragraph{Bayesian approaches to bandits}
A natural way to incorporate side information about the arms is to encode it as a prior in a Bayesian formulation. Bayesian UCB and Thompson Sampling variants are known to be asymptotically optimal under mild conditions on the prior \citep{kaufmann2012bayesian}, and this optimality typically holds regardless of the specific prior chosen: the influence of the prior on the regret vanishes as $T\to\infty$, since it is eventually overwhelmed by the accumulating data \citep{atsidakou2023finite}. Nonparametric extensions relax the need for a parametric prior on the reward distribution altogether, placing a prior directly on the space of distributions, e.g.\ via Dirichlet Processes \citep{vashishtha2025leveraging}, while retaining similar asymptotic guarantees through information-theoretic analyses in the spirit of \citet{russo2016information}. These approaches could be used in our setting but raise several problems. It is not clear how to choose the prior to encode the prior distribution $\hat \nu'_k$ with our condition on the mean $| \mu_k - \mu'_k| < L_k$ and in our setting this side information can remain influential in the regret as $T\to\infty$ whenever $N'_k$ itself scales with $T$ while in the current analysis of Bayesian approach it is used that the prior become asymptotically negligible.

\paragraph{Transfer learning in contextual and causal bandits}
Beyond encoding prior belief through a distribution, another route to exploiting side information is to transfer directly from related, previously observed environments, an approach studied extensively in contextual bandits and causal settings. For example, \cite{deng2025transferlearninglatentcontextual} analyze an offline contextual dataset subject to covariate shift and propose algorithms that exploit the prior data; \cite{10.1214/23-AOS2341} derive minimax lower bounds and near-optimal algorithms for Hölder-smooth reward functions in a contextual transfer setting. Recent work on causal bandits studies structural transfer via graphical models and proposes Thompson-Sampling–type procedures with regret analyses \citep{NEURIPS2023_8a8ce53b,ijcai2017p186}. These approaches crucially rely on contextual or causal structure to transfer information effectively which can be either unknown or too complex to derive tight problem-dependent bounds.

\paragraph{Transfer learning in non-contextual bandits}
The non-contextual case, closer to ours, has also attracted attention. Early work \cite{NIPS2013_062ddb6c} considered a finite set of candidate models and obtained problem-dependent regret bounds. Recurrent/episodic settings where multiple related bandit instances are played have been studied in \cite{10.1007/978-3-030-86486-6_1}, with matching lower bounds.

\paragraph{Multi-agent bandits}
Another way in which prior data could be used is from another agent playing a related bandit instance, as studied in multi-agent (or cooperative / collaborative) bandit problems, in which several agents interact with the same, or related, instances and exchange information, typically over a communication graph or protocol \citep{barnea2025individual, baron2019individual, yi2022regretoptimal, wang2022thompson,asadi2022collaborative}. These works study how communication constraints affect either the collective (group) regret or the individual regret of each agent, and the algorithms are co-designed so that every agent's policy, and the communication or aggregation protocol itself, are known and controlled by the analysis. This is the main point of departure from our setting: we do not design, coordinate with, or even observe the strategy $\pi'$ that produced the prior data, nor do we assume it interacts with our target environment $\nu$ in any structured way beyond the discrepancy bound $L_k$. To the best of our knowledge, existing multi-agent bandit models do not consider a setting in which one agent's data-generating strategy is neither known nor coordinated with the learner's, which is precisely the situation we study when transferring from an arbitrary, unobserved policy $\pi'$.

\paragraph{Closest related work}
Few works come particularly close to our setting and merit a closer comparison. \cite{sentenac2025balancingoptimismpessimismofflinetoonline} is close to our setting as they start with offline data. However the offline data comes from the same distributions of the game, and their goal is to minimize the regret with a known $T$ by trading off between exploiting the best arm with only offline data and exploring arms. Structured algorithms are also quite close to our setting \citep{pmlr-v119-degenne20b,combes2017minimalexplorationstructuredstochastic}, especially the algorithm for Lipschitz bandits in \cite{magureanu2014lipschitzbanditsregretlower}, which is close to our algorithm; however, they work with Bernoulli distributions, add forced exploration, and assume structure on the arms, without offline data, while we do not assume structure between the arms but have prior data from an unknown strategy. \cite{rahul2024transfer,rahul2024exploiting} study a similar structure across sequential episodes, transferring raw reward samples via UCB-type algorithms under a known (or online-estimated) bound on the mean-reward shift between consecutive tasks, however it is only for Gaussian distribution with fixed variance with a regret bound far from the optimal one. Closest to our setting are \cite{pmlr-v235-cheung24a} and \cite{prevost2025asymptoticallyoptimalproblemdependentbandit}, who consider the same setting where we get samples from a close distribution $\nu'_a$ such that $|\mu_a - \mu'_a| < L_a$ except they consider offline data and Gaussian distribution with fixed variance. The first one notably establishes an impossibility result in the absence of a known discrepancy bound while the second one gives an algorithm that has a regret that matches the problem-dependent lower bound with the exact constant. In comparison, the prior sample size $N'_k$ in our setting is not assumed fixed in advance: it may be random and generated by an unknown, possibly data-dependent policy $\pi'$ and our set of distributions is the nonparametric set of distributions in $[0,1]$.

\paragraph{Gap with the literature}
Across these lines of work, a recurring pattern emerges: exploiting side information about the bandit problem typically requires either (i) encoding it through a prior whose influence vanishes asymptotically, (ii) assuming a structural relationship (context, causal graph) between source and target, (iii) assuming the side information is offline and fixed rather than generated by an ongoing, unknown decision process, or (iv) assuming that any other data-generating agent follows a policy known to, and coordinated with, the learner. Even the closest existing results relax only one or two of these assumptions at a time: \citet{sentenac2025balancingoptimismpessimismofflinetoonline} keep the offline data on-distribution and rely on a known horizon; \citet{magureanu2014lipschitzbanditsregretlower} allow no prior data but assume structure between the arms; \citet{prevost2025asymptoticallyoptimalproblemdependentbandit} allow prior data but require it to be deterministic in size and drawn i.i.d.\ from a known family. Our setting relaxes all four assumptions simultaneously: the prior data may come from an arbitrary, unobserved, and possibly data-dependent policy $\pi'$, with no contextual or causal structure and no vanishing influence, connected to the target only through a known discrepancy bound $L_k$. This combination is, to the best of our knowledge, not covered by existing results, and requires new proof techniques to obtain both problem-dependent lower and upper bounds.

\section{Problem formulation}\label{sec:formulation}
In this section, we introduce the setting formally, together with notation and assumptions.

\subsection{Classical setting}

Let $\mathcal{D}$ be a known family of real-valued distributions and let the unknown arm distributions be $\nu = (\nu_1, \ldots, \nu_K) \in \mathcal{D}^K$ with means $\mu_k = \mathbb{E}_{X \sim \nu_k}[X]$. Although the classical framework recalled below holds for a general family $\mathcal{D}$, throughout this paper we specialize to $\mathcal{D} = \mathcal{P}([0,1])$, the set of distributions supported on $[0,1]$ with mean in $[0,1)$.

\paragraph{Notational convention} Throughout the paper, for any distribution (possibly decorated with a tilde, prime, or hat, and possibly empirical) we denote by the corresponding decorated $\mu$ its mean. For instance, $\mu'_k := \mathbb{E}_{X \sim \nu'_k}[X]$, $\tilde\mu_k := \mathbb{E}_{X \sim
\tilde\nu_k}[X]$, and $\hat\mu_a(t) := \mathbb{E}_{X \sim \hat\nu_a(t)}[X]$, where $\hat\nu_a(t)$ denotes the empirical distribution of arm $a$ at time $t$, introduced in \autoref{sec:algo}. We use this convention silently for any such decoration introduced later in the paper.
\\
\\
At each round $t \in \{1, \ldots, T\}$ a learner $\pi$ selects an arm $a_t$ and observes reward $X_t \sim \nu_{a_t}$. Denoting by $\mathbf{1}(\cdot)$ the indicator function, we let
$$
N_k(T) = \sum_{t=1}^T \mathbf{1}(a_t = k)
$$
denote the number of times arm $k$ has been pulled up to time $T$. The horizon $T$ is not assumed to be known by the learner, which is hence anytime. 
\\
\\
We assume without loss of generality that arm $1$ is the arm with the highest mean, that is $\mu_1 = \max_k \mu_k$. This arm is called an optimal arm, while other arms with smaller means are the suboptimal arms. We further assume that there is a single optimal arm (hence arm $1$), a standard assumption in the bandit literature (see e.g.\ \citealp{lattimore2020bandit}).
\\
The expected cumulative regret of a learner policy $\pi$ over $T$ rounds interacting with $\nu$ is
$$
R^\pi_{T,\nu} = \sum_{k=1}^K \mathbb{E}_{\nu,\pi}[N_k(T)](\mu_1 - \mu_k).
$$

\subsection{Transfer learning setting}\label{sec:subtransfer}

We now formalize the setting we refer to as \emph{transfer with process-generated samples}. Before interacting with the target environment $\nu$, for each arm $k$ the learner observes $N'_k$ prior samples drawn i.i.d.\ from a source distribution $\nu'_k$. The counts $(N'_k)_k$ may themselves come from an online game, making them data-dependent on the prior observations.
\\
\\
We denote $\nu' = (\nu'_1, \ldots, \nu'_K) \in \mathcal{D}'^K$ the corresponding source distributions, which may differ from the target $\nu$. We assume a known discrepancy bound between the mean of the source game and the target game:
$$
|\mu_k - \mu'_k| < L_k, \qquad k = 1, \ldots, K,
$$
where $(L_k)_k$ are given constants.

\paragraph{Prior data} Let $\mathcal{X}'$ denote the set of received data. This is a multiset of realizations: for each realization, we know which arm $k$ it comes from, and realizations are all independent. Thus, denoting $N'_k$ the number of realizations for arm $k$, the data set writes
$$
\mathcal{X}' := \big\{ x'_{k,i} \;\big|\; k \in \{1, \ldots, K\},\ i \in \{1, \ldots, N'_k\} \big\}.
$$
We do not assume knowledge of the strategy $\pi'$ that determined the counts $(N'_k)_k$, so that the prior data may, for instance, originate from a bandit game. We do, however, require that $(N'_k)_k$ be \emph{independent} of the target distribution $\nu$ of the current game, making $(\nu', \pi')$ and $\nu$ independent; the only source of dependence between the prior observations and the current game comes through the pull counts $(N_k(T))_k$ chosen by the learner.
\\
\\
The prior counts $(N'_k)_k$ may depend on $T$. This is intentional: since our guarantees are asymptotic in $T$, a prior of fixed size would become negligible as $T \to +\infty$ and the setting would be trivial. Allowing $N'_k$ to scale with $T$ is also natural in practice, since the amount of prior data available before the game may be comparable to, or even exceed, the number of online rounds. Throughout the paper we keep the notation $N'_k$ for brevity, but the reader should keep in mind that $N'_k = N'_k(T)$.

\paragraph{Regret minimization} The goal is still to choose a strategy $\pi$ minimizing the regret with respect to the target environment,
$$
R^\pi_{T,\nu',\pi',\nu} = \sum_{k=1}^K \mathbb{E}_{\nu',\pi',\nu,\pi}[N_k(T)](\mu_1 - \mu_k),
$$
except that now the expectation also accounts for the randomness of the unknown source $\nu'$ and unknown prior strategy $\pi'$. We only know that for each arm $k$, $|\mu_k - \mu'_k| < L_k$. Everything with prime is about the transfer data/distribution.

\section{Lower bound}\label{sec:lowerbound}

In this section, we first review the classical instance-dependent regret lower bound for stochastic multi-armed bandits \citep{graves1997asymptotically}, before deriving our lower bound for the transfer setting introduced in \autoref{sec:subtransfer}.

\subsection{Lower bound in the classical setting}

Following \citet{graves1997asymptotically, garivier2019explore}, we define the set of consistent algorithms $\pi$ on the set of distributions $\mathcal{D}$ as
$$
\mathcal{C}^1_{\mathcal{D}} := \left\{ \pi \;\middle|\;
\begin{array}{l}
\forall \nu \in \mathcal{D}, \\
\forall \text{ suboptimal arm } k, \\
\forall \alpha > 0
\end{array}
,\ \mathbb{E}_{\nu,\pi}[N_k(T)] = o(T^\alpha) \right\}.
$$

\paragraph{Problem-dependent bounds} Then we have the following classical theorem (we refer the reader to e.g.\ \citealp{lattimore2020bandit} for a proof):

\begin{theorem}
For all algorithms $\pi \in \mathcal{C}^{1}_{\mathcal{D}}$, for all $\nu \in
\mathcal{D}$, for all suboptimal arms $k$,
$$
\liminf_{T \to \infty} \inf_{\substack{\tilde\nu_k \in \mathcal{D} \\ \tilde\mu_k > \mu_1}}
\frac{\mathbb{E}_{\nu,\pi}[N_k(T)]\, \mathrm{KL}(\nu_k, \tilde\nu_k)}{\ln T} \geq 1.
$$
\end{theorem}
Equivalently, introducing the notation
$$
\mathcal{K}_{inf}(\nu_k, \mu_1, \mathcal{D}) := \inf_{\substack{\tilde\nu_k \in \mathcal{D} \\ \tilde\mu_k > \mu_1}} \mathrm{KL}(\nu_k, \tilde\nu_k),
$$
one obtains the classical lower bound
\begin{equation}
    \mathbb{E}_{\nu,\pi}[N_k(T)] \geq \frac{\ln T}{\mathcal{K}_{inf}(\nu_k, \mu_1, \mathcal{D})} + o(\ln T),
\label{eq:borne-classique}
\end{equation}
and therefore, for every $\pi \in \mathcal{C}_{\mathcal{D}}$, for every $\nu \in \mathcal{D}^K$,
$$
R^\pi_{T,\nu} \geq \left( \sum_{k=2}^K \frac{\mu_1 - \mu_k}{\mathcal{K}_{inf}(\nu_k, \mu_1, \mathcal{D})} \right) \ln T + o(\ln T).
$$

\subsection{Instance-dependent lower bound in the transfer setting}\label{sec:lowerboundT}

We now extend this classical framework to the transfer with process-generated samples setting of \autoref{sec:subtransfer}. The extension is not immediate: the classical consistency class $\mathcal{C}_{\mathcal{D}}$ is defined with respect to a single, fixed environment $\nu$, whereas here the learner faces both an unknown target $\nu$ and an unknown, possibly data-dependent source $(\nu', \pi')$. We therefore need a notion of consistency that holds uniformly over all admissible sources.
\\
\\
For convenience, introduce the set
$$
\mathcal{P}([0,1])_L := \left\{ (\nu, \nu') \in \mathcal{P}([0,1])^K \times \mathcal{P}([0,1])^K \;:\; \forall k,\ |\mu_k - \mu'_k| \le L_k \right\}.
$$
We extend the notion of consistency to this transfer setting as follows. We define our class of consistent algorithms as
$$
\mathcal{C}_L := \left\{ \pi \;\middle|\;
\begin{array}{l}
\forall (\nu, \nu') \in \mathcal{P}([0,1])_L, \\
\forall \text{ prior strategy } \pi', \\
\forall \text{ suboptimal arm } k, \\
\forall \alpha > 0
\end{array}
,\ \mathbb{E}_{\nu',\pi',\nu,\pi}[N_k(T)] = o(T^\alpha) \right\}.
$$
From this notion of consistency, we derive the following lower bound.
\begin{theorem}
\label{thm:lowerbound}
For every algorithm $\pi \in \mathcal{C}_L$, for any $(\nu, \nu') \in \mathcal{P}([0,1])_L$,
for any prior strategy $\pi'$ that comes from a bandit game played over
$T' = f(T)$ rounds, for any suboptimal arm $a$:
\begin{itemize}
\item[(i)] If $\mu'_a \le \mu_1 - L_a$ and $\limsup_{T \to +\infty} \mathbb{E}_{\nu',\pi'}[N'_a] / \ln T < +\infty$, then
$$
\liminf_{T \to +\infty} \frac{\mathbb{E}_{\nu',\pi',\nu,\pi}[N_a(T)]}{\ln T}
\geq \liminf_{T \to +\infty} \left( 1 - \frac{\mathbb{E}_{\nu',\pi'}[N'_a]}{\ln T} \mathcal{K}_{inf}(\nu'_a, \mu_1 - L_a) \right)_+ \Big/ \mathcal{K}_{inf}(\nu_a, \mu_1).
$$
\item[(ii)] If $\mu'_a > \mu_1 - L_a$, then
$$
\liminf_{T \to +\infty} \frac{\mathbb{E}_{\nu',\pi',\nu,\pi}[N_a(T)]}{\ln T} \geq \frac{1}{\mathcal{K}_{inf}(\nu_a, \mu_1)}.
$$
\end{itemize}
\end{theorem}
(see the proof in \autoref{sec:prooflowerbound}).
\\
Our lower bound applies only when the prior game is a classical regret-minimization bandit game, but we believe this could be extended to other prior game types (e.g.\ best-arm identification).

\paragraph{Necessary and sufficient condition to have an improvement. } We can see that it is necessary to have $\mu'_a \le \mu_1 - L_a$ to obtain an improvement compared to the classical lower bound. From the regret bound of our algorithm in \autoref{sec:regret}, the strict condition $\mu'_a < \mu_1 - L_a$ is sufficient to obtain an improvement. Only the degenerate boundary case $\mu'_a = \mu_1 - L_a$ remains unclear (see also Proposition \ref{prop:matching} in \autoref{sec:regret}, where the same boundary is the only case left open once a concentration condition on $N'_a$ is assumed).

\paragraph{Tightness of the lower bound.}
We show that the lower bound of Theorem \ref{thm:lowerbound} is not tight in general, and we conjecture that closing this gap requires a fundamentally different proof technique, and we keep this for future work. The change-of-measure argument underlying Theorem \ref{thm:lowerbound} (Lemma \ref{lem:jointure}) compares two adaptive bandit instances through a KL divergence that depends on the source sample size only through the expectation $\mathbb{E}_{\nu',\pi'}[N'_a]$. This is enough to correctly locate the phase transition: the threshold it identifies coincides with the sufficient condition of our upper bound (Theorem \ref{thm:regret1}), $\mu'_a < \mu_1 - L_a$. However, the leading constant it produces can be arbitrarily loose, as the following example shows.

Intuitively, we want our consistency class to encode that, without any prior information, $\pi$ should behave as a consistent algorithm in the classical setting. Formally, let $\varpi_{\pi}$ denote the restriction of $\pi$ to the case where no prior samples are received, i.e.
\begin{equation*}
    \varpi_{\pi} \ : \ \underbrace{\mathcal{A}_t}_{\substack{\text{history up} \\ \text{to turn }t}} \mapsto \pi\big( \underbrace{\varnothing}_{\substack{\text{no prior}\\ \text{sample}}}, \mathcal{A}_t\big).
\end{equation*}
We want a consistency class containing at least $\mathcal{C}_{L} \cap \left\{ \pi \ \middle| \ \varpi_{\pi} \in \mathcal{C}^{1} \right\}$. Take $\pi$ in this set, and consider $(\pi', \nu')$ such that $\mathcal{K}_{\inf}(\nu'_a, \mu_1 - L_a) > 0$, $N'_k = 0$ a.s.\ for all $k \neq a$, and
$$
N'_a =
\begin{cases}
T & \text{with probability } \varepsilon, \\[1ex]
0 & \text{with probability } 1 - \varepsilon.
\end{cases}
$$
Since $\mathbb{E}_{\pi',\nu'}[N'_a]/\ln(T)  \to \infty$, the expectation-based threshold of Theorem~\ref{thm:lowerbound} is vacuous on this instance and provides no useful lower bound. Yet our restricted consistency class still forces the classical rate on the event where the source turns out to be uninformative:
\begin{align*}
    \mathbb{E}_{\pi',\nu',\pi,\nu}[N_a(T)] &\geq \mathbb{P}\big(\forall k, \ N'_k = 0\big) \ \mathbb{E}_{\pi',\nu',\pi,\nu}[N_a(T) \mid \forall k, \ N'_k = 0] \\
    & =(1 - \varepsilon) \ \mathbb{E}_{\varpi_{\pi},\nu}[N_a(T)] \\
    & \geq (1 - \varepsilon) \frac{\ln T}{\mathcal{K}_{\inf}(\nu_a,\mu_1)} \neq o(\ln T).
\end{align*}
This exhibits an instance on which Theorem~\ref{thm:lowerbound} is not tight. Closing the gap in full generality likely requires a consistency notion and a change-of-measure argument built on the conditional law of $N_a(T)$ given $N'_a$, rather than on an unconditional expectation-based comparison. We show in \autoref{sec:regret} (Proposition \ref{prop:matching}) that the gap in fact vanishes under a mild concentration condition on $N'_a$, except at the boundary $\mu'_a = \mu_1 - L_a$; whether it persists without such regularity, and what happens exactly at this boundary, remain open.

\paragraph{Adaptability for every $L$. } A natural question is whether it is possible to design an algorithm that exploits the prior data without any knowledge of an upper bound $L$ on $|\mu_k - \mu'_k|$. The answer is no. Indeed, since by uniform goodness, we want logarithmic regret for every $L$ satisfying the discrepancy condition, we must have
$$
\pi \in \bigcap_{\substack{L = (a_1,\ldots,a_K) \\ \forall k,\ a_k > |\mu_k - \mu'_k|}} \mathcal{C}_L = \mathcal{C}_{(1,\ldots,1)}.
$$
Consequently, we always have $\mu'_a > \mu_1 - L_a$, and Theorem~\ref{thm:lowerbound} shows that no improvement over the classical case is then possible. 

\section{Algorithm KL-UCB}\label{sec:algo}

Having established a lower bound on the achievable regret in the transfer setting, we now turn to the design of an algorithm that exploits the prior data to approach this bound. We first recall the classical KL-UCB algorithm and its optimality guarantee in the setting without prior data. We then present an instantiate of this algorithm, KL-UCB-Transfer, which aggregates the online samples with the prior data through a single index.

\subsection{Algorithm \texttt{KL-UCB} in the classical setting}

The \texttt{KL-UCB} algorithm was originally proposed in \cite{pmlr-v19-maillard11a}, inspired from \cite{lai1987adaptive}. The core idea is to pull, at each time $t$, the arm whose index $U_a(t)$ (an upper confidence bound on the arm mean) is maximal. For completeness we recall a general definition of the index used in nonparametric/bounded settings.
\\
\\
Let $\hat\nu_a(t)$ denote the empirical distribution of the rewards observed from arm $a$ up to time $t$, and let $N_a(t)$ be the corresponding pull count. For a threshold $\delta_t$, the index, which is an upper confidence bound on $\mu_a$, is defined as
$$
U_a(t) := \sup \left\{ \tilde\mu \;\middle|\; \tilde\nu \in \mathcal{D},\ N_a(t)\, \mathrm{KL}(\hat\nu_a(t), \tilde\nu) \le \delta_t \right\},
$$
where a standard choice is $\delta_t = \ln t + c \ln \ln t$ for a small constant $c$ that may depend on $\mathcal{D}$. The arm with the largest $U_a(t)$ is selected.
\\
\\
In the case $\mathcal{D} = \mathcal{P}([0,1])$ we define
$$
U^{\mathrm{KL}}(N, \nu, \delta) := \max \{ \mu \mid N\, \mathcal{K}_{inf}(\nu, \mu) \le \delta \}.
$$
The quantity $\mathcal{K}_{inf}(\nu, \mu)$ is computable from $\hat\nu_N$, the empirical distribution of observations $(X_i)_{i \le N}$ in $[0,1]$, using its dual form (see \cite{Honda2010AnAO}):
$$
N \mathcal{K}_{inf}(\hat\nu_N, U) = \max_{\lambda \in [0, \frac{1}{1-U}]} \sum_{i=1}^N \ln(1 - \lambda(X_i - U)).
$$
With $\delta_t := \ln t + \ln \ln t$, the index of \texttt{KL-UCB} is defined for each arm $a$ at each time $t$ as
$$
U_a(t) := U^{\mathrm{KL}}(N_a(t), \hat\nu_a(t), \delta_t).
$$
\texttt{KL-UCB} and its variants have been shown to be asymptotically optimal (in the sense of the problem-dependent bound \autoref{eq:borne-classique}) in many settings: Bernoulli rewards \citep{pmlr-v19-garivier11a}, one-parameter exponential families and bounded distributions with finite support \citep{Cappe2013KL}, with further refinements for both problem-dependent and problem independent optimality for bounded distributions \citep{garivier2022kl}.

\subsection{Algorithm \texttt{KL-UCB-Transfer}}\label{sec:KLT}

The \texttt{KL-UCB-Transfer} index was proposed in \cite{prevost2025asymptoticallyoptimalproblemdependentbandit} for the Gaussian setting with deterministic prior sample sizes. Here we adapt the idea to the bounded nonparametric setting $\mathcal{D} = \mathcal{D}' = \mathcal{P}([0,1])$, with $N'_k$ generated by an unknown sequential strategy $\pi'$, making $N'_k$ random and dependent on the prior data.
\\
\\
Let $\hat\nu'_a$ denote the empirical distribution of the prior samples for arm $a$. If $N'_a = 0$, we adopt the convention that the term $N'_a \mathcal{K}_{inf}(\hat\nu'_a, \cdot)$ is equal to $0$, regardless of the (undefined) value of $\hat\nu'_a$. Using the problem-dependent divergence $\mathcal{K}_{inf}$ introduced above, we redefine, for the remainder of the paper, the index $U_a(t)$ for each arm $a$ at each time $t > K$ as
$$
U_a(t) := \sup \left\{ \mu \;\middle|\; N_a(t)\, \mathcal{K}_{inf}(\hat\nu_a(t), \mu) + N'_a\, \mathcal{K}_{inf}(\hat\nu'_a, \mu - L_a) \le \delta_t \right\},
$$
with $\delta_t := \ln t + \ln \ln t$. Intuitively, the index aggregates the information coming from the online samples (first term) and from the prior samples (second term), while shifting the prior mean by the tolerance $L_a$.
\\
\\
We give below the pseudocode for \texttt{KL-UCB-Transfer}:
\begin{algorithm}[H]
\caption{\texttt{KL-UCB-Transfer}}
\begin{algorithmic}[1]
\State Receive the number of arms $K$, prior observations $X'$, upper bound of the distance $L$
\For{$t=1,\dots,K$}
    \State Pull each arm once
\EndFor
\For{$t=K+1,\dots,T$}
    \For{$a = 1,\dots,K$}
        \State Compute $U_a(t)$
    \EndFor
    \State Pull arm $a_t := \arg \max_a U_a(t)$ (pull uniformly at random in $\arg \max$ if it is not a singleton) 
\EndFor
\end{algorithmic}
\end{algorithm}

\section{Regret upper bound}\label{sec:regret}

We obtain the following guarantee on the regret of \texttt{KL-UCB-Transfer}.

\begin{theorem}
\label{thm:regret1}
For all distributions $\nu, \nu' \in \mathcal{P}([0,1])$ such that for all $k$ we have $|\mu_k - \mu'_k| < L_k$, for every suboptimal arm $a$ and for every $T \ge 3$, the algorithm \texttt{KL-UCB-Transfer} satisfies:
\begin{itemize}
\item[(i)] If $\mu'_a < \mu_1 - L_a$,
\begin{equation*}
    \mathbb{E}_{\nu',\pi',\nu,\pi}\big[N_a(T)\big]
\leq
\frac{\mathbb{E}_{\nu',\pi'}\big[\big(\ln T - N'_a\,\mathcal{K}_{inf}(\nu'_a,\mu_1-L_a)\big)_+\big]}
{\mathcal{K}_{inf}(\nu_a,\mu_1)}
\;+\; h_1(T).
\end{equation*}
\item[(ii)] If $\mu'_a \geq \mu_1 - L_a$,
\begin{equation*}
    \mathbb{E}_{\nu',\pi',\nu,\pi}\big[N_a(T)\big]
\leq
\frac{\ln T}
{\mathcal{K}_{inf}(\nu_a,\mu_1)}
\;+\; h_2(T),
\end{equation*}
\end{itemize}
where $h_1(T) = O\big((\ln T)^{4/5} (\ln \ln T)^{1/5}\big)$ and $h_2(T) = O\big((\ln T)^{4/5} (\ln \ln T)^{1/5}\big)$.
\end{theorem}
It is enough to have $\mu'_a < \mu_1 - L_a$ to obtain an improvement compared to the classical case.

\paragraph{Comparison with classical \texttt{KL-UCB}} Since $(\ln T - N'_a\, \mathcal{K}_{inf}(\hat\nu'_a, \mu_1 - L_a))_+ \le \ln T$, the first bound (i) is always at most the second (ii), so \texttt{KL-UCB-Transfer} never performs worse, at first order in $T$, than the classical rate $\ln T / \mathcal{K}_{inf}(\nu_a, \mu_1)$ of \texttt{KL-UCB} run without any prior information.

\paragraph{Comparison with the lower bound.} As discussed in \autoref{sec:lowerboundT} (paragraph ``Tightness of the lower bound''), Theorem \ref{thm:lowerbound} does not match the upper bound of Theorem \ref{thm:regret1} in general. We show below that this gap closes under a mild concentration condition on the prior sample size $N'_a$, except possibly at the boundary $\mu'_a = \mu_1 - L_a$.

\begin{proposition}\label{prop:matching}
If $\mathbb{E}_{\pi',\nu'} \left[ \ \big|  N'_a - \mathbb{E}_{\pi',\nu'}[N'_a]  \big| \ \right] = o(\ln T)$, then the lower bound of Theorem \ref{thm:lowerbound} matches the upper bound of Theorem \ref{thm:regret1}, except possibly when $\mu'_a = \mu_1 - L_a$.
\end{proposition}
The proof is given in \autoref{sec:proofmatching}. Combined with the discussion following Theorem \ref{thm:lowerbound}, this isolates the single case left open in the well-specified analysis: the boundary $\mu'_a = \mu_1 - L_a$, at which neither the improvement condition of \autoref{sec:lowerboundT} nor Proposition \ref{prop:matching} determines the exact rate.


\section{Robustness to misspecification of $L$}\label{sec:robust}

We now study how the regret guarantee depends on these assumptions, and to what extent they can be relaxed. We show that the algorithm remains meaningful even without any discrepancy assumption, and that the quality of the regret bound degrades in a precise and structured way.
\\
\\
In particular, we show that all regret guarantees can be expressed through a single comparison level $\mu^\dagger$, which allows us to characterize exactly when logarithmic regret is preserved and how it degrades otherwise.

\begin{theorem}\label{thm:general}
For all distributions $\nu,\nu'\in\mathcal{P}([0,1])$, for every suboptimal arm $a$ and for every $T\geq 3$, the algorithm \texttt{KL-UCB-Transfer} (run with prior data and discrepancy bound $L$) satisfies the following.
\\
Let $\mu^\dagger\in(\mu_a,\mu_1]$ be any real number such that $\mu^\dagger < \mu'_1 + L_1$. Then
\begin{equation*}
\mathbb{E}_{\nu',\pi',\nu,\pi}[N_a(T)] \leq \frac{\ln T}{\mathcal{K}_{inf}(\nu_a,\mu^\dagger)} + h_{\nu,\nu',\mu^\dagger,L}(T).
\end{equation*}
If moreover $\mu'_a + L_a < \mu^\dagger$, then
\begin{equation*}
\mathbb{E}_{\nu',\pi',\nu,\pi}[N_a(T)] \leq \frac{\mathbb{E}_{\nu',\pi'}\big[\big(\ln T - N'_a\,\mathcal{K}_{inf}(\nu'_a,\mu^\dagger-L_a)\big)_+\big]}{\mathcal{K}_{inf}(\nu_a,\mu^\dagger)} + g_{\nu,\nu',\mu^\dagger,L}(T),
\end{equation*}
where $h_{\nu,\nu',\mu^\dagger,L}(T) = \mathcal{O}\big((\ln T)^{4/5}(\ln\ln T)^{1/5}\big)$ and $g_{\nu,\nu',\mu^\dagger,L}(T) = \mathcal{O}\big((\ln T)^{4/5}(\ln\ln T)^{1/5}\big)$.
\end{theorem}
An explicit expression for both $h_{\nu,\nu',\mu^\dagger,L}$ and $g_{\nu,\nu',\mu^\dagger,L}$ is given in the proof in section \ref{sec:proofgeneral}.
\\
\\
We remind that $\mu^{\dagger}$ is not an input of the algorithm. The algorithm only uses the prior data $(N'_a,\hat{\nu}'_a)_a$ and the discrepancy bounds $(L_a)_a$.

\begin{figure}[h]
\centering

\begin{tikzpicture}[x=8.5cm, y=1.5cm, >=stealth]

\draw[->, thick] (-0.05,0) -- (1.2,0) node[right] {$\mu'_1+L_1$};

\draw[thick] (0.3,0.1) -- (0.3,-0.1) node[below=4pt] {$\mu_a$};
\draw[thick] (0.8,0.1) -- (0.8,-0.1) node[below=4pt] {$\mu_1$};

\draw[red!80!black, ultra thick] (-0.05,0) -- (0.3,0);
\draw[orange!90!black, ultra thick] (0.3,0) -- (0.8,0);
\draw[green!50!black, ultra thick] (0.8,0) -- (1.15,0);


\node[
  anchor=south,
  align=center,
  font=\small,
  inner sep=2pt
] at (0.125,0.32) {
  \color{red!80!black}\textbf{Severely misspecified}\\
  \small No guarantee
};

\node[
  anchor=south,
  align=center,
  font=\small,
  inner sep=2pt
] at (0.55,0.32) {
  \color{orange!90!black}\textbf{Mildly misspecified}\\
  \small Logarithmic regret\\
  \small Degraded constant
};

\node[
  anchor=south,
  align=center,
  font=\small,
  inner sep=2pt
] at (0.975,0.32) {
  \color{green!50!black}\textbf{Well-specified}\\
  \small Logarithmic regret\\
  \small Best constant\\
  \small (Recovers Theorem~\ref{thm:regret1})
};

\end{tikzpicture}

\caption{
Regret regimes of \texttt{KL-UCB-Transfer} as a function of the effective transfer level $\mu'_1 + L_1$ from theorem \ref{thm:general}.
}
\label{fig:regimes_transfer}

\end{figure}
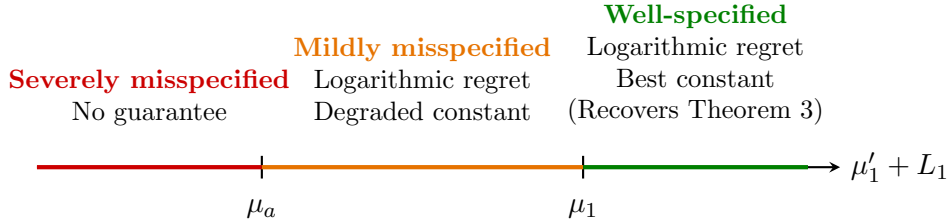

\paragraph{Three regimes.}
In the well-specified regime, where $|\mu_a - \mu'_a| < L_a$ holds for all arms $a$, we in particular have $\mu_1 < \mu'_1 + L_1$, and choosing $\mu^{\dagger} := \mu_1$ in Theorem~\ref{thm:general} recovers Theorem~\ref{thm:regret1}. When this condition is violated, i.e., $\mu'_1 + L_1 < \mu_1$, logarithmic regret may still hold as long as $\mu'_1 + L_1 > \mu_a$: any $\mu^{\dagger} \in (\mu_a, \mu'_1 + L_1)$ yields a mildly misspecified regime with a degraded constant. If instead $\mu'_1 + L_1 \leq \mu_a$, no admissible choice of $\mu^{\dagger}$ exists and our theorem gives no guarantee, a severely misspecified regime. Figure~\ref{fig:regimes_transfer} summarizes the three regimes.

\paragraph{Asymmetry of discrepancy requirements.}
The role of the discrepancy bounds is \emph{asymmetric}. The condition on the optimal arm, through $L_1$, is critical for ensuring the existence of a logarithmic regret regime. By contrast, the bounds $(L_a)_{a \neq 1}$ only affect the leading constant when tightened, and are not required for logarithmic regret. Since the optimal arm is unknown, a uniform condition across all arms is imposed in practice.

\paragraph{One-sided discrepancy conditions.}
The analysis only relies on the one-sided quantities $\mu'_a+L_a$, rather than on the symmetric discrepancies $|\mu_a-\mu'_a|$. In practice, since the optimal arm is unknown, it is sufficient to choose $(L_a)_a$ so that
\[
\mu_a<\mu'_a+L_a,\qquad\forall a,
\]
which guarantees in particular that the optimal arm satisfies the condition required by Theorem~\ref{thm:general}. The lower inequalities $\mu'_a-L_a<\mu_a $ play no role in the regret analysis.

\paragraph{Favorable overestimation regime.}
If $\mu'_1 > \mu_1$, one may set $L_1 = 0$ while preserving $\mu'_1 + L_1 > \mu_1$. In particular, if $\mu'_a > \mu_a$ holds for all arms, then setting $L_a = 0$ for all $a$ is valid and yields the optimal logarithmic regime.

\medskip

A typical example is a systematic shift in the environment, where the source distribution corresponds to a more favorable setting (e.g., a richer soil in an agricultural harvesting task), while the target environment is less favorable. In such cases, all arms are uniformly better in the source than in the target, leading to $\mu'_a > \mu_a$ for all $a$, and hence we can take $L_a = 0$ for all arm $a$.

\paragraph{Small number of prior data. } If the number of prior samples is bounded by a quantity independent from $T$ (i.e $\forall k$, $N'_k \leq B$ a.s.) then even in the Severely misspecified regime we recover the classical rate of \texttt{KL-UCB}. We do not provide a proof of it but it can be deduced easily from the proof of Theorem \ref{thm:general} by using Lemma \ref{lem:supKinf}.

\section{Extension to Multiple Independent Sources}
\label{sec:multisource}

The analysis developed for Theorem \ref{thm:general} extends naturally to several independent prior sources. This extension further illustrates the flexibility of the effective transfer level $\mu^\dagger$.
\\
\\
We consider $M-1$ independent source games, played on distributions $\nu^{(2)}, \ldots, \nu^{(M)} \in \mathcal{P}([0,1])$ by strategies $\pi^{(2)}, \ldots, \pi^{(M)}$. Each source game is associated with its own discrepancy bound $L^{(i)} = (L_a^{(i)})_a$. We denote the target game by $(\nu^{(1)}, \pi^{(1)}) = (\nu, \pi)$, and we write $(\nu) := (\nu^{(2)}, \ldots, \nu^{(M)})$ and $(\pi) := (\pi^{(2)}, \ldots, \pi^{(M)})$ for brevity. As in the single-source setting, each source game is assumed independent of the target $\nu$. We further assume the $M-1$ source games to be mutually independent of one another. \texttt{KL-UCB-Transfer} is run with the index generalized in the natural way, aggregating all $M-1$ prior batches alongside the online samples.

\begin{theorem}
\label{thm:multisource}
For all distributions $\nu^{(1)}, \ldots, \nu^{(M)} \in \mathcal{P}([0,1])$, for every suboptimal arm $a$ and for every $T \geq 3$, the algorithm \texttt{KL-UCB-Transfer} run with the $M-1$ independent prior games and bounds $(L^{(2)}, \ldots, L^{(M)})$ satisfies the following.
\\
Let $\mu^\dagger \in (\mu_a, \mu_1]$ be any real number such that
\[
\mu^\dagger < \min_{i \in \{2,\ldots,M\}} \mu_1^{(i)} + L_1^{(i)}.
\]
Denote by
\[
\mathcal{S}_a^\dagger := \left\{ i \in \{2,\ldots,M\} \; \middle| \; \mu_a^{(i)} + L_a^{(i)} < \mu^\dagger \right\},
\]
the set of sources that are informative for arm $a$ at level $\mu^\dagger$. Then
\[
\mathbb{E}_{(\nu),(\pi),\nu,\pi}[N_a(T)] \leq \frac{\mathbb{E}_{(\nu),(\pi)}\left[\left(\ln T - \sum_{i \in \mathcal{S}_a^\dagger} N_a^{(i)} \mathcal{K}_{inf}\big(\nu_a^{(i)}, \mu^\dagger - L_a^{(i)}\big)\right)_+\right]}{\mathcal{K}_{inf}(\nu_a, \mu^\dagger)} + g_{(\nu),\mu^\dagger,(L)}(T),
\]
where $g_{(\nu),\mu^\dagger,(L)}(T) = O\big((\ln T)^{4/5} (\ln \ln T)^{1/5}\big)$.
\end{theorem}
The proof follows the same structure as that of Theorem \ref{thm:general} and is given in \autoref{sec:proofmultisource}.

\paragraph{Recovering Theorem \ref{thm:general}.} Taking $M := 2$ recovers exactly Theorem~\ref{thm:general}. In that case $\mathcal{S}_a^\dagger \in \{\emptyset, \{2\}\}$ depending on whether the single source is informative for arm $a$ at level $\mu^\dagger$.

\section{Numerical Experiments}\label{sec:simu}

We evaluate \texttt{KL-UCB-Transfer} against classical \texttt{KL-UCB} (no transfer). All rewards are Beta distributions with variance fixed to $\sigma^2 = 0.05$ across all experiments (only the mean varies). Regret curves are averaged over $M = 100$ independent Monte Carlo repetitions over an horizon $T = 10^5$, with shaded $\pm 3\frac{\hat\sigma_M}{\sqrt{M}}$ bands, and reported in log scale for the abscissa. All the experiments conducted in this section are available on \url{https://gitlab.inria.fr/waris.radji/kl-ucb-transfer-companion}.

\subsection{The effectiveness of transfer learning (Theorem~\ref{thm:regret1})}
\label{subsec:exp-effectiveness}

\begin{figure}[h]
    \centering
    \includegraphics[width=0.48\textwidth]{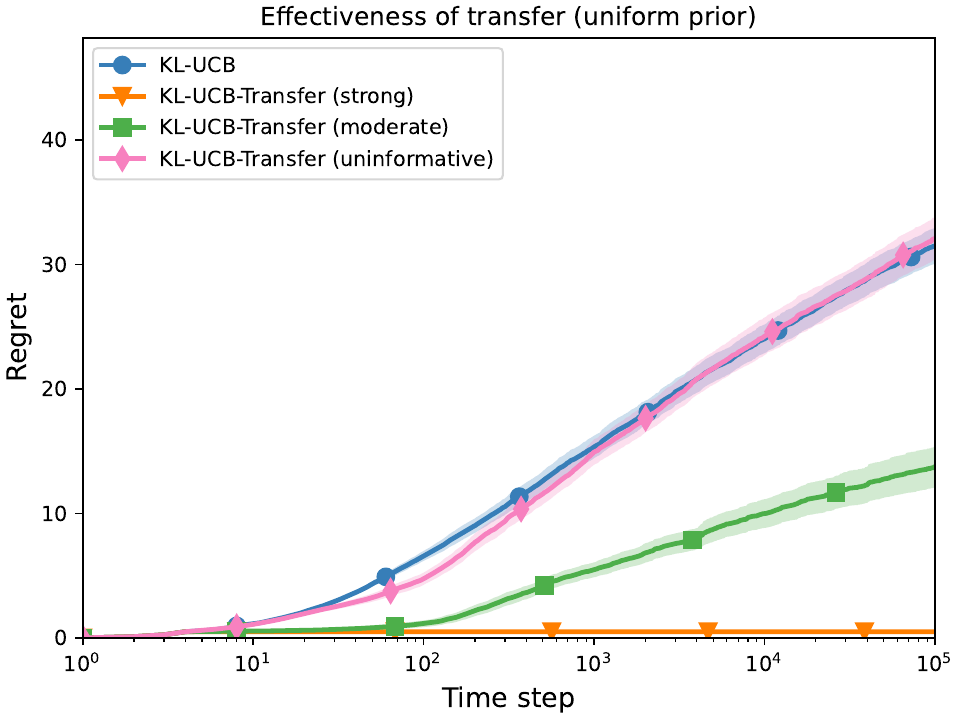}
    \includegraphics[width=0.48\textwidth]{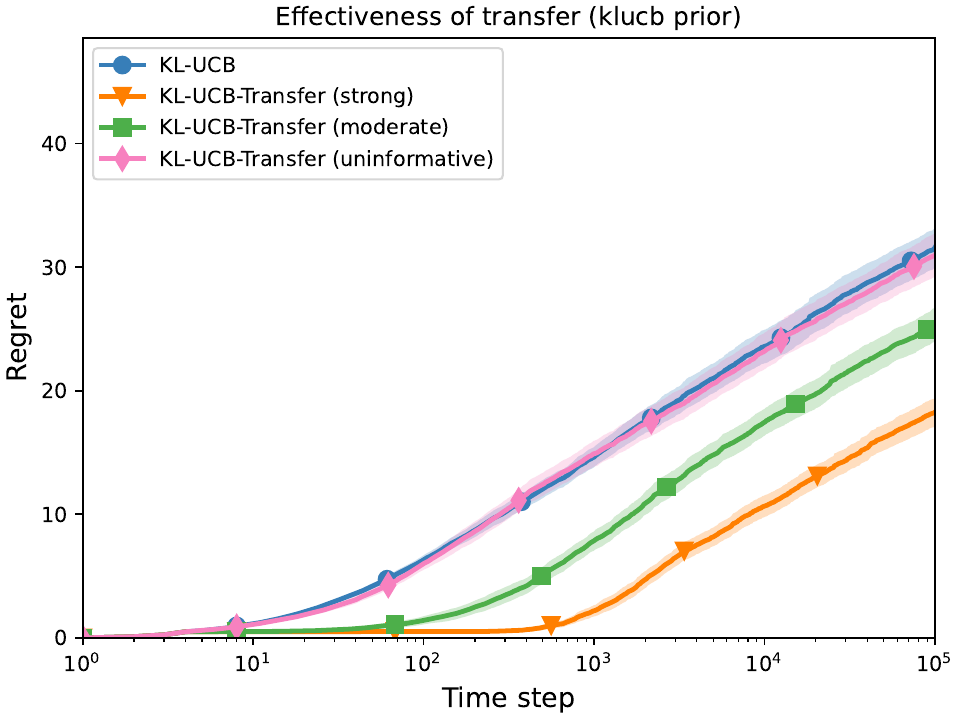}
    \caption{Cumulative regret of \texttt{KL-UCB-Transfer} vs.\ \texttt{KL-UCB}, well-specified regime. Left: $N'_a = 500$ for all $a$. Right: prior generated by a \texttt{KL-UCB} run on $T' = 2000$ rounds.}
    \label{fig:expA}
\end{figure}

We take $K=4$ arms with $\mu = (0.6, 0.5, 0.45, 0.35)$ (arm 1 optimal). The choice of $\mu'$ is summarized in Table~\ref{tab:exp1}. Recall that prior samples are informative for a suboptimal arm $a$ if $\mu_1 > \mu'_a + L_a$.
\\
\\
On the left panel of Figure~\ref{fig:expA} we take the non-adaptive $N'_a = 500$ for each arm $a$. On the right panel, we instead play \texttt{KL-UCB} on $\mu'$ for $T' = 2000$ rounds to generate the prior samples $(N'_a, \hat\nu'_a)_a$ themselves, illustrating that the guarantee also holds when $N'_a$ is random and data-dependent, as allowed by our setting.

\begin{table}[h]
\centering
\begin{tabular}{lcccc}
\toprule
Source & $\mu'$ & $L$ & Informative for & Role \\
\midrule
$1$ & $(0.58, 0.47, 0.42, 0.32)$ & $0.05$ (all arms) & all $a$ & strong source \\ $2$ & $(0.58, 0.56, 0.44, 0.34)$ & $0.07$ (all arms) & $a \in\{3,4\}$ only & moderate source \\ $3$ & $(0.5, 0.63, 0.5, 0.48)$ & $0.15$ (all arms) & none & uninformative source \\
\bottomrule
\end{tabular}
\caption{Parameters for the source prior}
\label{tab:exp1}
\end{table}

\noindent
\textbf{Interpretation of Figure \ref{fig:expA}.} On the left panel, the strong source is informative for all suboptimal arms. The $500$ prior samples per arm are enough to rule out these arms over the whole horizon, so the regret remains almost zero. The moderate source is informative only for arms 3 and 4. Arm 2 must therefore still be explored online, and its contribution to the regret is close to its contribution under \texttt{KL-UCB}. The uninformative source provides no benefit, so its curve matches that of \texttt{KL-UCB} at first order.
\\
\\
On the right panel, the gains are smaller even though the sources are the same. A regret-minimizing policy samples an arm less often when it performs poorly in the source. These are also the arms whose samples would provide the most information. The prior therefore saves less exploration, and the regret starts increasing earlier than on the left panel.

\subsection{The three regimes of Theorem~\ref{thm:general} (misspecification of $L$)}
\label{subsec:exp-misspecification}

We fix $\mu = (0.6, 0.5, 0.45, 0.35)$, $L_a = 0.05$ for all $a$, and $\mu'_{2,3,4} = (0.51, 0.47, 0.37)$, well-specified throughout (as in Section~\ref{subsec:exp-effectiveness}). Prior sample sizes are fixed to $N'_a = 500$ for all arms and both panels reuse the two prior generation protocols of Figure~\ref{fig:expA} ($N'_a=500$ non-adaptive, and \texttt{KL-UCB} with $T'=2000$). We track the regret contribution of arm $a=2$, the second-best arm ($\mu_2 = 0.5$), since it is the suboptimal arm closest to $\mu_1$ and hence the one responsible for the transition between regimes. We vary $\mu'_1$ only:

\begin{table}[h]
\centering

\begin{tabular}{lccc}
\toprule
Regime & $\mu'_1$ & $\mu'_1 + L_1$ & Guarantee (arm $a=2$, $\mu_2=0.5$) \\
\midrule
Well-specified & $0.58$ & $0.63$ & 
log-regret, best constant ($0.63 > \mu_1 = 0.6$) \\ Mildly misspecified & $0.49$ & $0.54$ &\hspace{-1em}  degraded constant ($0.54 \in (\mu_2, \mu_1) = (0.5, 0.6)$) \\ Severely misspecified & $0.40$ & $0.45$ & none ($0.45 \leq \mu_2 = 0.5$) \\
\bottomrule
\end{tabular}
\caption{Parameters for the three regimes of Theorem~\ref{thm:general}}
\label{tab:expC}

\end{table}

\begin{figure}[h]
    \centering
    \includegraphics[width=0.48\textwidth]{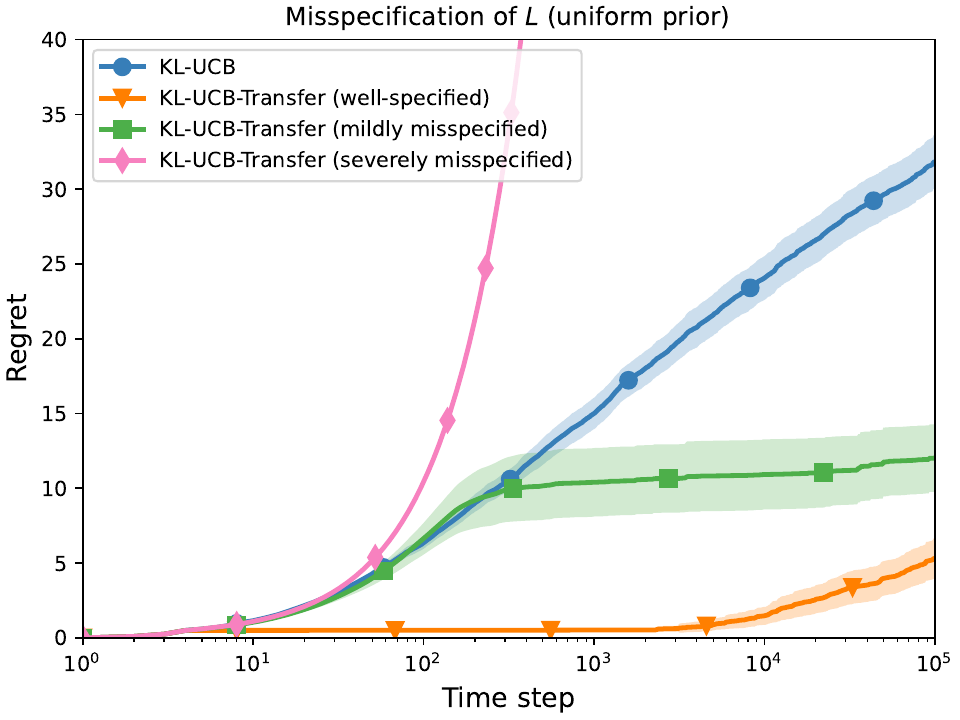}
    \includegraphics[width=0.48\textwidth]{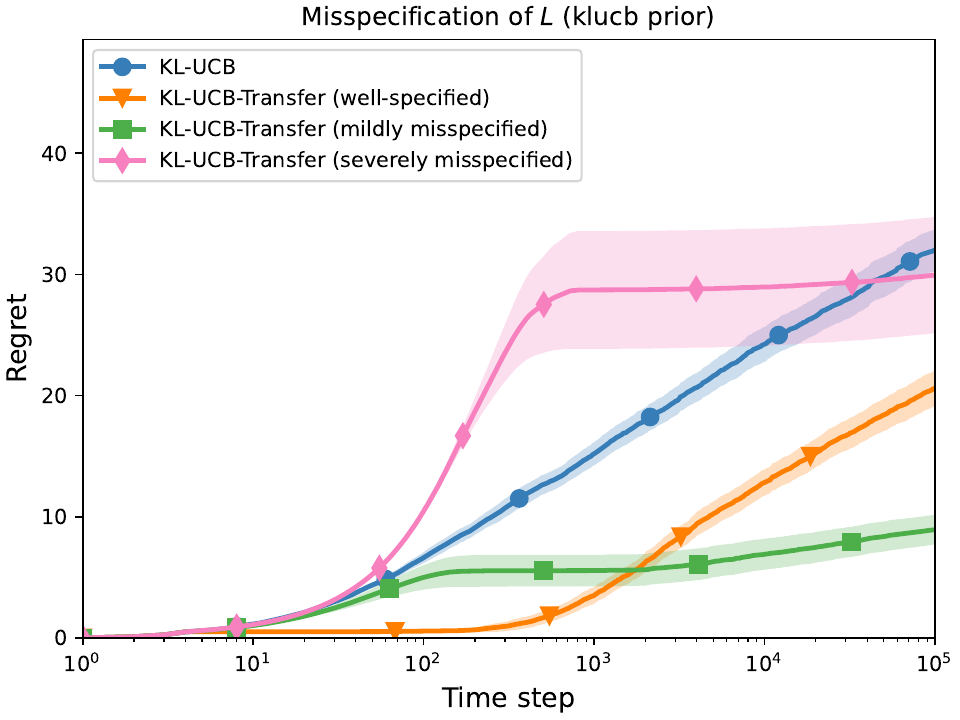}
    \caption{Cumulative regret of \texttt{KL-UCB Transfer} vs.\ \texttt{KL-UCB} on the 3 regimes of Theorem~\ref{thm:general}. Left: $N'_a = 500$ for all $a$. Right: prior from a \texttt{KL-UCB} run on $T' = 2000$ rounds.}
    \label{fig:exp-misspec}
\end{figure}
\noindent
\textbf{Interpretation of Figure \ref{fig:exp-misspec}.} On the left panel, the well-specified curve behaves as in Figure \ref{fig:expA}. In the mildly misspecified case, the prior underestimates arm 1 but still ranks it above arm 2 because $\mu'_1+L_1>\mu_2$. The algorithm can therefore still rule out arm 2 and the regret remains logarithmic. It even remains below that of \texttt{KL-UCB} because the prior samples from arm 2 help to rule it out. In the severely misspecified case, the prior incorrectly ranks arm 2 above arm 1 because $\mu'_1+L_1\le\mu_2$. The algorithm then pulls arm 2 on almost every round, and the regret grows linearly.
\\
\\
On the right panel, the prior is generated by a regret-minimizing policy. In the severe case, arm 1 is clearly suboptimal in the source and receives few prior samples. The misleading information therefore has a limited effect and the regret remains reasonable. Changing $\mu'_1$ also changes the source allocation. When $\mu'_1=0.49$, arm 2 is the best source arm and receives many more samples than when $\mu'_1=0.58$. This explains why the mildly misspecified curve lies below the well-specified one.

\subsection{Hard well-specified case}
\label{subsec:exp-struggle}

We fix $\mu = (0.6, 0.58)$ and add prior samples only to the optimal arm $1$. We set $\mu'_1 = 0.5$ and $L_1 = 0.11$, so that $\mu'_1 + L_1 = 0.61 \sim 0.6 = \mu_1$: the well-specified regime holds, but only barely. Since the prior term costs nothing below $\mu = 0.61$, the index $U_1(t)$ stays pinned near $0.61$ regardless of $N'_1$, causing the algorithm to over-explore arm $2$ until its own index falls below this level.

\begin{figure}[h]
    \centering
    \includegraphics[width=0.46\textwidth]{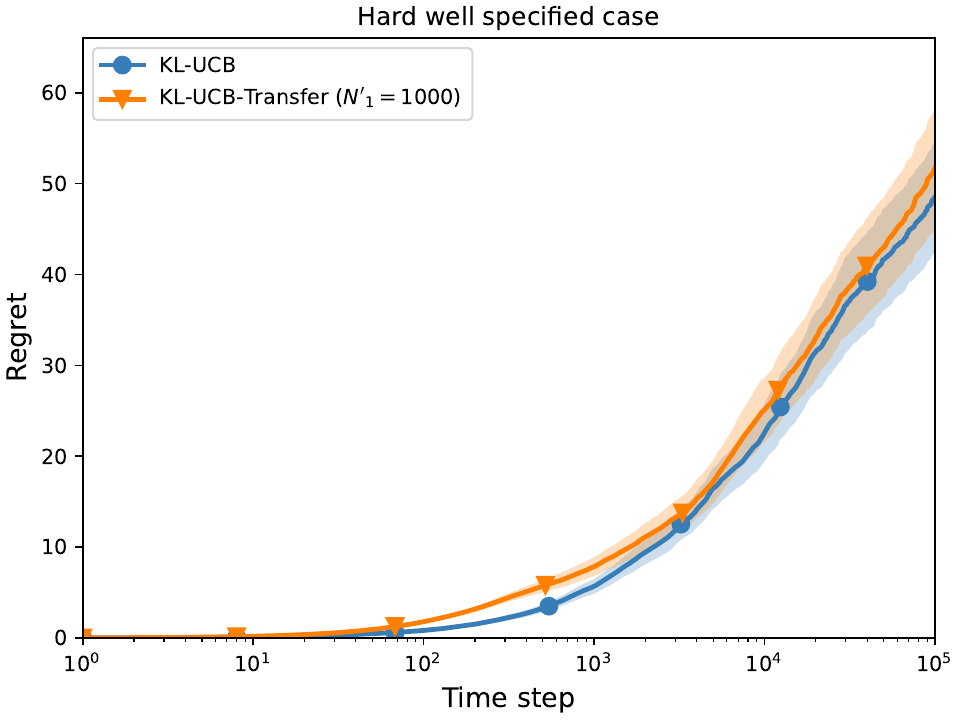}
    
    \caption{Cumulative regret of \texttt{KL UCB-Transfer} vs.\ \texttt{KL-UCB}, prior samples only on arm $1$, $N'_1 = 1000$.}
\vspace{-1em}
    \label{fig:exp-struggle}
\end{figure}
\noindent
\textbf{Interpretation of Figure \ref{fig:exp-struggle}.} We illustrate this with Figure~\ref{fig:exp-struggle}.
The well-specification margin
$\eta := \mu'_1+L_1-\mu_1 = 0.01$ is small. Since the prior term is nonnegative, transfer can only decrease the indices. In particular, $U_1(t)$ is capped just above $\hat\mu'_1+L_1 \approx 0.61$, while the \texttt{KL-UCB} index of arm~1 is much larger in early rounds. Arm 2 is therefore explored earlier. 
\\
\\
The analysis captures both effects. The second-order term in Theorem \ref{thm:general} contains $(1-e^{-2\eta^2})^{-3} \approx (2\eta^2)^{-3}$ (Lemmas \ref{lem:terme3} and \ref{lem:inter}), which diverges as $\eta\to0$. By contrast, Lemma \ref{lem:inter} bounds the event $\hat\mu'_1+L_1 < \mu_1-\varepsilon$ independently of $N'_1$, suggesting that this effect does not increase with the amount of prior data. The first-order term is unchanged, so the gap with \texttt{KL-UCB} remains $o(\ln T)$, although the difference persists throughout the horizon. Designing an algorithm that uses $T$ to better handle short horizons, as in \cite{sentenac2025balancingoptimismpessimismofflinetoonline}, is left for future work.

\section{Conclusion}\label{sec:conclusion}
In this paper we characterize the problem of transfer of a random number of prior data in Bandit problem with bounded distributions.

Several questions remain open. First, our lower bound (Theorem \ref{thm:lowerbound}) does not match the upper bound of Theorem \ref{thm:regret1} in full generality: the change-of-measure argument underlying it only sees the source sample size $N'_a$ through its expectation, and \autoref{sec:lowerboundT} exhibits an instance on which this makes the bound loose. We show, however, that this gap vanishes under a mild concentration condition on $N'_a$ (Proposition \ref{prop:matching}), namely $\mathbb{E}_{\pi',\nu'} \left[ \ \big|  N'_a - \mathbb{E}_{\pi',\nu'}[N'_a]  \big| \ \right] = o(\ln T)$, except at the boundary $\mu'_a = \mu_1 - L_a$. Whether the gap persists for source policies violating this regularity, and what happens exactly at this boundary case, remain open, and would likely require a change-of-measure argument built on the conditional law of $N_a(T)$ given $N'_a$ rather than on an unconditional, expectation-based comparison.

Second, we assumed the discrepancy bounds $(L_k)_k$ to be known exactly. A natural relaxation would be to only require that the condition $|\mu_k - \mu'_k| < L_k$ hold with some known probability, allowing for a degree of uncertainty in the discrepancy specification itself, and to compare the resulting guarantees to those obtained here. Pushing this further, one could place a prior distribution directly over $L$ rather than treating it as a fixed, adversarially chosen constant, which would connect our framework to the Bayesian approaches discussed in \autoref{sec:related} while retaining a discrepancy-based rather than distributional link between source and target.

Third, our results are stated for a discrepancy measured through the difference of means. It would be of interest to consider other notions of distance between $\nu_k$ and $\nu'_k$, for example based on higher moments, or on an information theoretic divergence such as the KL divergence itself. The construction of the algorithm remains unclear for now, and it would possibly require new concentration inequalities.

Finally, our analysis is asymptotic in $T$. A non asymptotic study, in the spirit of \cite{sentenac2025balancingoptimismpessimismofflinetoonline}, would be valuable both for practical deployment and for a finer understanding of the regimes identified in Theorem \ref{thm:general}, particularly when the horizon $T$ is known in advance.


\acks{The authors gratefully acknowledge support from the French Ministry of Higher Education and Research, the Hauts-de-France Region, Inria and the MEL. They also thank the Scool research group for its excellent working environment, and Inria-IISc Associate Team “FOSSIL” for supporting the project.  This work was supported by the PEPR FOUNDRY project (ANR-23-PEIA-0003) and the ANR STRESS (ANR-25-CE23-6825). 
}

\vskip 0.2in
\bibliography{sample}
\newpage

\appendix

\section{Useful lemmas}

In the section we present lemmas from other paper that we will use in our proof.
\\
The next two lemmas are some useful concentration of the $\mathcal{K}_{inf}$ for distributions in $[0,1]$. 
\begin{lemma}[From \cite{garivier2022kl}, Proposition 14]\label{lem:concentre1}
For distributions in $[0,1]$, 
\begin{equation*}
    \mathbb{P}_{\nu}\left( \mathcal{K}_{inf}(\hat \nu_s,\mathbb{E}[\nu]) >u  \right) \leq e(2s+1)e^{-su}.
\end{equation*}
\end{lemma}
\begin{lemma}[From \cite{garivier2022kl}, Proposition 16]\label{lem:concentre2}
With the same notations and assumptions as in the previous proposition,
consider a real number $\mu \in (\mathbb{E}(\nu),1)$ and define
\[
\gamma=\frac{1}{\sqrt{1-\mu}}\left(16e^{-2}+\ln^2\left(\frac{1}{1-\mu}\right)\right) \in \left[2 , \frac{4}{(1 - \mu)^{5/2}} \right].\] 

Then for all $y>0$,
\[
\mathbb{P}\!\left(
\mathcal{K}_{inf}(\hat \nu_n,\mu)\leq \mathcal{K}_{inf}(\nu,\mu) - y\right)
\leq
\left\{\begin{array}{ll}
\exp(-n\gamma/8)\le \exp(-n/4),& \text{if } y\geq \gamma/2,
\\[1ex]\exp\left(-\dfrac{ny^2}{2\gamma}\right),& \text{if } y < \gamma/2.
\end{array}
\right.
\]
\end{lemma}
Next lemma gives a bound of the $\mathcal{K}_{inf}$ for distributions bounded in $[0,1]$. 
\begin{lemma}[From \cite{Cappe2013KL} in the proof of lemma 8 (HAL version)]\label{lem:supKinf}
Let $\nu\in [0,1]$ and $\mu \in [0,1)$, we have
\[
\mathcal{K}_{inf}(\nu,\mu) \leq \ln \left(\frac{1}{1-\mu}\right).
\]
\end{lemma}
The last lemma gives nice properties on the $\mathcal{K}_{inf}$,
\begin{lemma}[From \cite{garivier2022kl}, Lemma 11]\label{lem:Kinfepsilon}
For all $\nu\in\mathcal{F}$, all $\mu\in(0,1)$, and all $0<\varepsilon<\mu$,
\[
\mathcal{K}_{inf}(\nu,\mu)\le
\mathcal{K}_{inf}(\nu,\mu-\varepsilon)+\frac{\varepsilon}{1-\mu},
\]
and, under the additional condition that $\mathbb{E}(\nu)<\mu-\varepsilon$,
\[
\mathcal{K}_{inf}(\nu,\mu)\ge
\mathcal{K}_{inf}(\nu,\mu-\varepsilon)+2 \varepsilon^2.
\]
\end{lemma}
We will often use the following fundamental inequality in bandit for every $s$
\begin{equation}\label{eq:fondaineq}
    \sum_{t=1}^T \mathbf{1}\left( A_t = a, \ N_a(t) = s \right) \leq 1.
\end{equation}

\section{Proof of Theorem \ref{thm:general}}\label{sec:proofgeneral}

To prove this theorem we will use the following decomposition:
\begin{lemma}\label{lem:decomposition}
For all $\nu,\nu'$, for all $\mu^{\dagger}$ such that
\begin{equation*}
    \mu^{\dagger} > \mu_a, \quad \mu^{\dagger} \leq \mu_1, \quad \mu^{\dagger} \leq \mu'_1 + L_1,
\end{equation*}
for every $\varepsilon$ such that
\begin{equation*}
    \varepsilon > 0, \quad \varepsilon < \mu^{\dagger}, \quad \varepsilon \leq \frac{(1 - \mu^{\dagger})\mathcal{K}_{inf}(\nu_a,\mu^{\dagger})}{4},
\end{equation*}
for all $T > 3$, with \texttt{KL-UCB-Transfer} using $L$ and $\delta_t := \ln t + \ln \ln t$ we have for every suboptimal arm $a$ that
    \begin{align*}
       N_a(T) & \leq \frac{(\ln T - N'_a\mathcal{K}_{inf}(\hat \nu'_a,\mu^{\dagger} - \varepsilon - L_a))_+}{\mathcal{K}_{inf}(\nu_a,\mu^{\dagger}) } + \frac{\ln \ln T}{\mathcal{K}_{inf}(\nu_a,\mu^{\dagger})} +  \frac{4 \varepsilon (\ln T + \ln \ln T)}{\mathcal{K}_{inf}(\nu_a,\mu^{\dagger})^2(1 - \mu^{\dagger})} \\
    & + \sum_{s = 1}^{+ \infty}\mathbf{1}\left(\mathcal{K}_{inf}(\hat \nu_{a,s},\mu^{\dagger}) \leq \mathcal{K}_{inf}(\nu_a,\mu^{\dagger}) - \frac{ \varepsilon}{1 - \mu^{\dagger}} \right) \\
    & + \sum_{t=K+1}^T \sum_{s = 1}^{t}  \mathbf{1}\left( \mathcal{K}_{inf}(\hat \nu_{1,s},\mu_1)  > \frac{\delta_t -  N'_1  \mathcal{K}_{inf}(\hat \nu_1',\mu^{\dagger} - \varepsilon-L_1)}{s} + 2\varepsilon^2 \right) \\
    & + \sum_{t=K+1}^T\sum_{s = 1}^{+ \infty}\mathbf{1}\left( \mathcal{K}_{inf}(\hat \nu_{1,s}',\mu'_1) \geq \frac{\delta_t}{s} + 2\varepsilon^2 \right) + K.
    \end{align*}
\end{lemma}
(see proof of Lemma \ref{lem:decomposition} in \autoref{sec:proofdecomposition}). \\
From this decomposition we will use the following Lemmas to control each term:
\begin{lemma}\label{lem:terme1}
With the same assumption as the Lemma \ref{lem:decomposition}
    \begin{equation*}
        \mathbb{E}_{\nu',\pi',\nu,\pi}\left[\sum_{s = 1}^{+ \infty}\mathbf{1}\left(\mathcal{K}_{inf}(\hat \nu_{a,s},\mu^{\dagger}) \leq \mathcal{K}_{inf}(\nu_a,\mu^{\dagger}) - \frac{ \varepsilon}{1 - \mu^{\dagger}} \right) \right] \leq 10 + \frac{16}{\varepsilon^2(1 - \mu^{\dagger})^{1/2}}.
    \end{equation*}
\end{lemma}
(see proof in \autoref{sec:proofterme1}). \\
Lemma \ref{lem:terme1} is similar to what we do in the classical proof of \texttt{KL-UCB} \cite{Cappe2013KL} by using the concentration Lemma \ref{lem:concentre2}, no contribution is made here

\begin{lemma}\label{lem:terme2}
With the same assumption as the Lemma \ref{lem:decomposition} with also $\varepsilon \leq 1$,
    \begin{align*}
      \mathbb{E}_{\nu',\pi',\nu,\pi}\left[\sum_{t=K+1}^T\sum_{s = 1}^{+ \infty}\mathbf{1}\left( \mathcal{K}_{inf}(\hat \nu_{1,s}',\mu'_1) \geq \frac{\delta_t}{s} + 2\varepsilon^2 \right) \right] \leq \frac{81(2 + \ln \ln T)}{\varepsilon^4}.
    \end{align*}
\end{lemma}
(see proof of Lemma \ref{lem:terme2} in \autoref{sec:proofterme2}). \\
Lemma \ref{lem:terme2} is also using the same reasoning as in the proof of \texttt{KL-UCB} by using the concentration Lemma \ref{lem:concentre1}.

\begin{lemma}\label{lem:terme3}
With the same assumption as the Lemma \ref{lem:decomposition} with also $\varepsilon \leq 1$ and $\eta := \mu'_1 + L_1 - \mu^{\dagger} + \varepsilon > 0$,
\begin{align*}
    &\mathbb{E}_{\nu',\pi',\nu,\pi}\left[\sum_{t=K+1}^T \sum_{s = 1}^{t}  \mathbf{1}\left( \mathcal{K}_{inf}(\hat \nu_{1,s},\mu_1)  > \frac{\delta_t -  N'_1  \mathcal{K}_{inf}(\hat \nu_1',\mu^{\dagger} - \varepsilon-L_1)}{s} + 2\varepsilon^2 \right) \right] \\
    &\leq \left(\frac{ 2 +18  \ln(1/(1-\mu'_1))}{\left( 1 - e^{-2 \eta^2} \right)^3} + 2 \right)\frac{81(2 + \ln \ln T)}{\varepsilon^4}.
\end{align*}
\end{lemma}
(see proof of Lemma \ref{lem:terme3} in \autoref{sec:proofterme3}). \\
Lemma \ref{lem:terme3} is using the independence between $(\nu',\pi')$ and $\nu$ to do concentration of conditional distribution. Then it requires a control of the prior terms.

\begin{lemma}\label{lem:terme4}

We always have
\begin{equation*}
    \mathbb{E}_{\nu',\pi',\nu,\pi}\left[(\ln T - N'_a\mathcal{K}_{inf}( \hat \nu'_a,\mu^{\dagger} - \varepsilon - L_a))_+ \right] \leq \ln T.
\end{equation*}
If we also have $\mu'_a < \mu^{\dagger} - L_a$, then for $\varepsilon \leq \frac{( 1 - (\mu^{\dagger} - L_a))\mathcal{K}_{inf}(\nu'_{a},\mu^{\dagger} - L_a) }{4}$,
\begin{align*}
    &\mathbb{E}_{\nu',\pi',\nu,\pi}\left[(\ln T - N'_a\mathcal{K}_{inf}(\hat \nu'_a,\mu^{\dagger} - \varepsilon - L_a))_+ \right]
    \\ & \leq \mathbb{E}_{\nu',\pi'}\left[(\ln T - N'_a\mathcal{K}_{inf}( \nu'_a,\mu^{\dagger} - L_a))_+ \right] \\ 
    &+ \frac{4 \varepsilon \ln T}{(1 - \mu^{\dagger})\mathcal{K}_{inf}(\nu'_{a},\mu^{\dagger} - L_a)} + \frac{7\sqrt{\ln(T) \times \ln(\ln T)}}{(1 - \mu^{\dagger})^{5/4}\mathcal{K}_{inf}(\nu'_{a},\mu^{\dagger} - L_a)^{1/2}} + \frac{40}{(1 - \mu^{\dagger})^{5/4}\mathcal{K}_{inf}(\nu'_{a},\mu^{\dagger} - L_a)^{3/2}} \\
    & + \frac{4128}{(1 - \mu^{\dagger})^{5} (\mathcal{K}_{inf}(\nu'_{a},\mu^{\dagger} - L_a) \wedge1)^3} + 32 \mathcal{K}_{inf}(\nu'_{a},\mu^{\dagger} - L_a).
\end{align*}

\end{lemma}
(see proof of Lemma \ref{lem:terme4} in \autoref{sec:proofterme4}). \\
From those Lemmas the proof of the theorem will be easy

\begin{proof}[of Theorem \ref{thm:general}] We define
\begin{itemize}
    \item $\varepsilon := \left(1 \wedge \mu^{\dagger} \wedge \frac{(1 - \mu^{\dagger})\mathcal{K}_{inf}(\nu_a,\mu^{\dagger})}{4} \right)\left( \frac{\ln \ln T}{\ln T} \right)^{1/5}$ 
    \item $\varepsilon := \left(1 \wedge \mu^{\dagger} \wedge \frac{(1 - \mu^{\dagger})\mathcal{K}_{inf}(\nu_a,\mu^{\dagger})}{4} \wedge \frac{( 1 - (\mu^{\dagger} - L_a))\mathcal{K}_{inf}(\nu'_{a},\mu^{\dagger} - L_a) }{4} \right)\left( \frac{\ln \ln T}{\ln T} \right)^{1/5}$ \text{ if } $\mu'_a + L_a < \mu^{\dagger}$.
\end{itemize}
This choice of constant is to be able to use the Lemmas. For every admissible $\mu^{\dagger}$
\begin{align*}
    \mathbb{E}_{\nu',\pi',\nu,\pi}[N_a(T)] & \leq \frac{\ln T}{\mathcal{K}_{inf}(\nu_a,\mu^{\dagger})} + \frac{\ln \ln T}{\mathcal{K}_{inf}(\nu_a,\mu^{\dagger})} +\frac{4 \varepsilon (\ln T + \ln \ln T)}{\mathcal{K}_{inf}(\nu_a,\mu^{\dagger})^2(1 - \mu^{\dagger})} + 10\\
    & + \left(\frac{ 2 +18  \ln(1/(1-\mu'_1))}{\left( 1 - e^{-2 \eta^2} \right)^3} + 3 \right)\frac{81(2 + \ln \ln T)}{\varepsilon^4} + K + \frac{16}{\varepsilon^2(1 - \mu^{\dagger})^{1/2}},
\end{align*}
where $\eta := \mu'_1 + L_1 - \mu^{\dagger} + \varepsilon$ converges to $\mu'_1 + L_1 - \mu^{\dagger} > 0$ because $\mu^{\dagger} < \mu'_1 + L_1$. From this decomposition and by using that $\varepsilon \asymp \left( \frac{\ln \ln T}{\ln T} \right)^{1/5}$ we got our result.
\\
\\
When $\mu'_a + L_a < \mu^{\dagger}$, using in addition the second part of Lemma \ref{lem:terme4} (with the extra $\varepsilon$-constraint from the second bullet above), we get
\begin{align*}
    \mathbb{E}_{\nu',\pi',\nu,\pi}[N_a(T)] & \leq \frac{\mathbb{E}_{\nu',\pi'}\left[(\ln T - N'_a\mathcal{K}_{inf}(\nu'_a,\mu^\dagger - L_a))_+\right]}{\mathcal{K}_{inf}(\nu_a,\mu^{\dagger})} + \frac{\ln \ln T}{\mathcal{K}_{inf}(\nu_a,\mu^{\dagger})} +  \frac{4 \varepsilon (\ln T + \ln \ln T)}{\mathcal{K}_{inf}(\nu_a,\mu^{\dagger})^2(1 - \mu^{\dagger})}  \\
    & + 10 + \left(\frac{ 2 +18  \ln(1/(1-\mu'_1))}{\left( 1 - e^{-2 \eta^2} \right)^3} + 3 \right)\frac{81(2 + \ln \ln T)}{\varepsilon^4} + K + \frac{16}{\varepsilon^2(1 - \mu^{\dagger})^{1/2}} \\
    & + \frac{4\varepsilon\ln T}{(1-\mu^\dagger)\mathcal K_{inf}(\nu'_a,\mu^\dagger-L_a)} + \frac{7\sqrt{\ln(T) \times \ln(\ln T)}}{(1 - \mu^{\dagger})^{5/4}\mathcal{K}_{inf}(\nu'_{a},\mu^{\dagger} - L_a)^{1/2}} \\
    &+ \frac{40}{(1 - \mu^{\dagger})^{5/4}\mathcal{K}_{inf}(\nu'_{a},\mu^{\dagger} - L_a)^{3/2}}  + \frac{4128}{(1-\mu^\dagger)^5(\mathcal K_{inf}(\nu'_a,\mu^\dagger-L_a)\wedge1)^3} \\
    & + 32\mathcal K_{inf}(\nu'_a,\mu^\dagger-L_a),
\end{align*}
with $\varepsilon \asymp \left( \frac{\ln \ln T}{\ln T} \right)^{1/5}$ we got again our result.
\end{proof}

\subsection{Proof of Lemma \ref{lem:decomposition}}\label{sec:proofdecomposition}

To prove it we will use the following Lemma 
\begin{lemma}\label{lem:noproof}
For all $u > 0$, \ $ \left\{ \mathcal{K}_{inf}(\nu,\mu- \varepsilon) \geq u \right\} \subset \left\{ \mathcal{K}_{inf}(\nu,\mu) \geq u + 2 \varepsilon^2 \right\}$.   
\end{lemma}
Lemma \ref{lem:noproof} is a direct consequence of the second part of Lemma \ref{lem:Kinfepsilon}.

\begin{proof}[ of Lemma \ref{lem:decomposition}]
    \begin{align*}
        N_a(T) & = \sum_{t=1}^T \mathbf{1}\left( A_t = a \right) \\
        & \leq K + \sum_{t=K+1}^T \mathbf{1}\left( A_t = a \right) \\
        & \leq K + \sum_{t=K+1}^T \mathbf{1}\left(U_1(t) < \mu^{\dagger} - \varepsilon\ \cap \ A_t = a \right) + \sum_{t=K+1}^T \mathbf{1}\left(U_1(t) \geq \mu^{\dagger} - \varepsilon\ \cap \ A_t = a \right) \\
        & \leq K + \underbrace{\sum_{t=K+1}^T \mathbf{1}\left(U_1(t) < \mu^{\dagger} - \varepsilon \right)}_{A} + \underbrace{\sum_{t=K+1}^T \mathbf{1}\left(U_a(t) \geq \mu^{\dagger} - \varepsilon\ \cap \ A_t = a \right)}_{B}.
    \end{align*}
Let us look at $A$, we have
\begin{align*}
    A&=\sum_{t=K+1}^T \mathbf{1}\left(U_1(t) < \mu^{\dagger} - \varepsilon \right) \\
    & \leq \sum_{t=K+1}^T \mathbf{1}\left(N_1(t)\mathcal{K}_{inf}(\hat \nu_1(t),\mu^{\dagger} - \varepsilon) + N'_1  \mathcal{K}_{inf}(\hat \nu_1',\mu^{\dagger} - \varepsilon-L_1) > \delta_t \right) \\
    & = \sum_{t=K+1}^T \mathbf{1}\left(N_1(t)\mathcal{K}_{inf}(\hat \nu_1(t),\mu^{\dagger} - \varepsilon)  > \delta_t -  N'_1  \mathcal{K}_{inf}(\hat \nu_1',\mu^{\dagger} - \varepsilon-L_1) \right). 
\end{align*}
Now we want to use Lemma \ref{lem:noproof} but it requires to have $G_{1,t}^{\dagger} := \delta_t -  N'_1  \mathcal{K}_{inf}(\hat \nu_1',\mu^{\dagger} - \varepsilon-L_1) > 0$. Then we will split according to this event
\begin{align*}
    &\mathbf{1}\left(N_1(t)\mathcal{K}_{inf}(\hat \nu_1(t),\mu^{\dagger} - \varepsilon)  > G_{1,t}^{\dagger} \right)  \\
    & = \mathbf{1}\left(G_{1,t}^{\dagger} > 0 \ \cap  \ N_1(t)\mathcal{K}_{inf}(\hat \nu_1(t),\mu^{\dagger} - \varepsilon)  > G_{1,t}^{\dagger} \right) + \mathbf{1}\left(G_{1,t}^{\dagger} \leq 0 \ \cap \ N_1(t)\mathcal{K}_{inf}(\hat \nu_1(t),\mu^{\dagger} - \varepsilon)  > G_{1,t}^{\dagger} \right) \\
    & \leq \mathbf{1}\left(G_{1,t}^{\dagger} > 0 \ \cap  \ N_1(t)\mathcal{K}_{inf}(\hat \nu_1(t),\mu^{\dagger} - \varepsilon)  > G_{1,t}^{\dagger} \right) + \mathbf{1}\left(G_{1,t}^{\dagger} \leq 0\right).
\end{align*}
Using Lemma \ref{lem:noproof} and that $\mu_1 - \mu^{\dagger}  \geq  0$ we get 
\begin{align*}
    &\mathbf{1}\left(\mathcal{K}_{inf}(\hat \nu_1(t),\mu^{\dagger} - \varepsilon)  > \frac{G_{1,t}^{\dagger}}{N_1(t)} \right)  \\
    & \leq   \mathbf{1}\left(G_{1,t}^{\dagger} > 0 \ \cap  \ N_1(t)\mathcal{K}_{inf}(\hat \nu_1(t),\mu_1)  > G_{1,t}^{\dagger} + N_1(t)2 (\mu_1 - \mu^{\dagger} + \varepsilon)^2 \right) + \mathbf{1}\left(G_{1,t}^{\dagger} \leq 0\right) \\
    & \leq  \mathbf{1}\left( N_1(t)\mathcal{K}_{inf}(\hat \nu_1(t),\mu_1)  > G_{1,t}^{\dagger} + N_1(t)2 \varepsilon^2 \right) + \mathbf{1}\left(G_{1,t}^{\dagger} \leq 0\right).
\end{align*}
However we have $ \mathbf{1}\left(G_{1,t}^{\dagger} \leq 0\right)  = \mathbf{1}\left(N'_1  \mathcal{K}_{inf}(\hat \nu_1',\mu^{\dagger} - \varepsilon-L_1) \geq \delta_t  \right)$.
\\
Then by using again Lemma \ref{lem:noproof} thanks to $\delta_t > 0$ and $\mu'_1 + L_1  -\mu^{\dagger} > 0$ we get
\begin{align*}
    \mathbf{1}\left(G_{1,t}^{\dagger} \leq 0 \right)
      &= \mathbf{1}\left(N'_1  \mathcal{K}_{inf}(\hat \nu_1',\mu^{\dagger} - \varepsilon-L_1) \geq \delta_t  \right) \\
    & \leq \mathbf{1}\left( N'_1\mathcal{K}_{inf}(\hat \nu_1',\mu'_1) \geq \delta_t + N'_12(\mu'_1 + L_1 + \varepsilon -\mu^{\dagger})^2 \right) \\
    & \leq \mathbf{1}\left( N'_1\mathcal{K}_{inf}(\hat \nu_1',\mu'_1) \geq \delta_t + N'_12\varepsilon^2 \right).
\end{align*}
Then we have
\begin{align*}
     \mathbf{1}\left(U_1(t) < \mu^{\dagger} - \varepsilon \right) & \leq \mathbf{1}\left( N_1(t)\mathcal{K}_{inf}(\hat \nu_1(t),\mu_1)  > G_{1,t}^{\dagger} + N_1(t)2 \varepsilon^2 \right) + \mathbf{1}\left( N'_1\mathcal{K}_{inf}(\hat \nu_1',\mu'_1) \geq \delta_t + N'_12\varepsilon^2 \right) \\
    & = \mathbf{1}\left( N_1(t)\mathcal{K}_{inf}(\hat \nu_1(t),\mu_1)  > \delta_t -  N'_1  \mathcal{K}_{inf}(\hat \nu_1',\mu^{\dagger} - \varepsilon-L_1) + N_1(t)2 \varepsilon^2 \right)\\
    & \hspace{0.5cm}+\mathbf{1}\left( N'_1\mathcal{K}_{inf}(\hat \nu_1',\mu'_1) \geq \delta_t + N'_12\varepsilon^2 \right).
\end{align*}
Now we decompose according to the possible values of $N_1(t)$ which are $\{1,...,t\}$ and for $N'_1$ which are $\{0,...,+ \infty\}$ but for $N'_1 = 0$ the term is $0$, it gives
\begin{align*}
    A & \leq \sum_{t=K+1}^T \sum_{s = 1}^{t}  \mathbf{1}\left( \mathcal{K}_{inf}(\hat \nu_{1,s},\mu_1)  > \frac{\delta_t -  N'_1  \mathcal{K}_{inf}(\hat \nu_1',\mu^{\dagger} - \varepsilon-L_1)}{s} + 2\varepsilon^2 \right)\\
    &+ \sum_{t=K+1}^T\sum_{s = 1}^{+ \infty}\mathbf{1}\left( \mathcal{K}_{inf}(\hat \nu_{1,s}',\mu'_1) \geq \frac{\delta_t}{s} + 2\varepsilon^2 \right).
\end{align*}
Now let us control $B$.
\begin{align*}
    B & = \sum_{t=K+1}^T \mathbf{1}\left(U_a(t) \geq \mu^{\dagger} - \varepsilon\ \cap \ A_t = a \right) \\
    & \leq \sum_{t=K+1}^T \mathbf{1}\left(N_a(t)\mathcal{K}_{inf}(\hat \nu_a(t),\mu^{\dagger} - \varepsilon) + N'_a\mathcal{K}_{inf}(\hat \nu'_a,\mu^{\dagger} - \varepsilon - L_a) \leq \delta_t\ \cap \ A_t = a \right) \\
    & \leq \sum_{t=K+1}^T \sum_{s = 1}^{t} \mathbf{1}\left(s\mathcal{K}_{inf}(\hat \nu_{a,s},\mu^{\dagger} - \varepsilon) + N'_a\mathcal{K}_{inf}(\hat \nu'_a,\mu^{\dagger} - \varepsilon - L_a) \leq \delta_t\ \cap \ A_t = a \ \cap \ N_a(t) = s \right) \\
    & \leq \sum_{t=K+1}^T \sum_{s = 1}^{T} \mathbf{1}\left(s\mathcal{K}_{inf}(\hat \nu_{a,s},\mu^{\dagger} - \varepsilon) + N'_a\mathcal{K}_{inf}(\hat \nu'_a,\mu^{\dagger} - \varepsilon - L_a) \leq \delta_T\ \cap \ A_t = a \ \cap \ N_a(t) = s \right) \\
    & = \sum_{s = 1}^{T}\mathbf{1}\left(s\mathcal{K}_{inf}(\hat \nu_{a,s},\mu^{\dagger} - \varepsilon) + N'_a\mathcal{K}_{inf}(\hat \nu'_a,\mu^{\dagger} - \varepsilon - L_a) \leq \delta_T \right) \sum_{t=K+1}^T \mathbf{1}\left( A_t = a \ \cap \ N_a(t) = s \right).
\end{align*}
Now using the fundamental inequality in bandit \eqref{eq:fondaineq},
\begin{align*}
    B \leq \sum_{s = 1}^{T}\mathbf{1}\left(\mathcal{K}_{inf}(\hat \nu_{a,s},\mu^{\dagger} - \varepsilon) \leq \frac{(\delta_T - N'_a\mathcal{K}_{inf}(\hat \nu'_a,\mu^{\dagger} - \varepsilon - L_a))_+}{s} \right).
\end{align*}
Now we define 
\begin{equation*}
    \Gamma := \frac{(\delta_T - N'_a\mathcal{K}_{inf}(\hat \nu'_a,\mu^{\dagger} - \varepsilon - L_a))_+}{\mathcal{K}_{inf}(\nu_a,\mu^{\dagger}) - \frac{2 \varepsilon}{1 - \mu^{\dagger}}}.
\end{equation*}
We get
\begin{align*}
    B& \leq \Gamma + \sum_{s = \Gamma \vee 1}^{T}\mathbf{1}\left(\mathcal{K}_{inf}(\hat \nu_{a,s},\mu^{\dagger} - \varepsilon) \leq \frac{(\delta_T - N'_a\mathcal{K}_{inf}(\hat \nu'_a,\mu^{\dagger} - \varepsilon - L_a))_+}{s} \right).
\end{align*}
Now we have two cases,
\begin{itemize}
    \item If $\Gamma = 0$, then 
    \begin{equation*}
        \frac{(\delta_T - N'_a\mathcal{K}_{inf}(\hat \nu'_a,\mu^{\dagger} - \varepsilon - L_a))_+}{s} = 0 \leq  \mathcal{K}_{inf}(\nu_a,\mu^{\dagger}) - \frac{2 \varepsilon}{1 - \mu^{\dagger}}.
    \end{equation*}
    \item If $\Gamma > 0$, then for $s \geq \Gamma$
    \begin{equation*}
        \frac{(\delta_T - N'_a\mathcal{K}_{inf}(\hat \nu'_a,\mu^{\dagger} - \varepsilon - L_a))_+}{s} \leq \frac{(\delta_T - N'_a\mathcal{K}_{inf}(\hat \nu'_a,\mu^{\dagger} - \varepsilon - L_a))_+}{\Gamma} = \mathcal{K}_{inf}(\nu_a,\mu^{\dagger}) - \frac{2 \varepsilon}{1 - \mu^{\dagger}}.
    \end{equation*}
\end{itemize}
Then in all cases we have
\begin{align*}
    B &\leq \Gamma + \sum_{s = \Gamma \vee 1}^{T}\mathbf{1}\left(\mathcal{K}_{inf}(\hat \nu_{a,s},\mu^{\dagger} - \varepsilon) \leq \mathcal{K}_{inf}(\nu_a,\mu^{\dagger}) - \frac{2 \varepsilon}{1 - \mu^{\dagger}} \right) \\
    & \leq \Gamma + \sum_{s = 1}^{+ \infty}\mathbf{1}\left(\mathcal{K}_{inf}(\hat \nu_{a,s},\mu^{\dagger} - \varepsilon) \leq \mathcal{K}_{inf}(\nu_a,\mu^{\dagger}) - \frac{2 \varepsilon}{1 - \mu^{\dagger}} \right).
\end{align*}
Now using the first part of Lemma \ref{lem:Kinfepsilon} we get
\begin{align*}
    B& \leq \Gamma + \sum_{s = 1}^{+ \infty}\mathbf{1}\left(\mathcal{K}_{inf}(\hat \nu_{a,s},\mu^{\dagger}) - \frac{\varepsilon}{1 - \mu^{\dagger}} \leq \mathcal{K}_{inf}(\nu_a,\mu^{\dagger}) - \frac{2 \varepsilon}{1 - \mu^{\dagger}} \right) \\
    & = \Gamma + \sum_{s = 1}^{+ \infty}\mathbf{1}\left(\mathcal{K}_{inf}(\hat \nu_{a,s},\mu^{\dagger}) \leq \mathcal{K}_{inf}(\nu_a,\mu^{\dagger}) - \frac{ \varepsilon}{1 - \mu^{\dagger}} \right).
\end{align*}
Now if we look at $\Gamma$, we have
\begin{align*}
    \Gamma  = \frac{(\delta_T - N'_a\mathcal{K}_{inf}(\hat \nu'_a,\mu^{\dagger} - \varepsilon - L_a))_+}{\mathcal{K}_{inf}(\nu_a,\mu^{\dagger}) } \times \frac{1}{1 -  \frac{2 \varepsilon}{\mathcal{K}_{inf}(\nu_a,\mu^{\dagger})(1 - \mu^{\dagger})}},
\end{align*}
the inequality $\frac{1}{1 - x} \leq 1 + 2x$ for $x \leq 1/2$ with  $\varepsilon \leq \frac{\mathcal{K}_{inf}(\nu_a,\mu^{\dagger})(1 - \mu^{\dagger})}{4}$ gives
\begin{align*}
    \Gamma & \leq \frac{(\delta_T - N'_a\mathcal{K}_{inf}(\hat \nu'_a,\mu^{\dagger} - \varepsilon - L_a))_+}{\mathcal{K}_{inf}(\nu_a,\mu^{\dagger}) } \left(1 + \frac{4 \varepsilon}{\mathcal{K}_{inf}(\nu_a,\mu^{\dagger})(1 - \mu^{\dagger})} \right) \\
    & \leq \frac{(\delta_T - N'_a\mathcal{K}_{inf}(\hat \nu'_a,\mu^{\dagger} - \varepsilon - L_a))_+}{\mathcal{K}_{inf}(\nu_a,\mu^{\dagger}) } + \frac{4 \varepsilon \delta_T}{\mathcal{K}_{inf}(\nu_a,\mu^{\dagger})^2(1 - \mu^{\dagger})}.
\end{align*}
As $\delta_T := \ln T + \ln \ln T$ we get
\begin{equation*}
    \Gamma \leq \frac{(\ln T - N'_a\mathcal{K}_{inf}(\hat \nu'_a,\mu^{\dagger} - \varepsilon - L_a))_+}{\mathcal{K}_{inf}(\nu_a,\mu^{\dagger}) } + \frac{\ln \ln T}{\mathcal{K}_{inf}(\nu_a,\mu^{\dagger})} +  \frac{4 \varepsilon (\ln T + \ln \ln T)}{\mathcal{K}_{inf}(\nu_a,\mu^{\dagger})^2(1 - \mu^{\dagger})}.
\end{equation*}
Finally we have
\begin{align*}
    N_a(T) & \leq \frac{(\ln T - N'_a\mathcal{K}_{inf}(\hat \nu'_a,\mu^{\dagger} - \varepsilon - L_a))_+}{\mathcal{K}_{inf}(\nu_a,\mu^{\dagger}) } + \frac{\ln \ln T}{\mathcal{K}_{inf}(\nu_a,\mu^{\dagger})} +  \frac{4 \varepsilon (\ln T + \ln \ln T)}{\mathcal{K}_{inf}(\nu_a,\mu^{\dagger})^2(1 - \mu^{\dagger})} \\
    & + \sum_{s = 1}^{+ \infty}\mathbf{1}\left(\mathcal{K}_{inf}(\hat \nu_{a,s},\mu^{\dagger}) \leq \mathcal{K}_{inf}(\nu_a,\mu^{\dagger}) - \frac{ \varepsilon}{1 - \mu^{\dagger}} \right) \\
    & + \sum_{t=K+1}^T \sum_{s = 1}^{t}  \mathbf{1}\left( \mathcal{K}_{inf}(\hat \nu_{1,s},\mu_1)  > \frac{\delta_t -  N'_1  \mathcal{K}_{inf}(\hat \nu_1',\mu^{\dagger} - \varepsilon-L_1)}{s} + 2\varepsilon^2 \right) \\
    & + \sum_{t=K+1}^T\sum_{s = 1}^{+ \infty}\mathbf{1}\left( \mathcal{K}_{inf}(\hat \nu_{1,s}',\mu'_1) \geq \frac{\delta_t}{s} + 2\varepsilon^2 \right) + K.
\end{align*}
\end{proof}

\subsection{Proof of Lemma \ref{lem:terme1}}\label{sec:proofterme1}

\begin{proof}[ of Lemma \ref{lem:terme1}]

\begin{align*}
    \mathbb{E}_{\nu',\pi',\nu,\pi}&\!\left[\sum_{s = 1}^{+ \infty}\mathbf{1}\!\left(\!\mathcal{K}_{inf}(\hat \nu_{a,s},\mu^{\dagger}) \leq \mathcal{K}_{inf}(\nu_a,\mu^{\dagger}) \!-\! \frac{ \varepsilon}{1 \!-\! \mu^{\dagger}} \right) \right]  
    \\&= \sum_{s = 1}^{+ \infty}\mathbb{P}_{\nu}\left(\mathcal{K}_{inf}(\hat \nu_{a,s},\mu^{\dagger}) \leq \mathcal{K}_{inf}(\nu_a,\mu^{\dagger}) - \frac{ \varepsilon}{1 - \mu^{\dagger}} \right).
\end{align*}
Using the concentration Lemma \ref{lem:concentre2} we get
\begin{align*}
    & \leq \sum_{s = 1}^{+ \infty} \exp\left( - s \left( \frac{1}{4} \wedge \frac{\varepsilon^2}{(1 - \mu^{\dagger})^2 2 \gamma_{\mu^{\dagger}}} \right) \right)  \leq \frac{1}{1 - \exp\left( -  \left( \frac{1}{4} \wedge \frac{\varepsilon^2}{(1 - \mu^{\dagger})^2 2 \gamma_{\mu^{\dagger}}} \right) \right)}.
\end{align*}
Using the inequalities $ \frac{1}{1 - e^{-x}} \leq 2 +  \frac{2}{x}$, \ $a \vee b \leq a + b$ when it is positive and $\gamma_{\mu^{\dagger}} \leq \frac{4}{(1 - \mu^{\dagger})^{5/2}}$,
\begin{equation*}
    \leq 2  + \frac{2}{\left( \frac{1}{4} \wedge \frac{\varepsilon^2}{(1 - \mu^{\dagger})^2 2 \gamma_{\mu^{\dagger}}} \right)}  = 2 + \left( 8  \vee \frac{4(1 - \mu^{\dagger})^2\gamma_{\mu^{\dagger}}}{\varepsilon^2}  \right) \leq 10 + \frac{4(1 - \mu^{\dagger})^2\gamma_{\mu^{\dagger}}}{\varepsilon^2}  \leq 10 + \frac{16}{\varepsilon^2(1 - \mu^{\dagger})^{1/2}}.
\end{equation*}

\end{proof}

\subsection{Proof of Lemma \ref{lem:terme2}}\label{sec:proofterme2}

\begin{proof}[ of Lemma \ref{lem:terme2}]

\begin{align*}
    &\mathbb{E}_{\nu',\pi',\nu,\pi}\left[\sum_{t=K+1}^T\sum_{s = 1}^{+ \infty}\mathbf{1}\left( \mathcal{K}_{inf}(\hat \nu_{1,s}',\mu'_1) \geq \frac{\delta_t}{s} + 2\varepsilon^2 \right) \right]  = \sum_{t=K+1}^T\sum_{s = 1}^{+ \infty}\mathbb{P}_{\nu'}\left( \mathcal{K}_{inf}(\hat \nu_{1,s}',\mu'_1) \geq \frac{\delta_t}{s} + 2\varepsilon^2 \right).
\end{align*}
Using the concentration Lemma \ref{lem:concentre1} we get
\begin{align*}
    &\leq \sum_{t=K+1}^T\sum_{s = 1}^{+ \infty} e(2s+1)e^{-s \left( \frac{\delta_t}{s} + 2 \varepsilon^2 \right)}  = \left(\sum_{t=K+1}^T e^{- \delta_t} \right) \left(\sum_{s = 1}^{+ \infty} e(2s+1)e^{-s2 \varepsilon^2}\right).
\end{align*}
We have
\begin{align*}
    \sum_{t=K+1}^T e^{- \delta_t} & = \sum_{t=K+1}^T \frac{1}{t \ln t} \leq 2 + \ln \ln T,
\end{align*}
and
\begin{align*}
    \sum_{s = 1}^{+ \infty} \underbrace{e}_{\leq 3}(2s+\underbrace{1}_{\leq s})e^{-s2 \varepsilon^2}  \leq 9\sum_{s = 1}^{+ \infty}se^{-s2 \varepsilon^2}  \leq \frac{9}{\left( 1 - e^{-2 \varepsilon^2} \right)^2}.
\end{align*}
Using that $\frac{1}{1 - e^{-x}} \leq 2 + \frac{2}{x}$  and that $\varepsilon \leq 1$, we get
\begin{align*}
    \frac{1}{\left( 1 - e^{-2 \varepsilon^2} \right)^2} \leq \left(2 + \frac{2}{2 \varepsilon^2} \right)^2  = \left(2 + \frac{1}{ \varepsilon^2} \right)^2 
     \leq  \frac{9}{ \varepsilon^4}. 
\end{align*}
Finally we have
\begin{align*}
    &\mathbb{E}_{\nu',\pi',\nu,\pi}\left[\sum_{t=K+1}^T\sum_{s = 1}^{+ \infty}\mathbf{1}\left( \mathcal{K}_{inf}(\hat \nu_{1,s}',\mu'_1) \geq \frac{\delta_t}{s} + 2\varepsilon^2 \right) \right]  \leq (2 + \ln \ln T) \times 9 \times \frac{9}{ \varepsilon^4}  = \frac{81(2 + \ln \ln T)}{\varepsilon^4}.
\end{align*}
\end{proof}

\subsection{Proof of Lemma \ref{lem:terme3}}\label{sec:proofterme3}

To prove this lemma we will use an other lemma

\begin{lemma}\label{lem:inter}
For $\eta := \mu'_1 + L_1 - \mu^{\dagger} + \varepsilon > 0$
\begin{align*}
    \mathbb{E}_{\nu',\pi'}\left[ \exp\left( N'_1  \mathcal{K}_{inf}(\hat \nu_1',\mu^{\dagger} - \varepsilon-L_1) \right)\right] \leq \frac{ 2 +18  \ln(1/(1-\mu'_1))}{\left( 1 - e^{-2 \eta^2} \right)^3} + 2.
\end{align*}
\end{lemma}
(see proof of Lemma \ref{lem:inter} in \autoref{sec:proofinter}). \\
\begin{proof}[ of Lemma \ref{lem:terme3}]
We have
\begin{align*}
    &\mathbb{E}_{\nu',\pi',\nu,\pi}\left[\sum_{t=K+1}^T \sum_{s = 1}^{t}  \mathbf{1}\left( \mathcal{K}_{inf}(\hat \nu_{1,s},\mu_1)  > \frac{\delta_t -  N'_1  \mathcal{K}_{inf}(\hat \nu_1',\mu^{\dagger} - \varepsilon-L_1)}{s} + 2\varepsilon^2 \right) \right] \\
    & = \mathbb{E}_{\nu',\pi'}\left[\sum_{t=K+1}^T \sum_{s = 1}^{t}  \mathbb{P}\left( \mathcal{K}_{inf}(\hat \nu_{1,s},\mu_1)  > \frac{\delta_t -  N'_1  \mathcal{K}_{inf}(\hat \nu_1',\mu^{\dagger} - \varepsilon-L_1)}{s} + 2\varepsilon^2 \  \middle | \ N'_1, \hat \nu'_1 \right) \right].
\end{align*}
Using that $(\pi',\nu')$ and $\nu$ are independent we can consider $h_t := \delta_t -  N'_1  \mathcal{K}_{inf}(\hat \nu_1',\mu^{\dagger} - \varepsilon-L_1)$ as a constant. Using the same reasoning as in the proof of Lemma \ref{lem:terme2} we get
\begin{align*}
    &\sum_{t=K+1}^T \sum_{s = 1}^{t}\mathbb{P}\left( \mathcal{K}_{inf}(\hat \nu_{1,s},\mu_1)  > \frac{h_t}{s} + 2\varepsilon^2 \  \middle | \ N'_1, \hat \nu'_1 \right) \\
    & \leq  \left(\sum_{t=K+1}^T e^{- h_t} \right) \left(\sum_{s = 1}^{+ \infty} e(2s+1)e^{-s2 \varepsilon^2}\right) \\
    & = e^{ N'_1  \mathcal{K}_{inf}(\hat \nu_1',\mu^{\dagger} - \varepsilon-L_1)}\left(\sum_{t=K+1}^T e^{- \delta_t} \right)  \left(\sum_{s = 1}^{+ \infty} e(2s+1)e^{-s2 \varepsilon^2}\right) \\
    & \leq e^{ N'_1  \mathcal{K}_{inf}(\hat \nu_1',\mu^{\dagger} - \varepsilon-L_1)} \frac{81(2 + \ln \ln T)}{\varepsilon^4}.
\end{align*}
Then,
\begin{align*}
    &\mathbb{E}_{\nu',\pi',\nu,\pi}\left[\sum_{t=K+1}^T \sum_{s = 1}^{t}  \mathbf{1}\left( \mathcal{K}_{inf}(\hat \nu_{1,s},\mu_1)  > \frac{\delta_t -  N'_1  \mathcal{K}_{inf}(\hat \nu_1',\mu^{\dagger} - \varepsilon-L_1)}{s} + 2\varepsilon^2 \right) \right] \\
     & = \mathbb{E}_{\nu',\pi'}\left[\sum_{t=K+1}^T \sum_{s = 1}^{t}  \mathbb{P}\left( \mathcal{K}_{inf}(\hat \nu_{1,s},\mu_1)  > \frac{\delta_t -  N'_1  \mathcal{K}_{inf}(\hat \nu_1',\mu^{\dagger} - \varepsilon-L_1)}{s} + 2\varepsilon^2 \  \middle | \ N'_1, \hat \nu'_1 \right) \right] \\
     & \leq \mathbb{E}_{\nu',\pi'}\left[ e^{ N'_1  \mathcal{K}_{inf}(\hat \nu_1',\mu^{\dagger} - \varepsilon-L_1)}\right]\frac{81(2 + \ln \ln T)}{\varepsilon^4}.
\end{align*}
We conclude by using Lemma \ref{lem:inter}
\end{proof}

\subsection{Proof of Lemma \ref{lem:terme4}}\label{sec:proofterme4}

To prove this lemma we will use an other lemma

\begin{lemma}\label{lem:inter2}
For all $a > c > 0$, $b > 0$ we have $\int_{c}^{a}e^{-by^2}dy \leq \frac{e^{-bc^2}}{2bc}$.
\end{lemma}
(see proof of Lemma \ref{lem:inter2} in \autoref{sec:proofinter2}). \\
\begin{proof}[ of Lemma \ref{lem:terme4}] If $\mu'_a < \mu^{\dagger} - L_a$ (so $\mathcal{K}_{inf}(\nu'_a, \mu^{\dagger} - L_a) > 0$), then by using the first part of  Lemma \ref{lem:Kinfepsilon},
\begin{equation*}
    \Big(\ln T - N'_a\mathcal{K}_{inf}(\hat \nu'_a,\mu^{\dagger} - \varepsilon - L_a) \Big)_+ \leq  \left(\ln T - N'_a\mathcal{K}_{inf}(\hat \nu'_a,\mu^{\dagger} - \varepsilon - L_a) + N'_a \frac{\varepsilon}{1 - (\mu^{\dagger} - L_a)} \right)_+.
\end{equation*}
Now we introduce some notations to reduce the length of the equations:
\begin{multicols}{2}
\begin{itemize}
    \item $\delta := \ln T$
    \item $\hat D_s := \mathcal{K}_{inf}(\hat \nu'_{a,s},\mu^{\dagger} - L_a)$
    \item $D := \mathcal{K}_{inf}(\nu'_{a},\mu^{\dagger} - L_a)$
    \item $N := N'_a$
    \item $c := \frac{1}{1 - (\mu^{\dagger} - L_a)} \leq \frac{1}{1 - \mu^{\dagger}}$
    \item $ \beta := \frac{D - \varepsilon c}{2} \geq \frac{D}{4}$. 
\end{itemize}
\end{multicols}
\noindent
We want to show that
\begin{align*}
    &\mathbb{E}_{\nu',\pi',\nu,\pi}\left[\Big(\delta - N \hat D_{N}  + N \varepsilon c \Big)_+\right]  = \mathbb{E}_{\nu',\pi'}\left[\Big(\delta - N \hat D_{N}  + N \varepsilon c \Big)_+ \right] \leq \mathbb{E}_{\nu',\pi'}\left[\Big(\delta - N \hat D_{N}  \Big)_+ \right]  + \text{ small}.
\end{align*}
We split between two events
\begin{equation*}
    \left\{ N \geq \frac{\delta}{\beta} \right\}, \qquad \left\{ N < \frac{\delta}{\beta} \right\}.
\end{equation*}
For the first event, we can see first that
\begin{equation}\label{eq:doublevent1}
    \left\{ N \geq \frac{\delta}{\beta} \right\} = \left\{\beta - \frac{\delta}{N} \geq 0 \right\} = \left\{2\beta - \frac{\delta}{N} \geq \beta \right\} = \left\{D - \frac{\delta}{N} - c \varepsilon \geq \beta \right\}.
\end{equation}
We keep this in mind. We also have
\begin{align*}
    \Big(\delta - N \hat D_{N}  + N \varepsilon c \Big)_+ 
    & = \int_{0}^{+ \infty} \mathbf{1}\left(\Big(\delta - N \hat D_{N}  + N \varepsilon c \Big)_+ \geq x \right) dx \\
    & = \int_{0}^{+ \infty} \mathbf{1}\left(\delta - N \hat D_{N}  + N \varepsilon c  \geq x \right) dx \\
    & = \int_{0}^{+ \infty} \mathbf{1}\left(\hat D_N \leq \frac{\delta - x}{N} + c \varepsilon \right) dx.
\end{align*}
As $\hat D_N \geq 0$ we can remove the $x$ larger than $\delta + N c \varepsilon$. It gives
\begin{align*}
    & = \int_{0}^{\delta + N c \varepsilon} \mathbf{1}\left(\hat D_N \leq \frac{\delta - x}{N} + c \varepsilon \right) dx.
\end{align*}
Now we do the following change of variable
\begin{itemize}
    \item $\frac{\delta - x}{N} + c \varepsilon = D - y $
    \item $ y = D - \frac{\delta - x}{N} - c \varepsilon$ 
    \item $N dy =  dx$.
\end{itemize}
Then it's equal to
\begin{align*}
    N\int_{D - \frac{\delta }{N} - c \varepsilon}^{D} \mathbf{1}\left(\hat D_{N} \leq D - y \right) dy.
\end{align*}
Then we have
\begin{align*}
    & \mathbb{E}_{\nu',\pi'}\left[\mathbf{1}\left(N \geq \frac{\delta}{\beta} \right)\Big(\delta - N \hat D_{N}  + N \varepsilon c \Big)_+ \right] \\
    & = \mathbb{E}_{\nu',\pi'}\left[\mathbf{1}\left(D - \frac{\delta}{N} - c \varepsilon \geq \beta\right)N\int_{D - \frac{\delta }{N} - c \varepsilon}^{D} \mathbf{1}\left(\hat D_{N} \leq D - y \right) dy \right] \\
    & \leq \mathbb{E}_{\nu',\pi'}\left[\mathbf{1}\left(D - \frac{\delta}{N} - c \varepsilon \geq \beta\right)N\int_{\beta}^{D} \mathbf{1}\left(\hat D_{N} \leq D - y \right) dy \right] \\ 
    & \leq \mathbb{E}_{\nu',\pi'}\left[N\int_{\beta}^{D} \mathbf{1}\left(\hat D_{N} \leq D - y \right) dy \right] \\
    & \leq \sum_{s = 1}^{+ \infty}s\int_{\beta}^{D} \mathbb{P}_{\nu'}\left(\hat D_{s} \leq D - y \right) dy  \\
    & \leq \sum_{s = 1}^{+ \infty} s \left(\int_{\beta}^{\gamma/2} \mathbb{P}_{\nu'}\left(\hat D_{s} \leq D - y \right) dy + \int_{\gamma/2}^{D} \mathbb{P}_{\nu'}\left(\hat D_{s} \leq D - y \right) dy \right).
\end{align*}
Using the concentration Lemma \ref{lem:concentre2} we get
\begin{align*}
    &\leq \sum_{s = 1}^{+ \infty} s \left(\int_{\beta}^{\gamma/2} \exp\left( -  \frac{s y^2}{2 \gamma} \right)dy + \int_{\gamma/2}^{D} \exp\left( - s/4 \right) dy \right).
\end{align*}
Using Lemma \ref{lem:inter2} for the first integral gives
\begin{align*}
    &\leq \sum_{s = 1}^{+ \infty} s \left( \frac{\exp\left( - \frac{s \beta^2}{2 \gamma}\right)}{2s \frac{\beta}{2 \gamma}} + D \exp\left( - s/4 \right)  \right) \\
    & = \frac{\gamma}{\beta} \sum_{s = 1}^{+ \infty}\exp\left( - \frac{s \beta^2}{2 \gamma}\right) + D \sum_{s = 1}^{+ \infty}s\exp\left( - s/4 \right)  \leq \frac{\gamma/ \beta}{1 -\exp\left( - \frac{ \beta^2}{2 \gamma}\right) } + D \frac{e^{-1/4}}{\left(1 - e^{-1/4}\right)^2}. 
\end{align*}
Using that $\frac{e^{-1/4}}{\left(1 - e^{-1/4}\right)^2} \leq 16$ and $\frac{1}{1 - e^{-x}} \leq 2 + \frac{2}{x}$ we get
\begin{align*}
    & \leq \frac{\gamma}{\beta} \left( 2 + \frac{2}{\frac{\beta^2}{2 \gamma}} \right) + 16 D  = \frac{2\gamma}{\beta} + \frac{4\gamma^2}{\beta^3} + 16 D.
\end{align*}
Using that $\beta \geq \frac{D}{4}$ and $\gamma \leq \frac{4}{(1 - (\mu^{\dagger} - L_a))^{5/2}}$ we get
\begin{align*}
    & \leq \frac{32}{(1 - (\mu^{\dagger} - L_a))^{5/2} D} + \frac{4096}{(1- (\mu^{\dagger} - L_a))^{5} D^3} + 16 D  \leq \frac{4128}{(1 - (\mu^{\dagger} - L_a))^{5} (D \wedge1)^3} + 16 D.
\end{align*}
For the second event \eqref{eq:doublevent1},
\begin{align*}
    \mathbb{E}_{\nu',\pi'}\left[\mathbf{1}\left(N < \frac{\delta}{\beta} \right)\Big(\delta - N \hat D_{N}  + N \varepsilon c \Big)_+ \right] 
    & \leq \mathbb{E}_{\nu',\pi'}\left[\mathbf{1}\left(N < \frac{\delta}{\beta} \right)\Big(\delta - N \hat D_{N}  + \frac{\delta}{\beta} \varepsilon c \Big)_+ \right] \\
    & \leq \mathbb{E}_{\nu',\pi'}\left[\mathbf{1}\left(N < \frac{\delta}{\beta} \right)\Big(\delta - N \hat D_{N}  \Big)_+ \right]  + \frac{c}{\beta} \varepsilon \delta.
\end{align*}
Then we have
\begin{align*}
   \Big(\delta - N \hat D_{N}  \Big)_+ & = \int_{0}^{+ \infty} \mathbf{1}\left(\Big(\delta - N\hat D_{N}\Big)_+ > x  \right) \\
   & = \int_{0}^{+ \infty} \mathbf{1}\left(\delta - N\hat D_{N} > x  \right) \\
   & = \int_{0}^{+ \infty} \mathbf{1}\left(\hat D_{N} < \frac{\delta - x}{N}  \right) \\
   & = \int_{0}^{\delta} \mathbf{1}\left(\hat D_{N} < \frac{\delta - x}{N}  \right).
\end{align*}
Now we do the change of variable $\frac{\delta - x}{N} = D - y/N$,
\begin{align*}
    & = \int_{ND - \delta}^{ND} \mathbf{1}\left(\hat D_{N} < D - \frac{y}{N}  \right)dy \\
    & = (\delta - ND)_+ + \int_{0}^{ND} \mathbf{1}\left(\hat D_{N} < D - \frac{y}{N}  \right)dy.
\end{align*}
Then we just need to control
\begin{equation*}
    \mathbb{E}_{\nu',\pi'}\left[\mathbf{1}\left(N < \frac{\delta}{\beta} \right)\int_{0}^{ND} \mathbf{1}\left(\hat D_{N} < D - \frac{y}{N}  \right)dy \right]. 
\end{equation*}
Let us note $t_0 := A \sqrt{\delta \ln (\delta)}$ with any constant $A > 0$. We have
\begin{align*}
    &\mathbf{1}\left(N < \frac{\delta}{\beta} \right)\int_{0}^{ND} \mathbf{1}\left(\hat D_{N} < D - \frac{y}{N}  \right)dy  \\
    & \leq t_0 + \mathbf{1}\left(N < \frac{\delta}{\beta} \right)\int_{t_0}^{ND} \mathbf{1}\left(\hat D_{N} < D - \frac{y}{N}  \right)dy \\
    & \leq t_0 + \sum_{n=1}^{\delta/\beta}\int_{t_0}^{nD} \mathbf{1}\left(\hat D_{n} < D - \frac{y}{n}  \right)dy \\
    & = t_0 + \sum_{n=1}^{\delta/\beta}\int_{t_0}^{n\gamma/2} \mathbf{1}\left(\hat D_{n} < D - \frac{y}{n}  \right)dy + \sum_{n=1}^{\delta/\beta}\int_{n\gamma/2}^{nD} \mathbf{1}\left(\hat D_{n} < D - \frac{y}{n}  \right)dy.
\end{align*}
By taking the mean and by using the Lemma \ref{lem:concentre2} and Lemma \ref{lem:inter2} we get
\begin{align*}
    &\mathbb{E}_{\nu',\pi'}\left[\mathbf{1}\left(N < \frac{\delta}{\beta} \right)\int_{0}^{ND} \mathbf{1}\left(\hat D_{N} < D - \frac{y}{N}  \right)dy \right]  \\
    & \leq t_0 + \sum_{n=1}^{\delta/\beta}\int_{t_0}^{n\gamma/2} \exp\left(- \frac{y^2}{2 \gamma n} \right)dy + \sum_{n=1}^{\delta/\beta}\int_{n\gamma/2}^{nD} \exp(-n/4)dy \\
    & \leq t_0 + \sum_{n=1}^{\delta/\beta} \frac{\exp\left(- \frac{t_0^2}{2 \gamma n} \right)}{2 \frac{t_0}{2 \gamma n}} + (D - \gamma/2)\sum_{n=1}^{\delta/\beta} n \exp(-n/4) \\
    & \leq t_0 + \frac{\gamma}{t_0}\sum_{n=1}^{\delta/\beta} n \exp\left(- \frac{t_0^2}{2 \gamma n} \right) + 16 D \\
    & \leq t_0 + \frac{\gamma}{t_0} \left( \frac{\delta}{\beta}\right)^2 \exp\left(- \frac{\beta}{2 \gamma}\frac{t_0^2}{\delta} \right) + 16 D.
\end{align*}
Then we have
\begin{align*}
     \frac{\gamma}{t_0} \left( \frac{\delta}{\beta}\right)^2 \exp\left(- \frac{\beta}{2 \gamma}\frac{t_0^2}{\delta} \right) & = \frac{\gamma}{\beta^2} \frac{\delta^2}{t_0} \exp\left(- \frac{\beta}{2 \gamma}\frac{t_0^2}{\delta} \right) \\
     & = \frac{\gamma}{A \beta^2} \frac{\delta^{3/2}}{\sqrt{\ln(\delta)}} \exp\left( - \frac{\beta}{2 \gamma}A^2 \ln(\delta) \right) \\
     & = \frac{\gamma}{A \beta^2} \frac{1}{\sqrt{\ln(\delta)}} \delta^{3/2 - \frac{\beta}{2 \gamma}A^2}.
\end{align*}
By taking $A = \sqrt{\frac{3 \gamma}{\beta}}$ we get by using $\beta \geq \frac{D}{4}$ and $\gamma \leq \frac{4}{(1 - \mu^{\dagger})^{5/2}}$
\begin{align*}
    = \frac{\sqrt{\gamma}}{\sqrt{3}\beta^{3/2}}\frac{1}{\sqrt{\ln(\delta)}} \leq \frac{1}{\sqrt{3}}\frac{2}{(1 - \mu^{\dagger})^{5/4}}\frac{4^{3/2}}{D^{3/2}}\frac{1}{1/4} \leq \frac{40}{(1 - \mu^{\dagger})^{5/4}D^{3/2}}.
\end{align*}
Finally we have
\begin{align*}
    \mathbb{E}_{\nu',\pi'}\left[\Big(\delta - N \hat D_{N}  + N \varepsilon c \Big)_+ \right]
     \leq& \ \mathbb{E}_{\nu',\pi'}\left[\Big(\delta - N D  \Big)_+ \right] + \frac{c}{\beta} \varepsilon \delta + \sqrt{\frac{3 \gamma}{\beta}}\sqrt{\delta \ln(\delta)}  \\
    &+ \frac{40}{(1 - \mu^{\dagger})^{5/4}D^{3/2}} + 16D + \frac{4128}{(1 - (\mu^{\dagger} - L_a))^{5} (D \wedge1)^3} + 16 D \\
     \leq&\   \mathbb{E}_{\nu',\pi'}\left[\Big(\delta - N D  \Big)_+ \right] + \frac{4 \varepsilon \delta}{(1 - \mu^{\dagger} )D} + \frac{7\sqrt{\delta \ln(\delta)}}{(1 - \mu^{\dagger})^{5/4}D^{1/2}} \\
     &+ \frac{40}{(1 - \mu^{\dagger})^{5/4}D^{3/2}} + \frac{4128}{(1- \mu^{\dagger})^{5} (D \wedge1)^3} + 32 D.
\end{align*}
\end{proof}

\subsection{Proof of Lemma \ref{lem:inter}}\label{sec:proofinter}

\begin{proof}[ of Lemma \ref{lem:inter}] If we note $\eta := \mu'_1 + L_1 - \mu^{\dagger} + \varepsilon > 0$, then we have 
\begin{align*}
     \mathbb{E}_{\nu',\pi'}\left[ \exp\left( N'_1  \mathcal{K}_{inf}(\hat \nu_1',\mu^{\dagger} - \varepsilon-L_1) \right)\right] & = \mathbb{E}_{\nu',\pi'}\left[ \exp\left( N'_1  \mathcal{K}_{inf}(\hat \nu_1',\mu'_1 - \eta) \right)\right].
\end{align*}
We want to use the second part of Lemma \ref{lem:Kinfepsilon} but we need that $\hat \mu'_1 < \mu'_1 - \eta$, then,
\begin{align*}
     &\mathbb{E}_{\nu',\pi'}\left[ \exp\left( N'_1  \mathcal{K}_{inf}(\hat \nu_1',\mu'_1 - \eta) \right)\right] \\
     & =  \mathbb{E}_{\nu',\pi'}\! \left[\mathbf{1}(\hat \mu'_1 < \mu'_1 - \eta) \exp\left( N'_1  \mathcal{K}_{inf}(\hat \nu_1',\mu'_1 - \eta) \right)\right] \\ & \phantom{=}+ \mathbb{E}_{\nu',\pi'}\! \left[\mathbf{1}(\hat \mu'_1 \geq \mu'_1 - \eta) \exp\left( N'_1  \underbrace{\mathcal{K}_{inf}(\hat \nu_1',\mu'_1 - \eta)}_{ = 0} \right)\right] \\
     & \leq \mathbb{E}_{\nu',\pi'}\left[\mathbf{1}(\hat \mu'_1 < \mu'_1 - \eta) \exp\left( N'_1  \mathcal{K}_{inf}(\hat \nu_1',\mu'_1 - \eta) \right)\right] + 1 \\
     & \leq \mathbb{E}_{\nu',\pi'}\left[\mathbf{1}(\hat \mu'_1 < \mu'_1 - \eta) \exp\left( N'_1 \left( \mathcal{K}_{inf}(\hat \nu_1',\mu'_1 ) - 2 \eta^2 \right)\right)\right] +1 \\
     & \leq \mathbb{E}_{\nu',\pi'}\left[\exp\left( N'_1 \left( \mathcal{K}_{inf}(\hat \nu_1',\mu'_1 ) - 2 \eta^2 \right)\right)\right] +1.
\end{align*}
Now we remove the dependency in $\pi'$ to be able to concentrate,
\begin{align*}
    \mathbb{E}_{\nu',\pi'}\left[\exp\left( N'_1 \left( \mathcal{K}_{inf}(\hat \nu_1',\mu'_1 ) - 2 \eta^2 \right)\right)\right] & \leq 1 + \sum_{s=1}^{+ \infty}\mathbb{E}_{\nu'}\left[\exp\left( s \left( \mathcal{K}_{inf}(\hat \nu_{1,s}',\mu'_1 ) - 2 \eta^2 \right)\right) \right],
\end{align*}
then we have
\begin{align*}
    &\mathbb{E}_{\nu'}\left[\exp\left( s \left( \mathcal{K}_{inf}(\hat \nu_{1,s}',\mu'_1 ) - 2 \eta^2 \right)\right) \right] = \int_{0}^{+ \infty} \mathbb{P}_{\nu'}\left(\exp\left(s \left( \mathcal{K}_{inf}(\hat \nu_{1,s}',\mu'_1 ) - 2 \eta^2 \right)\right) \geq x \right)dx.
\end{align*}
Using that $\mathcal{K}_{inf}(\hat \nu_{1,s}',\mu'_1 ) \leq \ln\left( 1/(1-\mu'_1) \right)$ we have a bound on $x$,
\begin{align*}
    & =  \int_{0}^{\exp\left(s \left( \ln(1/(1-\mu'_1)) - 2 \eta^2 \right)\right)} \mathbb{P}_{\nu'}\left(\exp\left(s \left( \mathcal{K}_{inf}(\hat \nu_{1,s}',\mu'_1 ) - 2 \eta^2 \right) \right) \geq x \right)dx \\
    & = \int_{0}^{\exp\left(s \left( \ln(1/(1-\mu'_1)) - 2 \eta^2 \right)\right)} \mathbb{P}_{\nu'}\left( \mathcal{K}_{inf}(\hat \nu_{1,s}',\mu'_1 ) \geq \frac{\ln x}{s} + 2 \eta^2 \right)dx.
\end{align*}
Now we upper bound the probability by 1 when $\frac{\ln x}{s} + 2 \eta^2 \leq 0$, it gives
\begin{align*}
    &\leq \exp\left(- s 2 \eta^2 \right) +   \int_{\exp\left(- s 2 \eta^2 \right)}^{\exp\left(s \left( \ln(1/(1-\mu'_1)) - 2 \eta^2 \right)\right)} \mathbb{P}_{\nu'}\left( \mathcal{K}_{inf}(\hat \nu_{1,s}',\mu'_1 ) \geq \frac{\ln x}{s} + 2 \eta^2 \right)dx.
\end{align*}
Using the change of variable $u = \ln(x)/s$, i.e. $x = e^{su}$, $dx = s e^{su} du$, we get
\begin{align*}
    & = \exp\left(- s 2 \eta^2 \right) +   \int_{-2\eta^2}^{\ln(1/(1-\mu'_1)) - 2\eta^2} \mathbb{P}_{\nu'}\left( \mathcal{K}_{inf}(\hat \nu_{1,s}',\mu'_1 ) \geq u + 2 \eta^2 \right) s e^{su} du.
\end{align*}
Using the concentration Lemma \ref{lem:concentre1} we get
\begin{align*}
    & \leq \exp\left(- s 2 \eta^2 \right) +   \int_{-2\eta^2}^{\ln(1/(1-\mu'_1)) - 2\eta^2} e(2s+1)e^{-s \left( u + 2 \eta^2 \right)} s e^{su} du \\
    & = \exp\left(- s 2 \eta^2 \right) + e(2s+1)s\exp\left(- s 2 \eta^2 \right)  \int_{-2\eta^2}^{\ln(1/(1-\mu'_1)) - 2\eta^2} du \\
    & = \exp\left(- s 2 \eta^2 \right) + e(2s+1)s\exp\left(- s 2 \eta^2 \right)  \ln(1/(1-\mu'_1)) \\
    & = \exp\left(- s 2 \eta^2 \right) \Bigg( 1 + e(2s+1)s  \ln(1/(1-\mu'_1))\Bigg) \\
    & \leq \exp\left(- s 2 \eta^2 \right) \Bigg( s^2 + 3(2s+s)s  \ln(1/(1-\mu'_1))\Big) \\
    & = s^2\exp\left(- s 2 \eta^2 \right) \Bigg( 1 +9  \ln(1/(1-\mu'_1))\Bigg). 
\end{align*}
Then we get
\begin{align*}
    \sum_{s=1}^{+ \infty}\mathbb{E}_{\nu'}\left[\exp\left( s \left( \mathcal{K}_{inf}(\hat \nu_{1,s}',\mu'_1 ) - 2 \eta^2 \right)\right) \right]
    & \leq \Big( 1 +9  \ln(1/(1-\mu'_1))\Big)\sum_{s=1}^{+ \infty} s^2\exp\left(- s 2 \eta^2 \right) \\
    & = \Big( 1 +9  \ln(1/(1-\mu'_1))\Big) \frac{e^{-2 \eta^2}\left( 1 + e^{-2 \eta^2} \right)}{\left( 1 - e^{-2 \eta^2} \right)^3} \\
    & \leq \Big( 1 +9  \ln(1/(1-\mu'_1))\Big) \frac{2}{\left( 1 - e^{-2 \eta^2} \right)^3}\\
    & = \frac{2 +18  \ln(1/(1-\mu'_1))}{\left( 1 - e^{-2 \eta^2} \right)^3}.
\end{align*}
Then finally 
\begin{align*}
\mathbb{E}_{\nu',\pi'}\left[ \exp\left( N'_1  \mathcal{K}_{inf}(\hat \nu_1',\mu'_1 - \eta) \right)\right] 
& \leq \mathbb{E}_{\nu',\pi'}\left[\exp\left( N'_1 \left( \mathcal{K}_{inf}(\hat \nu_1',\mu'_1 ) - 2 \eta^2 \right)\right)\right] +1 \\
& \leq  \sum_{s=1}^{+ \infty}\mathbb{E}_{\nu'}\left[\exp\left( s \left( \mathcal{K}_{inf}(\hat \nu_{1,s}',\mu'_1 ) - 2 \eta^2 \right)\right) \right] + 2 \\
& \leq \frac{ 2 +18  \ln(1/(1-\mu'_1))}{\left( 1 - e^{-2 \eta^2} \right)^3} + 2.
\end{align*}\vspace{-2em}
\end{proof}
\subsection{Proof of Lemma \ref{lem:inter2}}\label{sec:proofinter2}

\begin{proof}[of Lemma \ref{lem:inter2}]
    For all $y \in [c,a]$ we have
\begin{align*}
    - \frac{d }{dy}e^{-by^2} = 2bye^{-by^2} \geq 2bce^{-by^2}.
\end{align*}
Thus, we have
\begin{align*}
     \int_{c}^{a}e^{-by^2}dy & \leq  - \frac{1}{2bc}\int_{c}^{a}\frac{d }{dy}e^{-by^2} dy = \frac{1}{2bc} \left( e^{-bc^2} - e^{-ba^2} \right)  \leq \frac{e^{-bc^2}}{2bc}.
\end{align*}
\end{proof}

\section{Proof of Theorem \ref{thm:multisource}}\label{sec:proofmultisource}

The proof is very similar to the one of Theorem \ref{thm:general}. We however have to adapt some lemmas. The first one is the analog of Lemma \ref{lem:decomposition}:
\begin{lemma}\label{lem:decomposition2}
    For all $\nu,\nu'$, for all $\mu^{\dagger}$ such that
\begin{equation*}
    \mu^{\dagger} > \mu_a, \quad \mu^{\dagger} \leq \mu_1, \quad \mu^{\dagger} \leq \min_{i \in \{2,\ldots,M\}} \mu_1^{(i)} + L_1^{(i)},
\end{equation*}
for every $\varepsilon$ such that
\begin{equation*}
    \varepsilon > 0, \quad \varepsilon < \mu^{\dagger}, \quad \varepsilon \leq \frac{(1 - \mu^{\dagger})\mathcal{K}_{inf}(\nu_a,\mu^{\dagger})}{4},
\end{equation*}
for all $T > 3$, with \texttt{KL-UCB-Transfer}, for every suboptimal arm $a$ we have
    \begin{align*}
       N_a(T) & \leq \frac{\left(\ln T - \sum_{i \in \mathcal{S}_a^\dagger} N_a^{(i)}  \mathcal{K}_{inf}\big(\hat\nu_a^{(i)}, \mu^\dagger - \varepsilon - L_a^{(i)}\big)\right)_+}{\mathcal{K}_{inf}(\nu_a,\mu^{\dagger}) } + \frac{\ln \ln T}{\mathcal{K}_{inf}(\nu_a,\mu^{\dagger})} +  \frac{4 \varepsilon (\ln T + \ln \ln T)}{\mathcal{K}_{inf}(\nu_a,\mu^{\dagger})^2(1 - \mu^{\dagger})} \\
    & + \sum_{s = 1}^{+ \infty}\mathbf{1}\left(\mathcal{K}_{inf}(\hat \nu_{a,s},\mu^{\dagger}) \leq \mathcal{K}_{inf}(\nu_a,\mu^{\dagger}) - \frac{ \varepsilon}{1 - \mu^{\dagger}} \right) + K \\
    &+ \sum_{i=1}^{M} \sum_{t = K+1}^{T} \sum_{s=1}^{+ \infty} \mathbf{1}\left( \mathcal{K}_{inf}(\hat \nu^{(i)}_{1,s}, \mu_1^{(i)}) > \frac{1}{s}\left( \delta_t - \sum_{j=i+1}^{M}N_1^{(j)}\mathcal{K}_{inf}(\hat \nu^{(j)}_{1}, \mu^{\dagger} - \varepsilon - L_1^{(j)}) \right) + 2 \varepsilon^2 \right). 
    \end{align*}
\end{lemma}
(see proof of Lemma \ref{lem:decomposition2} in \autoref{sec:proofdecomposition2}). \\
Now if we note
\begin{equation*}
    \mathcal{A}^{(i)} := \sum_{t = K+1}^{T} \sum_{s=1}^{+ \infty} \mathbf{1}\left( \mathcal{K}_{inf}(\hat \nu^{(i)}_{1,s}, \mu_1^{(i)}) > \frac{1}{s}\left( \delta_t - \sum_{j=i+1}^{M}N_1^{(j)}\mathcal{K}_{inf}(\hat \nu^{(j)}_{1}, \mu^{\dagger} - \varepsilon - L_1^{(j)}) \right) + 2 \varepsilon^2 \right), 
\end{equation*}
then we have the following lemma which is an analogue of Lemma \ref{lem:terme2} and \ref{lem:terme3}:

\begin{lemma}\label{lem:terme22}
With the same assumption as the Lemma \ref{lem:decomposition2} with also $\varepsilon \leq 1$ and $\eta^{(j)} := \mu^{(j)}_1 + L^{(j)}_1 - \mu^{\dagger} + \varepsilon > 0$ for all $j > i$,
    \begin{equation*}
        \mathbb{E}\left[ \mathcal{A}^{(i)} \right] \leq \left(\prod_{j = i+1}^{M} \left( \frac{ 2 +18  \ln(1/(1-\mu^{(j)}_1))}{\left( 1 - e^{-2 (\eta^{(j)})^2} \right)^3} + 2 \right)  \right) \frac{81(2 + \ln \ln T)}{\varepsilon^4}.
    \end{equation*}
\end{lemma}
(see proof of Lemma \ref{lem:terme22} in \autoref{sec:proofterme22}). \\
And finally the analogue of Lemma \ref{lem:terme4}:
\begin{lemma}\label{lem:terme42}
    For $\varepsilon \leq \min_{i \in \mathcal{S}^{\dagger}_a}\frac{( 1 - (\mu^{\dagger} - L^{(i)}_a))\mathcal{K}_{inf}(\nu^{(i)}_{a},\mu^{\dagger} - L^{(i)}_a) }{4}$,
\begin{align*}
    &\mathbb{E}_{(\pi),(\nu)}\left[\left(\ln T - \sum_{i \in \mathcal{S}_a^\dagger} N_a^{(i)}  \mathcal{K}_{inf}\big(\hat\nu_a^{(i)}, \mu^\dagger - \varepsilon - L_a^{(i)}\big)\right)_+ \right] \\
    & \leq \mathbb{E}_{(\pi),(\nu)}\left[\left(\ln T - \sum_{i \in \mathcal{S}_a^\dagger} N_a^{(i)}  \mathcal{K}_{inf}\big(\nu_a^{(i)}, \mu^\dagger - L_a^{(i)}\big)\right)_+ \right] \\ 
    &+ \sum_{i \in \mathcal{S}^{\dagger}_a} \Bigg(\frac{4 \varepsilon \ln T}{(1 - \mu^{\dagger})\mathcal{K}_{inf}(\nu^{(i)}_{a},\mu^{\dagger} - L^{(i)}_a)} + \frac{7\sqrt{\ln(T) \times \ln(\ln T)}}{(1 - \mu^{\dagger})^{5/4}\mathcal{K}_{inf}(\nu^{(i)}_{a},\mu^{\dagger} - L^{(i)}_a)^{1/2}}  \\
    & + \frac{10}{(1 - \mu^{\dagger})^{5/4}\mathcal{K}_{inf}(\nu^{(i)}_{a},\mu^{\dagger} - L^{(i)}_a)^{3/2}} + \frac{4128}{(1 - \mu^{\dagger})^{5} (\mathcal{K}_{inf}(\nu^{(i)}_{a},\mu^{\dagger} - L^{(i)}_a) \wedge1)^3} + 32 \mathcal{K}_{inf}(\nu^{(i)}_{a},\mu^{\dagger} - L^{(i)}_a) \Bigg).
\end{align*}
\end{lemma}
(see proof of Lemma \ref{lem:terme42} in \autoref{sec:proofterme42}). \\
From those lemmas the proof of Theorem \ref{thm:multisource} is trivial. We get that $g_{(\nu),\mu^\dagger,(L)}(T)$ is equal to
\begin{align*}
    &\sum_{i \in \mathcal{S}^{\dagger}_a} \Bigg(\frac{4 \varepsilon \ln T}{(1 - \mu^{\dagger})\mathcal{K}_{inf}(\nu^{(i)}_{a},\mu^{\dagger} - L^{(i)}_a)} + \frac{7\sqrt{\ln(T) \times \ln(\ln T)}}{(1 - \mu^{\dagger})^{5/4}\mathcal{K}_{inf}(\nu^{(i)}_{a},\mu^{\dagger} - L^{(i)}_a)^{1/2}}  \\
    & + \frac{10}{(1 - \mu^{\dagger})^{5/4}\mathcal{K}_{inf}(\nu^{(i)}_{a},\mu^{\dagger} - L^{(i)}_a)^{3/2}} + \frac{4128}{(1 - \mu^{\dagger})^{5} (\mathcal{K}_{inf}(\nu^{(i)}_{a},\mu^{\dagger} - L^{(i)}_a) \wedge1)^3} + 32 \mathcal{K}_{inf}(\nu^{(i)}_{a},\mu^{\dagger} - L^{(i)}_a) \Bigg) \\
    & +\frac{\ln \ln T}{\mathcal{K}_{inf}(\nu_a,\mu^{\dagger})} +  \frac{4 \varepsilon (\ln T + \ln \ln T)}{\mathcal{K}_{inf}(\nu_a,\mu^{\dagger})^2(1 - \mu^{\dagger})} + K + 10 + \frac{16}{\varepsilon^2(1 - \mu^{\dagger})^{1/2}} \\
    & + \sum_{i=1}^{M} \left(\prod_{j = i+1}^{M} \left( \frac{ 2 +18  \ln(1/(1-\mu^{(j)}_1))}{\left( 1 - e^{-2 (\eta^{(j)})^2} \right)^3} + 2 \right)  \right) \frac{81(2 + \ln \ln T)}{\varepsilon^4},  
\end{align*}
with
\begin{equation*}
    \varepsilon := \left(1 \wedge \mu^{\dagger} \wedge \frac{(1 - \mu^{\dagger})\mathcal{K}_{inf}(\nu_a,\mu^{\dagger})}{4} \wedge \min_{i \in \mathcal{S}^{\dagger}_a}\frac{( 1 - (\mu^{\dagger} - L^{(i)}_a))\mathcal{K}_{inf}(\nu^{(i)}_{a},\mu^{\dagger} - L^{(i)}_a) }{4} \right)\left( \frac{\ln \ln T}{\ln T} \right)^{1/5}.
\end{equation*}

\subsection{Proof of Lemma \ref{lem:decomposition2}}\label{sec:proofdecomposition2}

The proof is very similar to Lemma \ref{lem:decomposition}, we will often refer to it.

\begin{proof}[of Lemma \ref{lem:decomposition2}]
As in the proof of Lemma \ref{lem:decomposition}, we have

    \begin{align*}
        N_a(T)  \leq K + \underbrace{\sum_{t=K+1}^T \mathbf{1}\left(U_1(t) < \mu^{\dagger} - \varepsilon \right)}_{A} + \underbrace{\sum_{t=K+1}^T \mathbf{1}\left(U_a(t) \geq \mu^{\dagger} - \varepsilon\ \cap \ A_t = a \right)}_{B}.
    \end{align*}
Using exactly the same steps as in Lemma \ref{lem:decomposition} with 
\begin{equation*}
    \Gamma := \frac{\left(\delta_T - \sum_{i =2}^{M} N_a^{(i)}  \mathcal{K}_{inf}\big(\hat\nu_a^{(i)}, \mu^\dagger - \varepsilon - L_a^{(i)}\big)\right)_+}{\mathcal{K}_{inf}(\nu_a,\mu^{\dagger}) - \frac{2 \varepsilon}{1 - \mu^{\dagger}}},
\end{equation*}
we get
\begin{align*}
    B &\leq \Gamma + \sum_{s = 1}^{+ \infty}\mathbf{1}\left(\mathcal{K}_{inf}(\hat \nu_{a,s},\mu^{\dagger} - \varepsilon) \leq \mathcal{K}_{inf}(\nu_a,\mu^{\dagger}) - \frac{2 \varepsilon}{1 - \mu^{\dagger}} \right), 
\end{align*}
and
\begin{align*}
    \Gamma &\leq \frac{\left(\ln T - \sum_{i \in \mathcal{S}_a^\dagger} N_a^{(i)}  \mathcal{K}_{inf}\big(\hat\nu_a^{(i)}, \mu^\dagger - \varepsilon - L_a^{(i)}\big)\right)_+}{\mathcal{K}_{inf}(\nu_a,\mu^{\dagger}) } + \frac{\ln \ln T}{\mathcal{K}_{inf}(\nu_a,\mu^{\dagger})} +  \frac{4 \varepsilon (\ln T + \ln \ln T)}{\mathcal{K}_{inf}(\nu_a,\mu^{\dagger})^2(1 - \mu^{\dagger})}.
\end{align*}
Now for $A$,
\begin{align*}
    A&=\sum_{t=K+1}^T \mathbf{1}\left(U_1(t) < \mu^{\dagger} - \varepsilon \right) \\
    & \leq \sum_{t=K+1}^T \mathbf{1}\left(N_1(t)\mathcal{K}_{inf}(\hat \nu^{(1)}_1(t),\mu^{\dagger} - \varepsilon) + \sum_{i=2}^{M}N^{(i)}_1  \mathcal{K}_{inf}(\hat \nu^{(i)}_1,\mu^{\dagger} - \varepsilon-L^{(i)}_1) > \delta_t \right) \\
    & = \sum_{t=K+1}^T \mathbf{1}\left(N_1(t)\mathcal{K}_{inf}(\hat \nu^{(1)}_1(t),\mu^{\dagger} - \varepsilon)   > \delta_t - \sum_{i=2}^{M}N^{(i)}_1  \mathcal{K}_{inf}(\hat \nu^{(i)}_1,\mu^{\dagger} - \varepsilon-L^{(i)}_1) \right).
\end{align*}
Now if we note 
\begin{equation*}
    G^t_{j} := \delta_t - \sum_{i=j+1}^{M}N^{(i)}_1  \mathcal{K}_{inf}(\hat \nu^{(i)}_1,\mu^{\dagger} - \varepsilon-L^{(i)}_1),
\end{equation*}
we have in particular that $G^t_{j} = G^t_{j+1} - N^{(j+1)}_1\mathcal{K}_{inf}(\hat \nu^{(j+1)}_1,\mu^{\dagger} - \varepsilon-L^{(j+1)}_1)$ and $G^t_{M} = \delta_t$ and
\begin{equation*}
    A \leq \sum_{t=K+1}^T \mathbf{1}\left(N_1(t)\mathcal{K}_{inf}(\hat \nu^{(1)}_1(t),\mu^{\dagger} - \varepsilon)   >  G^t_{1} \right).
\end{equation*}
Now we want to use Lemma \ref{lem:noproof} but it requires to have $G^t_{1}> 0$, then we will split according to this event:
\begin{align*}
    &\mathbf{1}\left(N_1(t)\mathcal{K}_{inf}(\hat \nu^{(1)}_1(t),\mu^{\dagger} - \varepsilon)   >  G^t_{1} \right) \\
    & \leq \mathbf{1}\left(G^t_1 > 0 \cap N_1(t)\mathcal{K}_{inf}(\hat \nu^{(1)}_1(t),\mu^{\dagger} - \varepsilon)   >  G^t_{1} \right) + \mathbf{1}\left(G^t_1 \leq 0 \right) \\
    & \leq \mathbf{1}\left(G^t_1 > 0 \cap N_1(t)\mathcal{K}_{inf}(\hat \nu^{(1)}_1(t),\mu^{(1)}_1 )   >  G^t_{1} + 2 \varepsilon^2 N_1(t) \right) + \mathbf{1}\left(G^t_1 \leq 0 \right) \\
    & \leq \mathbf{1}\left( N_1(t)\mathcal{K}_{inf}(\hat \nu^{(1)}_1(t),\mu^{(1)}_1 )   >  G^t_{1} + 2 \varepsilon^2 N_1(t) \right) + \mathbf{1}\left(G^t_1 \leq 0 \right) \\
    & = \mathbf{1}\left( N_1(t)\mathcal{K}_{inf}(\hat \nu^{(1)}_1(t),\mu^{(1)}_1 )   >  G^t_{1} + 2 \varepsilon^2 N_1(t) \right) + \mathbf{1}\left(N^{(2)}_1\mathcal{K}_{inf}(\hat \nu^{(2)}_1,\mu^{\dagger} - \varepsilon - L^{(2)}_1)   \geq  G^t_{2} \right). 
\end{align*}
Using the same reasoning, we have for all $2 \leq i< M$ that
\begin{align*}
    &\mathbf{1}\left(N^{(i)}_1\mathcal{K}_{inf}(\hat \nu^{(i)}_1,\mu^{\dagger} - \varepsilon - L^{(i)}_1)   \geq  G^t_{i} \right) \\
    & \leq \mathbf{1}\left( N^{(i)}_1 \geq 1 \cap N^{(i)}_1\mathcal{K}_{inf}(\hat \nu^{(i)}_1,\mu^{(i)}_1 )   \geq  G^t_{i} + 2 \varepsilon^2 N^{(i)}_1\right) + \mathbf{1}\left(N^{(i+1)}_1\mathcal{K}_{inf}(\hat \nu^{(i+1)}_1,\mu^{\dagger} \!-\! \varepsilon \!-\! L^{(i+1)}_1)   \geq  G^t_{i+1} \right). 
\end{align*}
Then we have
\begin{align*}
     &\mathbf{1}\left(N_1(t)\mathcal{K}_{inf}(\hat \nu^{(1)}_1(t),\mu^{\dagger} - \varepsilon)   >  G^t_{1} \right) \\
     & \leq \mathbf{1}\!\left(N^{(M)}_1\mathcal{K}_{inf}(\hat \nu^{(M)}_1,\mu^{\dagger} \!-\! \varepsilon \!-\! L^{(M)}_1)   \geq  G^t_{M} \right) + \sum_{i=1}^{M-1}\!\mathbf{1}\!\left( N^{(i)}_1 \geq 1 \cap N^{(i)}_1\mathcal{K}_{inf}(\hat \nu^{(i)}_1,\mu^{(i)}_1 )   \geq  G^t_{i} + 2 \varepsilon^2 N^{(i)}_1 \right).
\end{align*}
However we have $G^t_{M} = \delta_t > 0$, then we can directly use Lemma \ref{lem:noproof}, which gives
\begin{equation*}
    \mathbf{1}\!\left(N^{(M)}_1\mathcal{K}_{inf}(\hat \nu^{(M)}_1,\mu^{\dagger} \!-\! \varepsilon \!-\! L^{(M)}_1)   \geq  G^t_{M} \right) \leq \mathbf{1}\!\left(N^{(M)}_1 \!\geq 1 \cap N^{(M)}_1\mathcal{K}_{inf}(\hat \nu^{(M)}_1,\mu^{(M)}_1)   \geq  G^t_{M} + 2 \varepsilon^2N^{(M)}_1 \right), 
\end{equation*}
then we have
\begin{align*}
    A & \leq \sum_{t=K+1}^T \sum_{i=1}^{M}\mathbf{1}\left( N^{(i)}_1 \geq 1 \cap N^{(i)}_1\mathcal{K}_{inf}(\hat \nu^{(i)}_1,\mu^{(i)}_1 )   \geq  G^t_{i} + 2 \varepsilon^2 N^{(i)}_1 \right).
\end{align*}
Then we decompose according to the value of the $N^{(i)}_1$ which gives
\begin{align*}
    A & \leq \sum_{t=K+1}^T \sum_{i=1}^{M} \sum_{s = 1}^{+ \infty}\mathbf{1}\left(\mathcal{K}_{inf}(\hat \nu^{(i)}_{1,s},\mu^{(i)}_1 )   \geq  \frac{1}{s}G^t_{i} + 2 \varepsilon^2 \right) \\
    & = \sum_{i=1}^{M} \sum_{t=K+1}^T \sum_{s = 1}^{+ \infty}\mathbf{1}\left( \mathcal{K}_{inf}(\hat \nu^{(i)}_{1,s}, \mu_1^{(i)}) \geq \frac{1}{s}\left( \delta_t - \sum_{j=i+1}^{M}N_1^{(j)}\mathcal{K}_{inf}(\hat \nu^{(j)}_{1}, \mu^{\dagger} - \varepsilon - L_1^{(j)}) \right) + 2 \varepsilon^2 \right). 
\end{align*}

\end{proof}

\subsection{Proof of Lemma \ref{lem:terme22}}\label{sec:proofterme22}

\begin{proof}[of Lemma \ref{lem:terme22}]
Let us note $ h_{i+1,t} :=  \delta_t - \sum_{j=i+1}^{M}N_1^{(j)}\mathcal{K}_{inf}(\hat \nu^{(j)}_{1}, \mu^{\dagger} - \varepsilon - L_1^{(j)})$. We have
\begin{align*}
    &\mathbb{E}_{(\pi),(\nu),\nu,\pi}\left[\mathcal{A}^{(i)} \right]  = \mathbb{E}_{(\pi),(\nu),\nu,\pi}\!\left[\sum_{t = K+1}^{T} \sum_{s=1}^{+ \infty}\mathbb{P}_{(\pi),(\nu),\nu,\pi}\!\left( \mathcal{K}_{inf}(\hat \nu^{(i)}_{1,s}, \mu_1^{(i)}) > \frac{h_{i+1,t}}{s} \!+\! 2 \varepsilon^2\! \ \middle | \!\ (N^{(j)}_1,\hat \nu^{(j)}_{1})_{j \geq i+1} \right)\! \right]\!.
\end{align*}
Using the assumption of independence we can do the computation conditionally on $h_{i+1,t}$. Then using the same reasoning as in the proof of Lemma \ref{lem:terme2} we get
\begin{align*}
    &\mathbb{E}_{(\pi),(\nu),\nu,\pi}\left[\mathcal{A}^{(i)} \right] \\
    & \leq \mathbb{E}_{(\pi),(\nu),\nu,\pi}\left[\sum_{t = K+1}^{T} \sum_{s=1}^{+ \infty} e(2s+1)e^{-s \left( \frac{h_{i+1,t}}{s} + 2 \varepsilon^2 \right)} \right] \\
    & = \mathbb{E}_{(\pi),(\nu)}\left[\exp\left(\sum_{j=i+1}^{M}N_1^{(j)}\mathcal{K}_{inf}(\hat \nu^{(j)}_{1}, \mu^{\dagger} - \varepsilon - L_1^{(j)})\right)\right]\left(\sum_{t=K+1}^T e^{- \delta_t} \right)  \left(\sum_{s = 1}^{+ \infty} e(2s+1)e^{-s2 \varepsilon^2}\right) \\
    & \leq \mathbb{E}_{(\pi),(\nu)}\left[\exp\left(\sum_{j=i+1}^{M}N_1^{(j)}\mathcal{K}_{inf}(\hat \nu^{(j)}_{1}, \mu^{\dagger} - \varepsilon - L_1^{(j)})\right)\right] \frac{81(2 + \ln \ln T)}{\varepsilon^4}.
\end{align*}
Now using again the independence and the Lemma \ref{lem:inter} we get
\begin{align*}
    &\mathbb{E}_{(\pi),(\nu)}\left[\exp\left(\sum_{j=i+1}^{M}N_1^{(j)}\mathcal{K}_{inf}(\hat \nu^{(j)}_{1}, \mu^{\dagger} - \varepsilon - L_1^{(j)})\right)\right] \\
    & = \prod_{j=i+1}^{M}\mathbb{E}_{\pi^{(j)},\nu^{(j)}}\left[\exp\left(N_1^{(j)}\mathcal{K}_{inf}(\hat \nu^{(j)}_{1}, \mu^{\dagger} - \varepsilon - L_1^{(j)})\right)\right] \\
    & \leq \prod_{j = i+1}^{M} \left( \frac{ 2 +18  \ln(1/(1-\mu^{(j)}_1))}{\left( 1 - e^{-2 (\eta^{(j)})^2} \right)^3} + 2 \right).
\end{align*}\vspace{-2em}
\end{proof}

\subsection{Proof of Lemma \ref{lem:terme42}}\label{sec:proofterme42}

\begin{proof}[of Lemma \ref{lem:terme42}]
   Using Lemma \ref{lem:Kinfepsilon} we get
\begin{align*}
    &\mathbb{E}_{(\pi),(\nu)}\left[\left(\ln T - \sum_{i \in \mathcal{S}_a^\dagger} N_a^{(i)}  \mathcal{K}_{inf}\big(\hat\nu_a^{(i)}, \mu^\dagger - \varepsilon - L_a^{(i)}\big)\right)_+ \right] \\
    &\leq \mathbb{E}_{(\pi),(\nu)}\left[\left(\ln T - \sum_{i \in \mathcal{S}_a^\dagger} N_a^{(i)}  \mathcal{K}_{inf}\big(\hat\nu_a^{(i)}, \mu^\dagger - \varepsilon - L_a^{(i)}\big) + N^{(i)}_a \frac{\varepsilon}{1 - (\mu^{\dagger} - L^{(i)}_a} \right)_+ \right].
\end{align*}
Now we introduce some notations to reduce the length of the equations:
\begin{multicols}{2}
\begin{itemize}
    \item $\delta := \ln T$
    \item $\hat D^{(i)}_s := \mathcal{K}_{inf}(\hat \nu^{(i)}_{a,s},\mu^{\dagger} - L^{(i)}_a)$
    \item $D^{(i)} := \mathcal{K}_{inf}(\nu^{(i)}_{a},\mu^{\dagger} - L^{(i)}_a)$
    \item $N^{(i)} := N^{(i)}_a$
    \item $c^{(i)} := \frac{1}{1 - (\mu^{\dagger} - L^{(i)}_a)} \leq \frac{1}{1 - \mu^{\dagger}}$
    \item $ \beta^{(i)} := \frac{D^{(i)} - \varepsilon c^{(i)}}{2} \geq \frac{D^{(i)}}{4}$.
\end{itemize}
\end{multicols}
\noindent
We split between two events
\begin{equation}\label{eq:doublevent2}
    \left\{ \forall i \in \mathcal{S}_a^\dagger, \  N^{(i)} < \frac{\delta}{\beta^{(i)}} \right\}, \qquad \left\{ \exists i \in \mathcal{S}_a^\dagger, \ N^{(i)} \geq \frac{\delta}{\beta^{(i)}} \right\}.
\end{equation}
on this first event we have
\begin{align*}
    &\mathbb{E}_{(\nu),(\pi)}\left[\mathbf{1}\left(  \forall i \in \mathcal{S}_a^\dagger, \  N^{(i)} < \frac{\delta}{\beta^{(i)}}\right)\Big(\delta - \sum_{i \in \mathcal{S}_a^\dagger}N^{(i)} \hat D^{(i)}_{N^{(i)}}  +N^{(i)} \varepsilon c^{(i)} \Big)_+ \right] \\
    & \leq \mathbb{E}_{(\nu),(\pi)}\left[\mathbf{1}\left(  \forall i \in \mathcal{S}_a^\dagger, \  N^{(i)} < \frac{\delta}{\beta^{(i)}}\right)\Big(\delta - \sum_{i \in \mathcal{S}_a^\dagger}N^{(i)} \hat D^{(i)}_{N^{(i)}}  +\frac{\delta}{\beta^{(i)}} \varepsilon c^{(i)} \Big)_+ \right] \\
    & \leq \mathbb{E}_{(\nu),(\pi)}\left[\mathbf{1}\left(  \forall i \in \mathcal{S}_a^\dagger, \  N^{(i)} < \frac{\delta}{\beta^{(i)}}\right)\Big(\delta - \sum_{i \in \mathcal{S}_a^\dagger}N^{(i)} \hat D^{(i)}_{N^{(i)}}   \Big)_+ \right] + \delta \varepsilon\sum_{i \in \mathcal{S}_a^\dagger} \frac{c^{(i)}}{\beta^{(i)}}.
\end{align*}
Then we have
\begin{align*}
     &\mathbb{E}_{(\nu),(\pi)}\left[\mathbf{1}\left(  \forall i \in \mathcal{S}_a^\dagger, \  N^{(i)} < \frac{\delta}{\beta^{(i)}}\right)\Big(\delta - \sum_{i \in \mathcal{S}_a^\dagger}N^{(i)} \hat D^{(i)}_{N^{(i)}}   \Big)_+ \right] \\
     & \leq \Big(\delta - \sum_{i \in \mathcal{S}_a^\dagger}N^{(i)}  D^{(i)}   \Big)_+ \\
     &+ \mathbb{E}_{(\nu),(\pi)}\left[\mathbf{1}\left(  \forall i \in \mathcal{S}_a^\dagger, \  N^{(i)} < \frac{\delta}{\beta^{(i)}}\right)\Bigg(\Big(\delta - \sum_{i \in \mathcal{S}_a^\dagger}N^{(i)} \hat D^{(i)}_{N^{(i)}}   \Big)_+ - \Big(\delta - \sum_{i \in \mathcal{S}_a^\dagger}N^{(i)}  D^{(i)}   \Big)_+ \Bigg)_+ \right].
\end{align*}
However we have
\begin{align*}
    &\mathbf{1}\left(  \forall i \in \mathcal{S}_a^\dagger, \  N^{(i)} < \frac{\delta}{\beta^{(i)}}\right)\Bigg(\Big(\delta - \sum_{i \in \mathcal{S}_a^\dagger}N^{(i)} \hat D^{(i)}_{N^{(i)}}   \Big)_+ - \Big(\delta - \sum_{i \in \mathcal{S}_a^\dagger}N^{(i)}  D^{(i)}   \Big)_+ \Bigg)_+ \\
    & \leq \mathbf{1}\left(  \forall i \in \mathcal{S}_a^\dagger, \  N^{(i)} < \frac{\delta}{\beta^{(i)}}\right) \sum_{i \in \mathcal{S}_a^\dagger}N^{(i)} \left(D^{(i)} - D^{(i)}_{N^{(i)}} \right)_+ \\
    & \leq \sum_{i \in \mathcal{S}_a^\dagger}\mathbf{1}\left( N^{(i)} < \frac{\delta}{\beta^{(i)}}\right)N^{(i)} \left(D^{(i)} - D^{(i)}_{N^{(i)}} \right)_+ \\
    & = \sum_{i \in \mathcal{S}_a^\dagger}\mathbf{1}\left( N^{(i)} < \frac{\delta}{\beta^{(i)}}\right)\int_{0}^{N^{(i)}D^{(i)}}\mathbf{1}\left(D^{(i)}_{N^{(i)}} <D^{(i)} - \frac{y}{N^{(i)}}  \right)dy,
\end{align*}
and we are back to the same proof as in Lemma \ref{lem:terme4}.\\
On the second event \eqref{eq:doublevent2}, we have
\begin{align*}
    &\mathbb{E}_{(\nu),(\pi)}\left[\mathbf{1}\left(\exists i \in \mathcal{S}_a^\dagger, \ N^{(i)} \geq \frac{\delta}{\beta^{(i)}}\right)\Big(\delta - \sum_{i \in \mathcal{S}_a^\dagger}N^{(i)} \hat D^{(i)}_{N^{(i)}}  +N^{(i)} \varepsilon c^{(i)} \Big)_+ \right] \\
    &  \leq \sum_{j \in \mathcal{S}_a^\dagger}\mathbb{E}_{(\nu),(\pi)}\left[\mathbf{1}\left(  N^{(j)} \geq \frac{\delta}{\beta^{(j)}}\right)\Big(\delta - \sum_{i \in \mathcal{S}_a^\dagger}N^{(i)} \hat D^{(i)}_{N^{(i)}}  +N^{(i)} \varepsilon c^{(i)} \Big)_+ \right] \\
    & \leq \sum_{j \in \mathcal{S}_a^\dagger}\mathbb{E}_{\nu^{(j)},\pi^{(j)}}\left[\mathbf{1}\left(  N^{(j)} \geq \frac{\delta}{\beta^{(j)}}\right)\Big(\delta - N^{(j)} \hat D^{(j)}_{N^{(j)}}  +N^{(j)} \varepsilon c^{(j)} \Big)_+ \right], 
\end{align*}
we conclude by controlling each term with the same reasoning as in the proof of Lemma \ref{lem:terme4}.
\end{proof}

\section{Proof of Proposition \ref{prop:matching}}\label{sec:proofmatching}

\begin{proof}[ of Proposition \ref{prop:matching}] When $\mu'_a > \mu_1 - L_a$ we already have the same bound. Let $\mu'_a < \mu_1 - L_a$. By looking at both theorems we can see that it is sufficient to prove :
\begin{equation*}
    \mathbb{E}_{\nu',\pi'}\big[\big(\ln T - N'_a\,\mathcal{K}_{inf}(\nu'_a,\mu_1-L_a)\big)_+\big]\leq \left( \ln T - \mathbb{E}_{\pi',\nu'}[N'_a]\mathcal{K}_{inf}(\nu'_a,\mu_1 - L_a) \right)_+ + o(\ln T)
\end{equation*}
Let us introduce some notations
\begin{multicols}{2}
    \begin{itemize}
    \item $D := \mathcal{K}_{inf}(\nu'_a,\mu_1 - L_a)$
    \item $N := N'_a$
    \item $ \mathbb{E}_{\pi',\nu'}[N'_a] := \bar N$
    \item $\delta := \ln T$
    \item $g(n) := (\delta - nD)_+$.
\end{itemize}
\end{multicols}
\noindent
Then we want to prove that $ \mathbb{E}\big[\big(\delta - N D\big)_+\big]\leq \left( \delta - \bar N D \right)_+ + o(\delta)$.
\\
Using that $(\cdot)_+$ is $1$-Lipschitz we have that $g$ is $D$-Lipschitz. Indeed, 
\begin{align*}
    | g(n) - g(m)| = |(\delta - nD)_+ - (\delta - mD)_+ |  \leq |(\delta - nD) - (\delta - mD) |  = D  | n - m|.
\end{align*}
Then
\begin{align*}
     \mathbb{E}\big[\big(\delta - N  D\big)_+\big] - \left( \delta - \bar N D \right)_+ & = \mathbb{E}[g(N)] - g(\bar N)\big]  \leq \mathbb{E}[ |g(N) - g(\bar N)|]\ \leq D \ \mathbb{E}[ |N - \bar N|]. 
\end{align*}
\end{proof}

\section{Proof of Theorem \ref{thm:lowerbound}}\label{sec:prooflowerbound}

To prove it we need to redefine the transfer bandit as a single bandit game, which is almost the same as the classical bandit game.
\\
\\
For $t \in \{1,...,T'\} $, the game is on the distribution $\nu'$ with the strategy $\pi'$ while the second part $t \in \{T'+1,...,T' + T\}$ the game is on the distribution $\nu$ with the strategy $\pi$.
\\
\\
We set $N := T' + T$. We consider a sequence of iid variable $(U_n)_n$ uniform on $[0,1]$, independent from the rest, which is the source of randomness for both phases. Now for each turn $t$ we define the information available at turn $t$ as 
\begin{equation*}
    I_t := (U_1,Y_1,...,Y_{t-1},U_t),
\end{equation*}
and the sequence of strategy
\begin{equation*}
    \pi^{\dagger}_t := \pi' \text{ for } t \leq T', \qquad \pi^{\dagger}_t := \pi \text{ for } t > T',
\end{equation*}
such that at round $t$ we have $A_t = \pi^{\dagger}_t(I_t)$. We can note that for $t > T'$, $I_t$ contains all the history of the first phase. We note $Y_t$ the observation of the round $t$:
\begin{equation*}
    Y_t \sim \nu'_{A_t} \text{ for } t \leq T', \qquad Y_t \sim \nu_{A_t} \text{ for } t > T'.
\end{equation*}
Formally, $Y_t$ for $t\le T'$ (resp.\ $t>T'$) is read from an i.i.d.\ table $X'_{A_t}(\cdot)\sim\nu'_{A_t}$ (resp.\ $X_{A_t}(\cdot)\sim\nu_{A_t}$). We suppose that $U_{t+1}$ remains independent of $(I_t,Y_t)$ for every $t$.
\\
\\
We will note $P^{I_t}_{(\nu,\nu'), \pi^{\dagger}}$ the distribution of $I_{t}$ under $(\nu',\pi',\nu,\pi)$ and
\begin{equation*}
    N'_k := \sum_{t = 1}^{T'} \mathbf{1}\left(A_t = k \right), \qquad N_k(T) := \sum_{t = T'+1}^{N} \mathbf{1}\left(A_t = k \right).
\end{equation*}
Now we have the following lemma
\begin{lemma}\label{lem:jointure}
Taking an alternative instance $(\tilde \nu ,\tilde\nu')$ such that $\nu \ll \tilde \nu$ and $ \nu' \ll \tilde \nu'$ and using the same notation introduced before we have
$$\KL\left(P^{I_{N+1}}_{( \nu,\nu'), \pi^{\dagger}},P^{I_{N+1}}_{(\tilde \nu,\tilde\nu'), \pi^{\dagger}}\right)  =  \sum_{i=1}^K \mathbb{E}_{\nu', \pi',\nu,\pi }[N_i(T) ]\KL\left(\nu_i, \tilde{\nu}_i\right) + \sum_{i=1}^K \mathbb{E}_{\nu', \pi' }[N'_i ]\KL\left(\nu'_i, \tilde{\nu}'_i\right).$$
\end{lemma}
(see proof of Lemma \ref{lem:jointure} in \autoref{sec:proofjointure}).\\
The proof is using the same argument as in the classical case (see \cite{garivier2019explore} or  \cite{lattimore2020bandit}). We can note that we don't take an alternative strategy for the prior strategy $\pi'$.
\\
\\
We will also use the following lemma
\begin{lemma}\label{lem:suiteKinf}
    For every $\nu \in \mathcal{P}([0,1])$ and $\mu \in (0,1)$ such that $\mathbb{E}[\nu] < \mu$, we can construct a sequence of distribution $(\tilde \nu^{(n)})_n$ such that we have two positive sequences $(\varepsilon_n)_n$, $(y_n)_n$ that converge to $0$ and for all $n$,
    \begin{align*}
        & \KL(\nu,\tilde \nu^{(n)}) \leq \mathcal{K}_{inf}(\nu,\mu) + y_n, \\
        & \tilde \mu^{(n)} = \mu + \varepsilon_n.
    \end{align*}
    In particular we have for every $\mu' > \mu$, \ $\mathcal{K}_{inf}(\nu,(\mu,\mu')) = \mathcal{K}_{inf}(\nu,\mu)$.
\end{lemma}
(see proof of Lemma \ref{lem:suiteKinf} in \autoref{sec:proofsuiteKinf}).\\
We will also use the following lemma, which is simply a corollary of the data-processing inequality
\begin{lemma}\label{lem:fundamental}
For any $\mathbb{P} \ll \mathbb{Q}$ and any random variable $Z$ in $[0,1]$
    \begin{equation*}
        \KL\left( \mathbb{P},\mathbb{Q}\right) \geq kl(\mathbb{E}_{\mathbb{P}}[Z], \mathbb{E}_{\mathbb{Q}}[Z]).
    \end{equation*}
\end{lemma}
A proof of it can be found in \cite[Lemma.~3.3]{maillard2019mathematics}

\begin{proof}[ of Theorem \ref{thm:lowerbound}] Let us note $\mathcal{G}_a$ the set of couple $(\tilde \nu,\tilde \nu')$ such that $\tilde \mu_a > \mu_1$, \ $\nu_a  \ll \tilde \nu_a$, \ $\nu'_a \ll \tilde \nu'_a$, $|\tilde \mu_a - \tilde \mu'_a| \leq L_a$ and for all $k \neq a$, $(\tilde \nu_k,\tilde \nu'_k) = (\nu_k,\nu'_k)$.
\\
Thanks to Lemma \ref{lem:jointure}, we have for all $(\tilde \nu,\tilde \nu') \in \mathcal{G}_a$,
\begin{equation}\label{eq:ineqlower1}
    \KL\left(P^{I_{N+1}}_{( \nu,\nu'), \pi^{\dagger}},P^{I_{N+1}}_{(\tilde \nu,\tilde\nu'), \pi^{\dagger}}\right) = \mathbb{E}_{\nu',\pi',\nu, \pi }[N_a(T) ]\KL\left(\nu_a, \tilde{\nu}_a\right) + \mathbb{E}_{\nu', \pi' }[N'_a ]\KL\left(\nu'_a, \tilde{\nu}'_a\right).
\end{equation}
By using Lemma \ref{lem:fundamental} with $Z = \frac{N_a(T)}{T}$ we also have
\begin{align}\label{eq:ineqlower2}
    \KL\left(P^{I_{N+1}}_{( \nu,\nu'), \pi^{\dagger}},P^{I_{N+1}}_{(\tilde \nu,\tilde\nu'), \pi^{\dagger}}\right) \geq kl \left( \frac{\mathbb{E}_{\nu',\pi',\nu, \pi }[N_a(T) ]}{T},\frac{\mathbb{E}_{\tilde \nu',\pi',\tilde \nu, \pi }[N_a(T) ]}{T} \right).
\end{align}
For all $(\tilde \nu,\tilde \nu') \in \mathcal{G}_a$, as $\tilde \mu_a > \mu_1 > \mu_a$ we have $\KL\left(\nu_a, \tilde{\nu}_a\right)> 0$. Then by using \eqref{eq:ineqlower1}\eqref{eq:ineqlower2} we get
\begin{align*}
    \mathbb{E}_{\nu',\pi',\nu, \pi }[N_a(T) ] \geq \frac{1}{\KL\left(\nu_a, \tilde{\nu}_a\right)} \left(\!kl\! \left( \frac{\mathbb{E}_{\nu',\pi',\nu, \pi }[N_a(T) ]}{T},\frac{\mathbb{E}_{\tilde \nu',\pi',\tilde \nu, \pi }[N_a(T) ]}{T} \right) \!- \mathbb{E}_{\nu', \pi' }[N'_a ]\KL\left(\nu'_a, \tilde{\nu}'_a\right) \right)_+\!.
\end{align*}
Now we want to control $kl \left( \frac{\mathbb{E}_{\nu',\pi',\nu, \pi }[N_a(T) ]}{T},\frac{\mathbb{E}_{\tilde \nu',\pi',\tilde \nu, \pi }[N_a(T) ]}{T} \right)$. \\
By consistency we have two quantities $G_{1,T} := G_{1,\alpha, \nu',\pi',\nu, \pi,T} > 0$ and $G_{2,T} := G_{2,\alpha,\tilde \nu',\pi',\tilde \nu, \pi,T} > 0$ that both converge to $0$ such that
\begin{itemize}
    \item $\mathbb{E}_{\nu',\pi', \nu, \pi}[N_a(T) ] = G_{1,T} T^{\alpha}$ 
    \item $T - \mathbb{E}_{\tilde \nu',\pi',\tilde \nu, \pi}[N_a(T) ] = G_{2,T}T^{\alpha}$.
\end{itemize}
Then using that 
\begin{align*}
    kl(p,q) &= (1-p)\log\left(\frac{1}{1-q} \right) + \underbrace{p \log(1/q)}_{\geq 0} + \underbrace{p \log(p) + (1-p) \log(1-p)}_{\geq - \log(2)} \\
    & \geq (1-p)\log\left(\frac{1}{1-q} \right) - \log(2), 
\end{align*}
we have 
\begin{align*}
    & kl \left( \frac{\mathbb{E}_{\nu',\pi',\nu, \pi }[N_a(T) ]}{T},\frac{\mathbb{E}_{\tilde \nu',\pi',\tilde \nu, \pi }[N_a(T) ]}{T} \right)\\
    & \geq \left( 1 - G_{1,T}T^{\alpha - 1}  \right) \Big( (1-\alpha) \ln(T) - \ln(G_{2,T}) \Big) - \ln(2) \\
    & = \ln(T) - \Bigg(G_{1,T}T^{\alpha - 1} \ln(T) + \big(1- G_{1,T}T^{\alpha - 1}\big) \big(  \alpha \ln(T) + \ln(G_{2,T})\big) + \ln(2)\Bigg) \\
    & := \ln(T) - R_{\alpha,\nu,\tilde \nu,\nu',\tilde \nu', \pi ,\pi',T}.
\end{align*}
Then we have
\begin{align*}
    \mathbb{E}_{\nu',\pi',\nu, \pi  }[N_a(T)] &\geq \frac{1}{\KL(\nu_a,\tilde \nu_a)} \left(\ln(T) - R_{\alpha,\nu,\tilde \nu,\nu',\tilde \nu', \pi ,\pi',T} - \mathbb{E}_{\nu', \pi' }[N'_a ]\KL\left(\nu'_a, \tilde{\nu}'_a\right) \right)_+ \\
    & \geq \frac{1}{\KL(\nu_a,\tilde \nu_a)} \left(\ln(T)   - \mathbb{E}_{\nu', \pi' }[N'_a ]\KL\left(\nu'_a, \tilde{\nu}'_a\right) \right)_+ - \frac{(R_{\alpha,\nu,\tilde \nu,\nu',\tilde \nu', \pi ,\pi',T})_+}{\KL(\nu_a,\tilde \nu_a)}.
\end{align*}
By dividing by $\ln T$ we get
\begin{align*}
    \frac{ \mathbb{E}_{\nu',\pi',\nu, \pi  }[N_a(T)]}{\ln T} \geq \frac{\left(1 - \frac{\mathbb{E}_{\nu', \pi' }[N'_a ]}{\ln T}\KL\left(\nu'_a, \tilde{\nu}'_a\right) \right)_+}{\KL(\nu_a,\tilde \nu_a)} - \frac{1}{\KL(\nu_a,\tilde \nu_a)} \frac{(R_{\alpha,\nu,\tilde \nu,\nu',\tilde \nu', \pi ,\pi',T})_+}{\ln T}.
\end{align*}
We can note that  for $T$ large enough we have $G_{1,T} \leq 1$ and $G_{2,T} \leq 1$, then
\begin{align*}
    \limsup_{T \to + \infty} \frac{R_{\alpha,\nu,\tilde \nu,\nu',\tilde \nu', \pi ,\pi',T}}{\ln T} & = \limsup_{T \to + \infty} \Bigg(G_{1,T}T^{\alpha - 1}  + \big(1- G_{1,T}T^{\alpha - 1}\big) \bigg(  \alpha + \frac{\ln(G_{2,T})}{\ln T}\bigg) + \frac{\ln(2)}{\ln T}\Bigg)_+ \\
    & \leq \limsup_{T \to + \infty} \Bigg(T^{\alpha - 1}  +  \alpha + \frac{\ln(2)}{\ln T}\Bigg) = \alpha.
\end{align*}
\textbf{Case 1 : if $\mu'_a > \mu_1 - L_a$}
\\
We can take $\tilde \nu'_a := \nu'_a$, which gives $\KL\left(\nu'_a, \tilde{\nu}'_a\right)=0 $. As we are in $\mathcal{G}_a$ we must choose $\tilde \nu_a$ such that $\tilde \mu_a \in ( \mu'_a  - L_a, \mu'_a + L_a)$ and $\tilde \mu_a > \mu_1$. As $\mu'_a + L_a > \mu_1$ it is possible and it gives $\tilde \mu_a \in (\mu_1,  \mu'_a + L_a)$. 
\\
Then we get for $\tilde \nu'_a := \nu'_a$ and $\tilde \nu_a$ such that $\tilde \mu_a \in (\mu_1,  \mu'_a + L_a)$
\begin{align*}
     \frac{ \mathbb{E}_{\nu',\pi',\nu, \pi  }[N_a(T)]}{\ln T} \geq \frac{1}{\KL(\nu_a,\tilde \nu_a) } - \frac{1}{\KL(\nu_a,\tilde \nu_a)} \frac{R_{\alpha,\nu,\tilde \nu,\nu',\tilde \nu', \pi ,\pi',T}}{\ln T},
\end{align*}
by taking the limit when $T \to + \infty$ and then the limit when $\alpha \to 0$ we get
\begin{equation*}
    \liminf_{T \to + \infty}\frac{ \mathbb{E}_{\nu',\pi',\nu, \pi  }[N_a(T)]}{\ln T} \geq \frac{1}{\KL(\nu_a,\tilde \nu_a) },
\end{equation*}
by taking the supremum over $\tilde \nu_a$ such that $\tilde \mu_a \in (\mu_1,  \mu'_a + L_a)$ we get  
\begin{equation*}
    \liminf_{T \to + \infty}\frac{ \mathbb{E}_{\nu',\pi',\nu, \pi  }[N_a(T)]}{\ln T} \geq \frac{1}{\mathcal{K}_{inf}(\nu_a,(\mu_1,  \mu'_a + L_a)) }.
\end{equation*}
Thanks to Lemma \ref{lem:suiteKinf} we have
\begin{equation*}
    \mathcal{K}_{inf}(\nu_a,(\mu_1,  \mu'_a + L_a)) = \mathcal{K}_{inf}(\nu_a,\mu_1).
\end{equation*}
Then we get
\begin{equation*}
    \liminf_{T \to + \infty}\frac{ \mathbb{E}_{\nu',\pi',\nu, \pi  }[N_a(T)]}{\ln T} \geq \frac{1}{\mathcal{K}_{inf}(\nu_a,\mu_1) }.
\end{equation*}
\textbf{Case 2 : if $\mu'_a \leq \mu_1 - L_a$ and $\limsup_{T \to + \infty} \frac{ \mathbb{E}_{\nu',\pi'  }[N'_a]}{\ln T} = C < +\infty$}
\\
As $\tilde \mu_a > \mu_1$, we have $\tilde \mu'_a > \mu_1 - L_a \geq \mu'_a$. Then $\KL\left(\nu'_a, \tilde{\nu}'_a\right) > 0$.
\\
We have with the limit when $T \to + \infty$ and $\alpha \to 0$
\begin{align*}
     \liminf_{T \to + \infty}\frac{ \mathbb{E}_{\nu',\pi',\nu, \pi  }[N_a(T)]}{\ln T} \geq 
    &  \frac{\left(1 - C\KL\left(\nu'_a, \tilde{\nu}'_a\right) \right)_+}{\KL(\nu_a,\tilde \nu_a)}.
\end{align*}
Now by taking the supremum of $\mathcal{G}_a$,
\begin{align*}
     \liminf_{T \to + \infty}\frac{ \mathbb{E}_{\nu',\pi',\nu, \pi  }[N_a(T)]}{\ln T} \geq 
    &  \sup_{(\tilde \nu,\tilde \nu') \in \mathcal{G}_a}\frac{\left(1 - C\KL\left(\nu'_a, \tilde{\nu}'_a\right) \right)_+}{\KL(\nu_a,\tilde \nu_a)}.
\end{align*}
Using Lemma \ref{lem:suiteKinf}, we can take a sequence of $(\tilde \nu^{n}_{a})_n$ such that
\begin{itemize}
    \item $\KL(\nu_a,\tilde \nu^{n}_a) = \mathcal{K}_{inf}(\nu_a,\mu_1) + y_n$
    \item $\tilde \mu^{n}_a = \mu_1 + \varepsilon_n$.
\end{itemize}
where $\varepsilon_n$ and $y_n$ are positive and converge to $0$. 
\\
Because we are in $\mathcal{G}_a$ we must have $\tilde \mu'_a \in (\mu_1 + \varepsilon_n - L_a, \mu_1 + \varepsilon_n + L_a)$ and
\begin{align*}
    \liminf_{T \to + \infty}\frac{ \mathbb{E}_{\nu',\pi',\nu, \pi  }[N_a(T)]}{\ln T} \geq 
    &  \frac{\left(1 - C\mathcal{K}_{inf}(\nu'_a,(\mu_1 + \varepsilon_n - L_a, \mu_1 + \varepsilon_n + L_a) ) \right)_+}{\mathcal{K}_{inf}(\nu_a,\mu_1) + y_n}.
\end{align*}
Now using again Lemma \ref{lem:suiteKinf} we get
\begin{equation*}
    \mathcal{K}_{inf}(\nu'_a,(\mu_1 + \varepsilon_n - L_a, \mu_1 + \varepsilon_n + L_a)) = \mathcal{K}_{inf}(\nu'_a,\mu_1 + \varepsilon_n - L_a),
\end{equation*}
and using the inequality $\frac{1}{x + y} \geq \frac{1}{x} - \frac{y}{x^2}$ for $x>0$ and $y\geq 0$ we get
\begin{align*}
    \liminf_{T \to + \infty}\frac{ \mathbb{E}_{\nu',\pi',\nu, \pi  }[N_a(T)]}{\ln T} \geq 
    &  \frac{\left(1 - C\mathcal{K}_{inf}(\nu'_a,\mu_1 + \varepsilon_n - L_a ) \right)_+}{\mathcal{K}_{inf}(\nu_a,\mu_1) } - \frac{y_n}{\mathcal{K}_{inf}(\nu_a,\mu_1) ^2}.
\end{align*}
By continuity of the $\mathcal{K}_{inf}$ on its second argument (see \cite{JMLR:v16:honda15a}) we have by taking the limit when $n \to + \infty$ 
\begin{align*}
    \liminf_{T \to + \infty}\frac{ \mathbb{E}_{\nu',\pi',\nu, \pi  }[N_a(T)]}{\ln T}  
    & \geq \frac{\left(1 - C\mathcal{K}_{inf}(\nu'_a,\mu_1  - L_a ) \right)_+}{\mathcal{K}_{inf}(\nu_a,\mu_1) }  = \liminf_{T \to + \infty}\frac{\left(1 - \frac{\mathbb{E}_{\nu', \pi' }[N'_a ]}{\ln T}\mathcal{K}_{inf}(\nu'_a,\mu_1  - L_a ) \right)_+}{\mathcal{K}_{inf}(\nu_a,\mu_1)}. 
\end{align*}
\vspace{-2em}
\end{proof}

\subsection{Proof of Lemma \ref{lem:jointure}}\label{sec:proofjointure}

\begin{proof}[of Lemma \ref{lem:jointure}]
We assume throughout that $\nu_i \ll \tilde\nu_i$ and $\nu'_i \ll \tilde\nu'_i$ for all $i\in\{1,...,K\}$; otherwise both sides of the identity are infinite and the statement is vacuous.
\\
\\
Let $t \in \{0,1,...,N\}$. We have $U_{t+1}$ independent from $(I_n,Y_n)$ with an uniform distribution on $[0,1]$, then 
\begin{equation*}
    P^{I_{t+1}}_{(\tilde \nu,\tilde\nu')} = \mathbb P^{(I_t,Y_t)}_{(\tilde \nu,\tilde\nu')} \otimes \lambda, \qquad \lambda := \mathrm{Unif}[0,1],
\end{equation*}
using that $\KL(\lambda,\lambda)=0$ we have
\begin{equation*}
    \KL\left(P^{I_{t+1}}_{( \nu,\nu'), \pi^{\dagger}},P^{I_{t+1}}_{(\tilde \nu,\tilde\nu'), \pi^{\dagger}}\right) = \KL\left(P^{(I_t,Y_t)}_{( \nu,\nu'), \pi^{\dagger}},P^{(I_t,Y_t)}_{(\tilde \nu,\tilde\nu'), \pi^{\dagger}}\right).
\end{equation*}
Now we use the decomposition 
\begin{equation*}
    dP^{(I_t,Y_t)}_{(\tilde \nu,\tilde\nu'), \pi^{\dagger}}(i_t,y_t) = dP^{I_t}_{(\tilde \nu,\tilde\nu'), \pi^{\dagger}}(i_t) dP^{Y_t}_{(\tilde \nu,\tilde\nu'), \pi^{\dagger}}(y_t \mid I_t = i_t).
\end{equation*}
By integrating the log-ratio we get
\begin{align*}
    \KL\left(P^{(I_t,Y_t)}_{( \nu,\nu'), \pi^{\dagger}},P^{(I_t,Y_t)}_{(\tilde \nu,\tilde\nu'), \pi^{\dagger}}\right) \!&=  \KL\left(P^{I_{t}}_{( \nu,\nu'), \pi^{\dagger}},P^{I_{t}}_{(\tilde \nu,\tilde\nu'), \pi^{\dagger}}\right) + \mathbb{E}_{P^{I_{t}}_{(\tilde \nu,\tilde\nu'), \pi^{\dagger}}}\!\!\left[\KL \left(P^{Y_t}_{( \nu,\nu'), \pi^{\dagger}}( \cdot \mid I_t ),P^{Y_t}_{(\tilde \nu,\tilde\nu'), \pi^{\dagger}}(\cdot\! \mid I_t ) \right)\right]. 
\end{align*}
However, conditionally to $I_t$, the chosen arm is the same for both instance $(\tilde \nu,\tilde\nu')$ and $( \nu,\nu')$ as $A_t = \pi^{\dagger}(I_t)$, then
\begin{align*}
    \mathbb{E}_{P^{I_{t}}_{(\tilde \nu,\tilde\nu'), \pi^{\dagger}}}\left[\KL \left(P^{Y_t}_{( \nu,\nu'), \pi^{\dagger}}( \cdot \mid I_t ),P^{Y_t}_{(\tilde \nu,\tilde\nu'), \pi^{\dagger}}(\cdot \mid I_t ) \right)\right] =  \begin{cases}
\mathbb{E}_{(\nu,\nu'),\pi^{\dagger}}\left[\KL(\nu'_{A_t},\tilde\nu'_{A_t})\right], & t\le T',\\[2pt]
\mathbb{E}_{(\nu,\nu'),\pi^{\dagger}}\left[\KL(\nu_{A_t},\tilde\nu_{A_t})\right], & t> T'.
\end{cases}
\end{align*}
Then we have
\begin{align*}
     \KL\left(P^{I_{t+1}}_{( \nu,\nu'), \pi^{\dagger}},P^{I_{t+1}}_{(\tilde \nu,\tilde\nu'), \pi^{\dagger}}\right) & = \KL\left(P^{(I_t,Y_t)}_{( \nu,\nu'), \pi^{\dagger}},P^{(I_t,Y_t)}_{(\tilde \nu,\tilde\nu'), \pi^{\dagger}}\right) \\
     & = \KL\left(P^{I_{t}}_{( \nu,\nu'), \pi^{\dagger}},P^{I_{t}}_{(\tilde \nu,\tilde\nu'), \pi^{\dagger}}\right) +\begin{cases}
\mathbb{E}_{(\nu,\nu'),\pi^{\dagger}}\left[\KL(\nu'_{A_t},\tilde\nu'_{A_t})\right], & t\le T',\\[2pt]
\mathbb{E}_{(\nu,\nu'),\pi^{\dagger}}\left[\KL(\nu_{A_t},\tilde\nu_{A_t})\right], & t> T'.
\end{cases}
\end{align*}
Then using that 
\begin{equation*}
    \KL\left(P^{I_{1}}_{( \nu,\nu'), \pi^{\dagger}},P^{I_{1}}_{(\tilde \nu,\tilde\nu'), \pi^{\dagger}}\right) = \KL(\lambda,\lambda)=0,
\end{equation*}
we get
\begin{align*}
    \KL\left(P^{I_{N+1}}_{( \nu,\nu'), \pi^{\dagger}},P^{I_{N+1}}_{(\tilde \nu,\tilde\nu'), \pi^{\dagger}}\right) & = \sum_{t = 1}^{T'}\mathbb{E}_{(\nu,\nu'),\pi^{\dagger}}\left[\KL(\nu'_{A_t},\tilde\nu'_{A_t})\right] + \sum_{t = T'+1}^{N}\mathbb{E}_{(\nu,\nu'),\pi^{\dagger}}\left[\KL(\nu_{A_t},\tilde\nu_{A_t})\right].
\end{align*}
Now we have
\begin{align*}
    \sum_{t = 1}^{T'}\mathbb{E}_{(\nu,\nu'),\pi^{\dagger}}\left[\KL(\nu'_{A_t},\tilde\nu'_{A_t})\right] & = \sum_{t = 1}^{T'}\sum_{i = 1}^{K}\mathbb{E}_{(\nu,\nu'),\pi^{\dagger}}\left[\mathbf{1}\left(A_t = i \right)\KL(\nu'_{i},\tilde\nu'_{i})\right] \\
    & = \sum_{i = 1}^{K} \mathbb{E}_{(\nu,\nu'),\pi^{\dagger}}\left[\sum_{t = 1}^{T'}{1}\left(A_t = i \right) \right] \KL(\nu'_{i},\tilde\nu'_{i}) \\
    & = \sum_{i = 1}^{K} \mathbb{E}_{(\nu,\nu'),\pi^{\dagger}}\left[N'_i \right] \KL(\nu'_{i},\tilde\nu'_{i}), 
\end{align*}
we also have
\begin{equation*}
    \sum_{t = T'+1}^{N}\mathbb{E}_{(\nu,\nu'),\pi^{\dagger}}\left[\KL(\nu_{A_t},\tilde\nu_{A_t})\right] = \sum_{i = 1}^{K} \mathbb{E}_{(\nu,\nu'),\pi^{\dagger}}\left[N_i(T) \right] \KL(\nu_{i},\tilde\nu_{i}),
\end{equation*}
then
\begin{equation*}
    \KL\left(P^{I_{N+1}}_{( \nu,\nu'), \pi^{\dagger}},P^{I_{N+1}}_{(\tilde \nu,\tilde\nu'), \pi^{\dagger}}\right) = \sum_{i = 1}^{K} \mathbb{E}_{(\nu,\nu'),\pi^{\dagger}}\left[N'_i \right] \KL(\nu'_{i},\tilde\nu'_{i}) + \sum_{i = 1}^{K} \mathbb{E}_{(\nu,\nu'),\pi^{\dagger}}\left[N_i(T) \right] \KL(\nu_{i},\tilde\nu_{i}).
\end{equation*}

\vspace{-1em}
\end{proof}
\subsection{Proof of Lemma \ref{lem:suiteKinf}}\label{sec:proofsuiteKinf}

\begin{proof}[of Lemma \ref{lem:suiteKinf}] As $\mathcal{K}_{inf}(\nu,\mu)$ is an infimum, we have access to a sequence of distributions $(\lambda^{(n)})_n$ such that 
\begin{align*}
    &(\KL(\nu,\lambda^{(n)}))_n \text{ converges to }\mathcal{K}_{inf}(\nu,\mu) \\
    & \forall n, \ \mu_{\lambda^{(n)}} > \mu.
\end{align*}
We however have no information about an hypothetical convergence of $\mu_{\lambda^{(n)}}$ to $\mu$.
\\
Let us fix $n$, we will use the convexity of $\mathcal{P}([0,1])$ by defining for every $t \in (0,1)$
\begin{equation*}
    \gamma^{(n)}_t := (1-t) \lambda^{(n)} + t \nu.
\end{equation*}
We have $\mu_{\gamma^{(n)}_t} \in (\mathbb{E}[\nu], \mu_{\lambda^{(n)}})$. As $\mathbb{E}[\nu] < \mu < \mu_{\lambda^{(n)}}$ and by continuity of $t \mapsto \gamma^{(n)}_t$, for every positive number $\varepsilon_n$ small enough, we can find $t^*_n \in (0,1)$ such that $\mu_{\gamma^{(n)}_{t^*_n}} = \mu + \varepsilon_n$.
\\
We define $ \tilde \nu^{(n)} := \gamma^{(n)}_{t^*_n}$. With the convexity of the $\KL$ we have
\begin{equation*}
    \KL(\nu,\tilde \nu^{(n)}) \leq (1-t^*_n)\KL(\nu,\lambda^{(n)}) \leq \KL(\nu,\lambda^{(n)}) =: \mathcal{K}_{inf}(\nu,\mu) + y_n.
\end{equation*}
where $y_n$ is positive and converges to $0$ by definition of $\lambda^{(n)}$. This gives the first part of the lemma. Now to prove the second part of the Lemma, we have for the first inequality
\begin{equation*}
    \mathcal{K}_{inf}(\nu,\mu) := \mathcal{K}_{inf}(\nu,(\mu,1)) \leq \mathcal{K}_{inf}(\nu,(\mu,\mu')).
\end{equation*}
But at the same time we can find a sequence $(\tilde \nu^{n}_a)_n$ and a sequence $(y_n)_n$ that converges to $0$ such that for all $n > T_0$ large enough,
\begin{align*}
        & \KL(\nu,\tilde \nu^{(n)}) \leq \mathcal{K}_{inf}(\nu,\mu) + y_n \\
        & \tilde \mu^{(n)} \leq   \frac{\mu + \mu'}{2}.
    \end{align*}
Then we have
\begin{equation*}
    \mathcal{K}_{inf}(\nu,\mu) \geq \KL(\nu,\tilde \nu^{(n)}) - y_n \geq \mathcal{K}_{inf}(\nu,(\mu,\mu')) - y_n.
\end{equation*}
By taking the limit when $n \to + \infty$ we get $\mathcal{K}_{inf}(\nu,\mu) \geq  \mathcal{K}_{inf}(\nu,(\mu,\mu'))$.

\end{proof}

\end{document}